\PassOptionsToPackage{table}{xcolor} % acmart loads xcolor itself; a second \usepackage with options clashes
\documentclass[sigconf,nonacm]{acmart}

\usepackage{balance} % for balancing columns on the final page

\usepackage{multirow}
\usepackage[colorinlistoftodos]{todonotes}
\usepackage{cleveref}
\usepackage{natbib}

\usepackage{pifont}
\usepackage{booktabs}
\usepackage[most]{tcolorbox}

\definecolor{checkgreen}{HTML}{287A46}
\definecolor{crossred}{HTML}{B64040}

\newcommand{\cmark}{\textcolor{checkgreen}{\ding{51}}}
\newcommand{\xmark}{\textcolor{crossred}{\ding{55}}}
\usepackage{tikz}
\usetikzlibrary{arrows.meta,positioning,fit,calc,backgrounds,shapes.geometric,shapes.misc,decorations.pathreplacing,shadows}
\usepackage{lipsum}

\title[Swarm Telepathy]{Swarm Telepathy: Prefill-Free KV-Cache Transfer Across Families}

\newcommand{\aff}[1]{\textsuperscript{\normalfont #1}}
\newcommand{\projecturl}{https://github.com/Ar-Kareem/SwaT}
\usepackage{fontawesome5}

\author{\mbox{Abdulrahman Alabdulkareem\aff{1,$*$}}\quad \mbox{Irene Tenison\aff{1}}\quad \mbox{Yang Zhang\aff{2}}\\
  \mbox{Yada Zhu\aff{2}}\quad \mbox{Boris Katz\aff{1}}\quad \mbox{Dan Gutfreund\aff{2}}}
\affiliation[obeypunctuation=true]{
  \institution{\rule{0pt}{2.1em}\aff{1}MIT CSAIL\qquad\aff{2}MIT-IBM Computing Research Lab}\\
  \country{\rule{0pt}{2.0em}\normalsize\faGithub\ Project Page: \textcolor{ACMDarkBlue}{\url{\projecturl}}}}

\begin{abstract}

LLM multi-agent systems split work across agents that communicate through text. An agent summarizes its findings, another agent receives the text. This constrains communication to discrete tokens and costs redundant compute, limiting bandwidth and capabilities of multi-agent systems.
Recent works demonstrate KV-cache transfer between LLMs but they either require the LLMs to be within the same family or cannot function across context, thus requiring the receiver to prefill text.
We present \swat{} (\underline{Swa}rm \underline{T}elepathy), the first cross-family KV-cache transfer method that allows prefill-free communication in multi-family multi-agent systems.
In an agent-to-agent communication setting, combining \swat{} with recent advances in KV-cache compaction \citep{yoon_attentionmatching} beats text-to-text communication in both accuracy and speed, with a mean of $40\%$ higher accuracy and a mean of $4\times$ faster wall-clock time.
Experiments show \swat{} achieves high performance on 17 benchmarks across 12 LLMs and 6 families. Additionally, given a sender Vision-Language Model (VLM), \swat{} is able to transfer the vision capabilities to a text-only receiver LLM exceeding text-to-text communication despite \swat{} being trained on text-only data. 
\swat{} keeps both sender and receiver models frozen without fine-tuning and only requires $<50M$ tokens and $<1$ GPU-day to train.

\end{abstract}


\newcommand{\BibTeX}{\rm B\kern-.05em{\sc i\kern-.025em b}\kern-.08em\TeX}

\crefname{appendix}{Appendix}{Appendices}
\Crefname{appendix}{Appendix}{Appendices}

\newcommand{\swat}{\textsc{SwaT}}
\newcommand{\ttot}{\textsc{T2T}}
\newcommand{\am}{\textsc{AM}}

\begin{document}

%%% The following commands remove the headers in your paper. For final 
%%% papers, these will be inserted during the pagination process.

%\pagestyle{fancy}
%\fancyhead{}

%%% The next command prints the information defined in the preamble.

\maketitle
{\renewcommand{\thefootnote}{$*$}\footnotetext{Corresponding author: \href{mailto:arkareem@mit.edu}{arkareem@mit.edu}}} 

%%%%%%%%%%%%%%%%%%%%%%%%%%%%%%%%%%%%%%%%%%%%%%%%%%%%%%%%%%%%%%%%%%%%%%%%

\section{Introduction}

Multi-agent systems (MAS) built from large language models (LLMs) have progressed rapidly, with agents that specialize, divide tasks, and pool expertise~\citep{LLM_multitask1,LLM_multitask2,LLM_multitask3,LLM_multitask4}, including systems that drive scientific discovery~\citep{mas_scidiscover}. In nearly all of these systems agents communicate in natural language: one agent reads its inputs, writes what it found, and the next agent reads that text. This simple protocol has delivered consistent gains over single-model performance~\citep{mas_text1,mas_text2,mas_text3,mas_text4,mas_text5,mas_text6,mas_text7,mas_text8,vighnesh_finetune}.

% V17d: v17 with the grey cloud saying 'I get the gist'
% V17: v01 lanes (left 2/3) + v05 scoreboard (right 1/3)
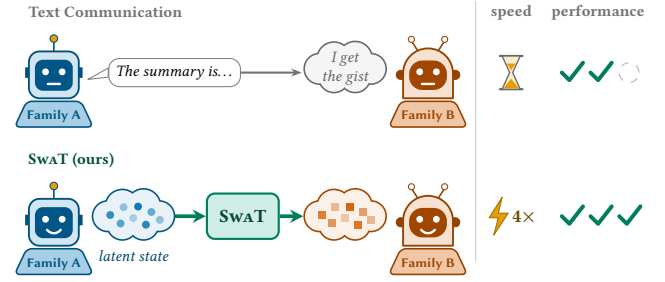
\begin{figure}
\centering
\definecolor{snd}{RGB}{0,114,178}\definecolor{rcv}{RGB}{213,94,0}\definecolor{swt}{RGB}{0,158,115}
\definecolor{vis}{RGB}{204,121,167}\definecolor{ylw}{RGB}{230,159,0}\definecolor{sky}{RGB}{86,180,233}
\definecolor{gry}{RGB}{140,140,140}\definecolor{bad}{RGB}{190,30,45}
\tikzset{>={Stealth[length=4pt]}, font=\footnotesize,
  flow/.style={->, thick, black!55},
  tp/.style={-{Stealth[length=5pt,width=5pt]}, line width=1.4pt, swt!80!black},
  swat/.style={draw=swt!80!black, fill=swt!18, thick, rounded corners=3pt, minimum height=0.6cm,
    minimum width=1.05cm, font=\footnotesize\bfseries, text=swt!45!black},
  say/.style={draw=black!50, fill=white, rounded corners=4pt, inner sep=2.5pt, font=\scriptsize\itshape},
  lab/.style={font=\scriptsize\bfseries, text=black!55, anchor=west},
  pics/botA/.style={code={
    \draw[#1!70!black, line width=0.6pt] (0,0.4) -- (0,0.55); \fill[#1!70!black] (0,0.6) circle (0.055);
    \draw[#1!70!black, line width=0.7pt, fill=#1!22, rounded corners=3pt] (-0.38,-0.36) rectangle (0.38,0.4);
    \fill[white] (-0.16,0.08) circle (0.1) (0.16,0.08) circle (0.1);
    \fill[#1!75!black] (-0.16,0.08) circle (0.055) (0.16,0.08) circle (0.055);
    \draw[#1!75!black, line width=0.7pt, line cap=round] (-0.13,-0.18) -- (0.13,-0.18);}},
  pics/botB/.style={code={
    \fill[#1!70!black, rounded corners=1pt] (-0.47,-0.1) rectangle (-0.35,0.12) (0.35,-0.1) rectangle (0.47,0.12);
    \draw[#1!70!black, line width=0.7pt, fill=#1!22] (0,0.01) circle (0.39);
    \fill[white] (-0.25,-0.01) rectangle (-0.06,0.17) (0.06,-0.01) rectangle (0.25,0.17);
    \fill[#1!75!black] (-0.195,0.04) rectangle (-0.115,0.12) (0.115,0.04) rectangle (0.195,0.12);
    \draw[#1!75!black, line width=0.7pt, line cap=round] (-0.12,-0.15) .. controls (-0.05,-0.23) and (0.05,-0.23) .. (0.12,-0.15);}},
  pics/botC/.style={code={
    \draw[#1!70!black, line width=0.6pt] (-0.14,0.36) -- (-0.24,0.54) (0.14,0.36) -- (0.24,0.54);
    \fill[#1!70!black] (-0.24,0.56) circle (0.045) (0.24,0.56) circle (0.045);
    \draw[#1!70!black, line width=0.7pt, fill=#1!22, line join=round] (-0.28,0.38) -- (0.28,0.38) -- (0.42,0.01) -- (0.28,-0.36) -- (-0.28,-0.36) -- (-0.42,0.01) -- cycle;
    \fill[white, rounded corners=2pt] (-0.26,0.0) rectangle (0.26,0.18);
    \fill[#1!75!black] (-0.12,0.09) circle (0.05) (0.12,0.09) circle (0.05);
    \foreach \x in {-0.1,0,0.1} \draw[#1!75!black, line width=0.6pt] (\x,-0.13) -- (\x,-0.24);}},
  pics/botD/.style={code={
    \draw[#1!70!black, line width=0.6pt] (0,0.42) -- (0.06,0.5) -- (-0.04,0.56) -- (0.02,0.64);
    \draw[#1!70!black, line width=0.7pt, fill=#1!22] (-0.32,-0.36) -- (-0.32,0.16) arc[start angle=180, end angle=0, radius=0.32] -- (0.32,-0.36) -- cycle;
    \fill[white] (0,0.12) circle (0.15); \fill[#1!75!black] (0,0.12) circle (0.075);
    \draw[#1!75!black, line width=0.6pt] (-0.14,-0.2) rectangle (0.14,-0.1) (0,-0.2) -- (0,-0.1);}},
  pics/cloud/.style={code={
    \foreach \x/\y/\r in {-0.42/0/0.24,-0.18/0.16/0.27,0.17/0.17/0.27,0.43/0/0.24,0.17/-0.14/0.24,-0.18/-0.14/0.24}
      \draw[#1!75!black, line width=1.2pt] (\x,\y) circle (\r);
    \foreach \x/\y/\r in {-0.42/0/0.24,-0.18/0.16/0.27,0.17/0.17/0.27,0.43/0/0.24,0.17/-0.14/0.24,-0.18/-0.14/0.24}
      \fill[#1!9] (\x,\y) circle (\r);
    \fill[#1!9] (-0.42,-0.12) rectangle (0.43,0.12);}},
  pics/thA/.style={code={\foreach \x/\y/\t in {-0.38/0.02/80,-0.16/0.15/45,0.08/0.16/95,0.3/0.07/60,-0.21/-0.12/65,0.0/-0.03/35,0.21/-0.14/85,0.44/-0.06/45}
      \fill[#1!\t] (\x,\y) circle (0.065);}},
  pics/thB/.style={code={\foreach \x/\y/\t in {-0.38/0.02/80,-0.16/0.15/45,0.08/0.16/95,0.3/0.07/60,-0.21/-0.12/65,0.0/-0.03/35,0.21/-0.14/85,0.44/-0.06/45}
      \fill[#1!\t] (\x-0.06,\y-0.06) rectangle (\x+0.06,\y+0.06);}},
  pics/thC/.style={code={\foreach \x/\y/\t in {-0.38/0.02/80,-0.16/0.15/45,0.08/0.16/95,0.3/0.07/60,-0.21/-0.12/65,0.0/-0.03/35,0.21/-0.14/85,0.44/-0.06/45}
      \fill[#1!\t] (\x,\y+0.075) -- (\x-0.07,\y-0.05) -- (\x+0.07,\y-0.05) -- cycle;}},
  pics/thD/.style={code={\foreach \x/\y/\t in {-0.38/0.02/80,-0.16/0.15/45,0.08/0.16/95,0.3/0.07/60,-0.21/-0.12/65,0.0/-0.03/35,0.21/-0.14/85,0.44/-0.06/45}
      \fill[#1!\t] (\x,\y+0.08) -- (\x+0.07,\y) -- (\x,\y-0.08) -- (\x-0.07,\y) -- cycle;}},
  pics/thBfew/.style={code={\foreach \x/\y/\t in {-0.16/0.15/30,0.21/-0.14/40} \fill[#1!\t] (\x-0.06,\y-0.06) rectangle (\x+0.06,\y+0.06);}},
  pics/tail/.style={code={\fill[white, draw=black!50, line join=round] (0,0.06) -- (-0.27,-0.09) -- (0,-0.06) -- cycle;}},
  pics/trail/.style={code={\fill[#1] (0,0) circle (0.035) (0.11,0.08) circle (0.05);}},
  pics/doc/.style={code={
    \draw[black!60, fill=white, line width=0.5pt] (-0.18,-0.23) -- (0.18,-0.23) -- (0.18,0.13) -- (0.08,0.23) -- (-0.18,0.23) -- cycle;
    \draw[black!60, line width=0.5pt] (0.08,0.23) -- (0.08,0.13) -- (0.18,0.13);
    \foreach \y in {-0.14,-0.06,0.02,0.1} \draw[black!40] (-0.12,\y) -- (0.12,\y);}},
  pics/book/.style={code={
    \draw[black!60, fill=white, line width=0.5pt, line join=round] (0,-0.2) -- (-0.4,-0.14) -- (-0.4,0.24) -- (0,0.18) -- cycle
      (0,-0.2) -- (0.4,-0.14) -- (0.4,0.24) -- (0,0.18) -- cycle;
    \foreach \y in {-0.06,0.02,0.1} \draw[black!35] (-0.34,\y+0.03) -- (-0.06,\y-0.01) (0.06,\y-0.01) -- (0.34,\y+0.03);}},
  pics/bus/.style={code={
    \draw[#1!60!black, rounded corners=1.5pt, fill=#1] (-0.45,-0.17) rectangle (0.45,0.25);
    \foreach \x in {-0.37,-0.17,0.03} \fill[white] (\x,0.03) rectangle ++(0.15,0.14);
    \fill[white] (0.25,-0.11) rectangle (0.38,0.17);
    \fill[black!80] (-0.25,-0.17) circle (0.07) (0.25,-0.17) circle (0.07);}},
  pics/scene/.style={code={
    \fill[#1!55] (0.33,0.2) circle (0.09);
    \fill[#1!50!black] (0.12,-0.28) rectangle (0.16,-0.12);
    \fill[#1!70] (0.14,-0.05) circle (0.11);
    \fill[#1!30] (-0.42,-0.28) rectangle (-0.1,-0.03);
    \fill[#1!85!black] (-0.47,-0.03) -- (-0.26,0.16) -- (-0.05,-0.03) -- cycle;
    \fill[white] (-0.3,-0.28) rectangle (-0.21,-0.13);}},
  pics/house/.style={code={
    \draw[#1, line width=0.6pt, line join=round] (-0.16,-0.28) rectangle (0.16,-0.03) (-0.21,-0.03) -- (0,0.16) -- (0.21,-0.03) -- cycle;}},
  pics/map/.style={code={
    \draw[#1!70!black, fill=#1!12, line width=0.5pt, line join=round] (-0.45,-0.3) -- (-0.15,-0.22) -- (0.15,-0.3) -- (0.45,-0.22) -- (0.45,0.3) -- (0.15,0.22) -- (-0.15,0.3) -- (-0.45,0.22) -- cycle;
    \draw[#1!50, line width=0.4pt] (-0.15,-0.22) -- (-0.15,0.3) (0.15,-0.3) -- (0.15,0.22);
    \draw[#1!80!black, dashed, line width=0.7pt] (-0.36,-0.16) .. controls (-0.1,-0.15) and (-0.25,0.15) .. (0.05,0.1) .. controls (0.2,0.06) and (0.2,-0.1) .. (0.3,0.1);
    \draw[bad, line width=0.9pt] (0.25,0.05) -- (0.35,0.15) (0.25,0.15) -- (0.35,0.05);}},
  pics/hourglass/.style={code={
    \draw[black!60, line width=0.8pt, line cap=round] (-0.15,0.22) -- (0.15,0.22) (-0.15,-0.22) -- (0.15,-0.22);
    \draw[black!60, fill=ylw!20, line width=0.5pt] (-0.12,0.22) -- (0.12,0.22) -- (0.02,0) -- (0.12,-0.22) -- (-0.12,-0.22) -- (-0.02,0) -- cycle;
    \fill[ylw] (-0.1,-0.205) -- (0.1,-0.205) -- (0,-0.08) -- cycle; \fill[ylw] (-0.06,0.12) -- (0.06,0.12) -- (0,0.03) -- cycle;}},
  pics/bolt/.style={code={
    \draw[ylw!70!black, fill=ylw, line width=0.4pt, line join=round] (0.05,0.25) -- (-0.12,-0.02) -- (0,-0.02) -- (-0.06,-0.25) -- (0.13,0.04) -- (0,0.04) -- cycle;}},
  pics/bulb/.style={code={
    \draw[ylw!70!black, fill=ylw!55, line width=0.5pt] (0,0.08) circle (0.17);
    \fill[black!45] (-0.08,-0.17) rectangle (0.08,-0.07);
    \foreach \a in {20,90,160} \draw[ylw!80!black, line width=0.7pt, line cap=round] ($(\a:0.24)+(0,0.08)$) -- ++(\a:0.1);}},
  pics/cross/.style={code={\draw[bad, line width=1.4pt, line cap=round] (-0.12,-0.12) -- (0.12,0.12) (-0.12,0.12) -- (0.12,-0.12);}},
  pics/check/.style={code={\draw[swt!80!black, line width=1.5pt, line cap=round, line join=round] (-0.13,0) -- (-0.03,-0.1) -- (0.15,0.13);}},
  pics/flake/.style={code={\foreach \a in {0,60,120}{\draw[#1, line width=0.6pt] (\a:0.12) -- (\a+180:0.12);
    \foreach \s in {1,-1}\foreach \b in {\a,\a+180}\draw[#1, line width=0.45pt] (\b:0.07) -- ++(\b+\s*45:0.045);}}},
  pics/flame/.style={code={
    \fill[orange!85!red] (0,0) .. controls (-0.13,0) and (-0.14,0.13) .. (-0.06,0.19) .. controls (-0.04,0.22) and (-0.04,0.26) .. (-0.01,0.31)
      .. controls (0.03,0.24) and (0.11,0.2) .. (0.11,0.1) .. controls (0.11,0.03) and (0.06,0) .. (0,0) -- cycle;
    \fill[yellow!80!orange] (0,0.015) .. controls (-0.07,0.015) and (-0.07,0.09) .. (-0.02,0.14) .. controls (0,0.12) and (0.06,0.1) .. (0.06,0.06)
      .. controls (0.06,0.03) and (0.03,0.015) .. (0,0.015) -- cycle;}},
  pics/waves/.style={code={\foreach \r in {0.12,0.24,0.36} \draw[#1, line width=0.9pt, line cap=round] (-40:\r) arc[start angle=-40, end angle=40, radius=\r];}},
  pics/gear/.style n args={3}{code={
    \foreach \i in {1,...,#2} \fill[#1!45, rotate=\i*360/#2] (-0.07,#3-0.05) rectangle (0.07,#3+0.12);
    \fill[#1!45] (0,0) circle (#3); \fill[white] (0,0) circle (#3*0.62);}},
}
\begin{tikzpicture}
\tikzset{
  pics/agentA/.style 2 args={code={
    \draw[#1!70!black, line width=0.6pt, fill=#1!35, rounded corners=2pt] (-0.52,-0.67) -- (-0.36,-0.33) -- (0.36,-0.33) -- (0.52,-0.67) -- cycle;
    \node[font=\sffamily\bfseries\tiny, text=#1!75!black, inner sep=0pt] at (0,-0.53) {Family A};
    \draw[#1!70!black, line width=0.6pt] (0,0.33) -- (0,0.46); \draw[#1!70!black, fill=ylw, line width=0.4pt] (0,0.5) circle (0.05);
    \fill[#1!70!black] (-0.38,-0.02) rectangle (-0.33,0.12) (0.33,-0.02) rectangle (0.38,0.12);
    \draw[#1!70!black, line width=0.7pt, fill=#1!18, rounded corners=4pt] (-0.33,-0.29) rectangle (0.33,0.33);
    \fill[#1!75!black, rounded corners=3pt] (-0.25,-0.19) rectangle (0.25,0.22);
    \fill[white] (-0.11,0.06) circle (0.045) (0.11,0.06) circle (0.045);
    \ifnum#2=1 \draw[white, line width=0.9pt, line cap=round] (-0.09,-0.08) .. controls (-0.04,-0.14) and (0.04,-0.14) .. (0.09,-0.08);
    \else \draw[white, line width=0.9pt, line cap=round] (-0.08,-0.1) -- (0.08,-0.1); \fi}},
  pics/agentB/.style 2 args={code={
    \draw[#1!70!black, line width=0.6pt, fill=#1!35, rounded corners=2pt] (-0.52,-0.67) -- (-0.36,-0.33) -- (0.36,-0.33) -- (0.52,-0.67) -- cycle;
    \node[font=\sffamily\bfseries\tiny, text=#1!75!black, inner sep=0pt] at (0,-0.53) {Family B};
    \draw[#1!70!black, line width=0.6pt] (-0.13,0.33) -- (-0.22,0.47) (0.13,0.33) -- (0.22,0.47);
    \draw[#1!70!black, fill=white, line width=0.4pt] (-0.22,0.49) circle (0.04) (0.22,0.49) circle (0.04);
    \draw[#1!70!black, fill=#1!70!black] (-0.36,0.02) circle (0.06) (0.36,0.02) circle (0.06);
    \draw[#1!70!black, line width=0.7pt, fill=#1!18, line join=round] (-0.33,-0.29) -- (-0.33,0.04) arc[start angle=180, end angle=0, radius=0.33] -- (0.33,-0.29) -- cycle;
    \fill[#1!75!black, rounded corners=5pt] (-0.25,-0.18) rectangle (0.25,0.21);
    \fill[white] (-0.11,0.06) circle (0.045) (0.11,0.06) circle (0.045);
    \ifnum#2=1 \draw[white, line width=0.9pt, line cap=round] (-0.09,-0.08) .. controls (-0.04,-0.14) and (0.04,-0.14) .. (0.09,-0.08);
    \else \draw[white, line width=0.9pt, line cap=round] (-0.08,-0.1) -- (0.08,-0.1); \fi}}}
\node[lab] at (-0.45,0.85) {Text Communication};
\pic at (0,0) {agentA={snd}{0}};
\node[say, anchor=west] (b) at (0.72,0.05) {The summary is\dots}; \pic at (b.west) {tail};
\draw[flow] (b.east) -- (3.28,0.05);
\pic[scale=0.8] at (3.85,0.1) {cloud=gry};
\node[font=\scriptsize\itshape, text=gry!70!black, align=center, inner sep=0pt, execute at begin node=\setlength{\baselineskip}{7pt}] at (3.85,0.11) {I get\\the gist};
\pic at (4.95,0) {agentB={rcv}{0}};
\begin{scope}[yshift=-1.95cm]
\node[lab, text=swt!50!black] at (-0.45,0.85) {\swat{} (ours)};
\pic at (0,0) {agentA={snd}{1}};
\pic[scale=0.8] at (1.05,0.1) {cloud=snd}; \pic[scale=0.8] at (1.05,0.1) {thA=snd};
\node[font=\scriptsize\itshape, text=snd!60!black] at (1.05,-0.42) {latent state};
\node[swat, minimum width=0.95cm] (s) at (2.5,0.1) {\swat{}};
\draw[tp] (1.6,0.1) -- (s.west); \draw[tp] (s.east) -- (3.3,0.1);
\pic[scale=0.8] at (3.85,0.1) {cloud=rcv}; \pic[scale=0.8] at (3.85,0.1) {thB=rcv};
\pic at (4.95,0) {agentB={rcv}{1}};
\end{scope}
\draw[black!20] (5.58,1.0) -- (5.58,-2.45);
\node[font=\scriptsize\bfseries, text=black!55] at (6.05,0.85) {speed};
\node[font=\scriptsize\bfseries, text=black!55] at (7.2,0.85) {performance};
\pic at (6.05,0.05) {hourglass};
\foreach \i in {0,1} \pic at (6.85+\i*0.38,0.05) {check};
\draw[black!30, dashed] (7.61,0.05) circle (0.13);
\pic at (5.88,-1.85) {bolt}; \node[font=\footnotesize\bfseries, text=ylw!45!black, anchor=west, inner sep=1pt] at (6.02,-1.85) {4$\times$};
\foreach \i in {0,1,2} \pic at (6.85+\i*0.38,-1.85) {check};
\end{tikzpicture}
\caption{
Communication between two LLM agents from different families. Top: communicating in words is slow and the receiver only gets the gist. Bottom: \swat{} translates the sender's latent state (KV-cache) to the receiver's latent space. Combined with equal-budget cache compression, \swat{} is on average 40\% more accurate than text summarization, at $4\times$ the speed.
}
\label{fig:intro}
\Description{Left two thirds, two rows. Top row, labelled Text Communication: a blue square-headed robot labelled Family A with a neutral face speaks a text bubble reading The summary is, and an orange dome-headed robot labelled Family B with a neutral face ends up with a grey thought cloud reading I get the gist. Bottom row, labelled SwaT (ours): both robots smile; the blue robot's thought cloud of blue circles, labelled latent state, passes through a green SwaT box and becomes a full thought cloud of orange squares next to the orange robot. Right third, a small scoreboard with columns speed and performance: the words row has an hourglass and two check marks plus an empty slot; the SwaT row has a lightning bolt marked 4 times and three check marks.}
\end{figure}

Text is nonetheless a narrow channel. A summary is bounded by what its author chooses to write and by what the reader can reconstruct from it. This process is also costly: the sender must generate the output (decode token by token), and the receiver must consume it from scratch (prefill) before doing any work of its own. Internally, however, the sender has already computed a far richer representation of its context, its key-value cache (KV-cache), which holds every layer's view of every token and represents the entire state an agent requires to generate the next token. The KV-cache is therefore a natural channel for agent communication: the sender agent has already computed it and it is denser than text.

Recent work confirms the promise of this communication channel when all agents share a single backbone. 
LatentMAS~\citep{LatentMAS} shows that agents exchanging latent states and caches carry substantially more information per token than agents exchanging text. 
Similarly, single-model latent chain-of-thought~\citep{latentCOT} reaches the same conclusion. 
But these results apply to a single model communicating within copies of itself, where sender and receiver caches have similar geometry.

Recent KV-cache transfer across context relaxes this constraint to models of different sizes and layer counts within one family~\citep{kv_nvidia,kv_cachebridge,kv_google,kv_dense_latent,kv_lcf,kv_icarus}. However, models in the same family (e.g., Qwen~3 at 4B, 8B, 14B, and 32B) are trained by the same team on similar pipelines and data, and smaller members are often distilled. This leads to the conclusion that a translation between the internal states of such models is a reasonable task. However, in this work we try to tackle the fundamental problem in replicating this protocol in more complex multi-agent systems which mix models of various sizes, architectures, and families while maintaining that the receiver does not prefill any text. This raises the question we address:

\begin{tcolorbox}[
  colback=gray!8, colframe=gray!60, boxrule=0.5pt, arc=1pt,
  left=6pt, right=6pt, top=4pt, bottom=4pt,
]
\emph{Can LLMs from different families communicate without text?}
\end{tcolorbox}

\begin{table*}[t]
\caption{
Cross-model KV transfer. \swat{} is the first method for cross-family KV-cache transfer with no receiver prefill (i.e., the receiver does not read the text).
}
\label{tab:cross-model-kv-comparison}
\centering
\begingroup
\definecolor{checkgreen}{HTML}{287A46}
\definecolor{crossred}{HTML}{B64040}

\providecommand{\cmark}{}\renewcommand{\cmark}{\textcolor{checkgreen}{\ding{51}}}
\providecommand{\xmark}{}\renewcommand{\xmark}{\textcolor{crossred}{\ding{55}}}
\begin{tabular}{lcccc}
\toprule
\multirow{2}{*}{\textbf{Paper}}
& \multirow{2}{*}{\textbf{Date}}
& \textbf{Different}
& \textbf{Cross-}
& \textbf{Skip} \\
& & \textbf{Sizes} & \textbf{Family} & \textbf{Prefill} \\
\midrule
Closed Form \cite{kv_nvidia}
& Aug 2026 & \cmark & \xmark & \cmark \\
CacheBridge \cite{kv_cachebridge}
& Sep 2026 & \cmark & \xmark & \cmark \\
KV Alignment \cite{kv_google}
& Jan 2026 & \cmark & \xmark & \cmark \\
Dense Latent \cite{kv_dense_latent}
& Jun 2026 & \cmark & \xmark & \cmark \\
LCF \cite{kv_lcf}
& May 2026 & \cmark & \xmark & \xmark \\
LCF-X \cite{kv_lcf}
& May 2026 & \xmark & \xmark & \cmark \\
ICaRus \cite{kv_icarus}
& Mar 2026 & \xmark & \xmark & \cmark \\
Cache-to-Cache \cite{kv_c2c}
& Oct 2025 & \cmark & \cmark & \xmark \\
XKV \cite{kv_xkv}
& Aug 2026 & \cmark & \cmark & \xmark \\
\midrule
\textbf{\swat{} (ours)} & --- & \cmark & \cmark & \cmark \\
\bottomrule
\end{tabular}
\endgroup
\end{table*}

There is reason to believe the answer is yes. The Platonic Representation Hypothesis~\citep{prh}
argues that models trained on the same world converge toward the same representation and are becoming more aligned due to selective pressures. 
If so, then LLMs built by different organizations with different architectures and data still share what they represent, differing mainly in how they express it. A sender's internal representations would then already contain what a receiver needs, and cross-family communication reduces to learning a mapping function.

If cross-family communication can bypass receiver prefill, it would enable a multitude of optimizations in the field. In a multi-agent system, a child can hand its parent a compacted cache instead of a summary which saves compute on both sides and carries more information. When switching models mid-conversation, the incoming model can inherit the outgoing model's cache instead of re-prefilling the entire context, which skips the cost of switching models mid-context. In routing systems that dispatch each query to the cheapest capable model~\citep{mas_route1}, a router that has already read the query can pass its cache to the chosen model rather than forcing a second read. In general, throughout LLM pipelines there are many moments when a cache for family A's model exists while family B's model is needed for decoding, and our method \swat{} (\cref{fig:intro}) enables these capabilities. Our contributions are:
\begin{itemize}
    \item We introduce \swat{}, a cross-family KV-cache transfer method that allows communication between LLMs without the receiver prefilling any text.
    \item Across many model--pairs, families, and sizes, an evaluation across 17 benchmarks shows that \swat{} is capable of transferring caches with high performance while exceeding the receiver model's own capabilities in certain setups.
    \item We demonstrate that communication in a multi-agent system is more effective when combining \swat{} with KV-cache compaction techniques such as \am{} \cite{yoon_attentionmatching} compared to textual summaries between agents. This is evaluated across many datasets, budgets, and sender--receiver pairs.
    \item We demonstrate that despite \swat{} only training on English text, it gives the receiver model zero-shot capabilities from the sender such as vision and Chinese capabilities.
\end{itemize}

\section{Background}
\label{sec:background}

% V00B: v00 with abstract depth: B blocks, slices S_b of L_S sender layers, groups R_b of L_R receiver layers
% Architecture source: multiagent/model/cache_renderer.py, CacheRendererV2.
% Notation synchronized with live/live.tex: zero-indexed blocks; trainable modules underlined.
\begin{figure*}
\centering\sffamily
\definecolor{snd}{RGB}{0,114,178}\definecolor{rcv}{RGB}{213,94,0}\definecolor{swt}{RGB}{0,158,115}
\definecolor{gry}{RGB}{140,140,140}\definecolor{bad}{RGB}{190,30,45}\definecolor{ylw}{RGB}{230,159,0}\definecolor{sky}{RGB}{86,180,233}
\tikzset{>={Stealth[length=4pt]}, font=\footnotesize,
  flow/.style={->, thick, black!55},
  xat/.style={->, line width=0.9pt, snd!80!black},
  kvo/.style={->, line width=0.9pt, rcv!85!black},
  tp/.style={-{Stealth[length=5pt,width=5pt]}, line width=1.4pt, swt!80!black},
  lab/.style={font=\scriptsize, text=black!65},
  ttl/.style={font=\footnotesize\bfseries},
  tag/.style={font=\scriptsize\itshape, text=black!55},
  sblk/.style={draw=swt!75!black, fill=swt!14, thick, rounded corners=2pt, align=center},
  agentA/.style={draw=snd!80!black, fill=snd!6, thick, rounded corners=4pt},
  agentB/.style={draw=rcv!80!black, fill=rcv!6, thick, rounded corners=4pt},
  swatbox/.style={draw=swt!80!black, fill=swt!5, thick, rounded corners=4pt},
  grp/.style={draw=#1!80!black, fill=#1!22, rounded corners=1.5pt, line width=0.5pt},
  tok/.style={draw=#1!70!black, fill=#1!12, rounded corners=1pt, inner sep=1.4pt, font=\scriptsize\sffamily, minimum height=0.36cm},
  op/.style={draw=black!45, fill=white, rounded corners=1.5pt, inner sep=1.6pt, font=\scriptsize, minimum height=0.36cm},
  pics/doc/.style={code={
    \draw[black!60, fill=white, line width=0.5pt] (-0.2,-0.26) -- (0.2,-0.26) -- (0.2,0.15) -- (0.09,0.26) -- (-0.2,0.26) -- cycle;
    \draw[black!60, line width=0.5pt] (0.09,0.26) -- (0.09,0.15) -- (0.2,0.15);
    \foreach \y in {-0.16,-0.07,0.02,0.11} \draw[black!40] (-0.13,\y) -- (0.13,\y);}},
  pics/botA/.style={code={
    \draw[#1!70!black, line width=0.6pt] (0,0.4) -- (0,0.55); \fill[#1!70!black] (0,0.6) circle (0.055);
    \draw[#1!70!black, line width=0.7pt, fill=#1!22, rounded corners=3pt] (-0.38,-0.36) rectangle (0.38,0.4);
    \fill[white] (-0.16,0.08) circle (0.1) (0.16,0.08) circle (0.1);
    \fill[#1!75!black] (-0.16,0.08) circle (0.055) (0.16,0.08) circle (0.055);
    \draw[#1!75!black, line width=0.7pt, line cap=round] (-0.13,-0.18) -- (0.13,-0.18);}},
  pics/botB/.style={code={
    \fill[#1!70!black, rounded corners=1pt] (-0.47,-0.1) rectangle (-0.35,0.12) (0.35,-0.1) rectangle (0.47,0.12);
    \draw[#1!70!black, line width=0.7pt, fill=#1!22] (0,0.01) circle (0.39);
    \fill[white] (-0.25,-0.01) rectangle (-0.06,0.17) (0.06,-0.01) rectangle (0.25,0.17);
    \fill[#1!75!black] (-0.195,0.04) rectangle (-0.115,0.12) (0.115,0.04) rectangle (0.195,0.12);
    \draw[#1!75!black, line width=0.7pt, line cap=round] (-0.12,-0.15) .. controls (-0.05,-0.23) and (0.05,-0.23) .. (0.12,-0.15);}},
  say/.style={draw=black!50, fill=white, rounded corners=4pt, inner sep=2.5pt, font=\scriptsize\itshape, align=left},
  pics/flake/.style={code={\foreach \a in {0,60,120}{\draw[#1, line width=0.6pt] (\a:0.12) -- (\a+180:0.12);
    \foreach \s in {1,-1}\foreach \b in {\a,\a+180}\draw[#1, line width=0.45pt] (\b:0.07) -- ++(\b+\s*45:0.045);}}},
  pics/flame/.style={code={\begin{scope}[yshift=-0.15cm]
    \fill[orange!85!red] (0,0) .. controls (-0.13,0) and (-0.14,0.13) .. (-0.06,0.19) .. controls (-0.04,0.22) and (-0.04,0.26) .. (-0.01,0.31)
      .. controls (0.03,0.24) and (0.11,0.2) .. (0.11,0.1) .. controls (0.11,0.03) and (0.06,0) .. (0,0) -- cycle;
    \fill[yellow!80!orange] (0,0.015) .. controls (-0.07,0.015) and (-0.07,0.09) .. (-0.02,0.14) .. controls (0,0.12) and (0.06,0.1) .. (0.06,0.06)
      .. controls (0.06,0.03) and (0.03,0.015) .. (0,0.015) -- cycle;\end{scope}}},
  pics/kvgrid/.style={code={\foreach \x in {0,...,5}\foreach \y in {0,1}{\pgfmathsetmacro\t{25+50*mod(3*\x+5*\y+1,7)/6}
    \fill[#1!\t] (\x*0.11,\y*0.11) rectangle ++(0.095,0.095);}}},
}
\newcommand{\qwen}{\raisebox{-0.35ex}{\includegraphics[height=1.6ex,draft=false]{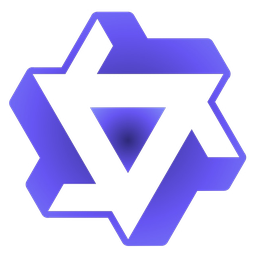}}}
\newcommand{\llama}{\raisebox{-0.35ex}{\includegraphics[height=1.6ex,draft=false]{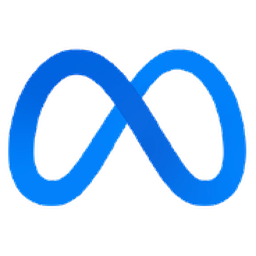}}}
\begin{tikzpicture}
% local text sizes: nothing below \scriptsize
\tikzset{lab/.style={font=\footnotesize, text=black!65}, tag/.style={font=\footnotesize\itshape, text=black!55},
  op/.style={draw=black!45, fill=white, rounded corners=1.5pt, inner sep=1.8pt, font=\footnotesize, minimum height=0.4cm},
  inp/.style={draw=black!55, fill=white, rounded corners=1pt, inner sep=1.6pt, font=\scriptsize\sffamily, minimum height=0.4cm}}
% left: three towers with abstract depth (rows 0, ..., b, ..., B)
\node[agentA, minimum width=1.32cm, minimum height=3.4cm, anchor=south] (A) at (0.66,-0.1) {};
\node[swatbox, minimum width=1.0cm, minimum height=1.95cm] (S) at (3.5,1.55) {};
\node[agentB, minimum width=1.32cm, minimum height=3.4cm, anchor=south] (B) at (6.34,-0.1) {};
\node[ttl, text=snd!70!black, align=center, anchor=south] at (A.north) {Sender Agent ($\mathcal{S}$)\\[-1pt]{\footnotesize\mdseries $L_{\mathcal{S}}$ layers}};
\node[ttl, text=swt!45!black, align=center, anchor=south] at ($(S.north)+(0,0.05)$) {\swat{}\\[-1pt]{\footnotesize\mdseries $\approx$300M params}\\[-1pt]{\footnotesize\mdseries $<$1 GPU-day}\\[-1pt]{\footnotesize\mdseries $<$50M tokens}};
\node[ttl, text=rcv!70!black, align=center, anchor=south] at (B.north) {Receiver Agent ($\mathcal{R}$)\\[-1pt]{\footnotesize\mdseries $L_{\mathcal{R}}$ layers}};
% real rows r = 0 (0), 2 (b), 4 (B); stripe counts differ by at most one (floor/ceil of L/B)
\foreach \r/\lab/\ns/\nr in {0/0/4/4, 2/b/5/4, 4/B/4/5}{
  \pgfmathsetmacro\y{0.1+0.6*\r}
  \node[grp=snd, minimum width=0.72cm, minimum height=0.38cm, anchor=south west] (a\r) at (0.1,\y) {};
  \foreach \i in {1,...,\ns} \draw[snd!70!black, line width=0.3pt] ($(a\r.south west)+(0.07,{0.38*\i/(\ns+1)})$) -- ++(0.58,0);
  \node[font=\footnotesize, text=snd!70!black, anchor=west, inner sep=1pt] at ($(a\r.east)+(0.03,0)$) {$\mathcal{S}_{\lab}$};
  \node[sblk, minimum width=0.7cm, minimum height=0.27cm, font=\footnotesize, inner sep=0pt, line width=0.6pt] (s\r) at (3.5,{1.49+0.31*(\r-2)}) {$\lab$};
  \node[grp=rcv, minimum width=0.72cm, minimum height=0.38cm, anchor=south east] (b\r) at (6.9,\y) {};
  \foreach \i in {1,...,\nr} \draw[rcv!70!black, line width=0.3pt] ($(b\r.south west)+(0.07,{0.38*\i/(\nr+1)})$) -- ++(0.58,0);
  \node[font=\footnotesize, text=rcv!70!black, anchor=east, inner sep=1pt] at ($(b\r.west)+(-0.03,0)$) {$\mathcal{R}_{\lab}$};
  \draw[xat, line width=0.7pt] (1.32,\y+0.19) -- (S.west |- s\r.west);
  \draw[kvo, line width=0.7pt] (S.east |- s\r.east) -- (5.68,\y+0.19);}
% ellipsis rows r = 1, 3
\foreach \r in {1,3}{
  \pgfmathsetmacro\y{0.29+0.6*\r}
  \foreach \d in {-0.1,0,0.1} {\fill[snd!70!black] (0.46,\y+\d) circle (0.016); \fill[rcv!70!black] (6.54,\y+\d) circle (0.016);}
  \foreach \d in {-0.07,0,0.07} \fill[swt!45!black] (3.5,{1.49+0.31*(\r-2)+\d}) circle (0.012);}
\draw[swt!40!black, line width=1.2pt, rounded corners=2pt] (s2.south west) rectangle (s2.north east);
\pic[scale=1.2] at ($(A.north east)+(-0.2,-0.24)$) {flake=sky!80!black};
\pic[scale=1.2] at ($(B.north east)+(-0.2,-0.24)$) {flake=sky!80!black};
\pic[scale=0.6] at ($(S.north east)+(-0.15,-0.17)$) {flame};
% inputs
\pic at (-0.3,-0.78) {doc}; \node[lab, anchor=base, inner sep=0pt] at (-0.3,-0.4) {$d$};
\node[inp] (qt) at (0.66,-0.78) {$\mathcal{S}$'s tokens};
\draw[flow] (-0.05,-0.78) -- (qt); \draw[flow] (qt.north) -- (0.66,-0.1);
\node[inp] (lt) at (3.5,-0.78) {$E_{\mathcal R}(\mathcal T_{\mathcal R})$};
\draw[flow] (-0.3,-1.08) |- (3.5,-1.25) -- (lt.south);
\draw[->, swt!60!black, thick] (lt.north) -- node[right, lab] {$\underline W_{\mathrm{in}}$} (S.south);
\node[inp] (qq) at (6.34,-0.78) {$q$ / $\emptyset$};
\draw[flow] (qq.north) -- (6.34,-0.1);
% zoom panel, shifted down as a whole
\begin{scope}[yshift=-0.3cm]
% zoom frame
\draw[swatbox, densely dashed] (7.65,-1.25) rectangle (17.15,3.92);
% zoom funnel: leaves block b nearly flat, then flares out towards the panel corners
\def\funtop{($(s2.east)+(0,0.1)$) .. controls +(0.9,0.3) and +(-0.9,-1.4) .. (7.65,3.92)}
\def\funbot{($(s2.east)+(0,-0.1)$) .. controls +(0.9,-0.3) and +(-0.9,1.8) .. (7.65,-1.25)}
\begin{scope}[on background layer]\fill[swt!6] ($(s2.east)+(0,0.1)$) .. controls +(0.9,0.3) and +(-0.9,-1.4) .. (7.65,3.92)
    -- (7.65,-1.25) .. controls +(-0.9,1.8) and +(0.9,-0.3) .. ($(s2.east)+(0,-0.1)$) -- cycle;
  \draw[densely dashed, swt!60!black, draw opacity=0.4] \funtop \funbot;\end{scope}
\node[ttl, text=swt!45!black, anchor=north west] at (7.75,3.87) {block $b$};
% memory branch
\node[lab, text=snd!70!black] at (8.95,-0.6) {$\mathcal{S}$ layers $\ell\in\mathcal{S}_b$};
\node[inner sep=1pt] (kg) at (8.35,-0.25) {\tikz\pic{kvgrid=snd};};
\node[inner sep=1pt] (vg) at (9.55,-0.25) {\tikz\pic{kvgrid=snd};};
\node[lab, text=snd!70!black, anchor=east, inner sep=0.5pt] at ($(kg.west)+(0.02,0)$) {K};
\node[lab, text=snd!70!black, anchor=west, inner sep=0.5pt] at ($(vg.east)+(-0.02,0)$) {V};
\node[op] (der) at (8.35,0.33) {de-RoPE};
\node[circle, draw=snd!70!black, fill=white, inner sep=0pt, minimum size=0.3cm, font=\scriptsize] (cat) at (8.95,0.72) {$\Vert$};
\node[op] (mli) at (8.95,1.3) {$\mathrm{LN}\to\underline W_b^{\mathrm{mem}}$};
\draw[xat] (kg) -- (der); \draw[xat] (der) |- (cat.west); \draw[xat] (vg) |- (cat.east); \draw[xat] (cat) -- (mli);
\node[tag, align=center] at (8.9,2.6) {cross-tokenizer-aware\\RoPE and\\causal masking};
% residual stream
\node[lab] (hin) at (12.0,-0.47) {$h_b$};
\node[sblk, minimum width=3.3cm, inner ysep=1.5pt, font=\footnotesize] (sa) at (12.0,0.2) {$\underline{\mathrm{LN}}$ $\to$ \underline{causal self-attn} (RoPE)};
\node[sblk, minimum width=3.3cm, inner ysep=1.5pt, font=\footnotesize] (xa) at (12.0,1.3) {$\underline{\mathrm{LN}}$ $\to$ \underline{causal cross-attn} (RoPE)};
\node[sblk, minimum width=3.3cm, inner ysep=1.5pt, font=\footnotesize] (ff) at (12.0,2.4) {$\underline{\mathrm{LN}}$ $\to$ \underline{FFN} (GELU)};
\foreach \n/\y in {p1/0.67, p2/1.77, p3/2.87} \node[circle, draw=swt!60!black, fill=white, inner sep=0pt, minimum size=0.23cm, font=\scriptsize] (\n) at (12.0,\y) {$+$};
\node[lab] (hout) at (12.0,3.64) {$h_{b+1}$};
\foreach \a/\b in {hin/sa, p1/xa, p2/ff, p3/hout} \draw[->, swt!60!black, thick] (\a) -- (\b);
\foreach \a/\b in {sa/p1, xa/p2, ff/p3} \draw[swt!60!black, thick] (\a) -- (\b);
\foreach \s/\p in {-0.27/p1, 0.88/p2, 1.98/p3} {\draw[->, swt!50!black] (12.0,\s) -- ++(1.85,0) |- (\p.east); \fill[swt!60!black] (12.0,\s) circle (0.035);}
\draw[xat] (mli) -- node[above, lab, text=snd!70!black] {$m_b$} (xa.west);
% output head
\coordinate (tap) at (12.0,3.12);
\node[op] (oln) at (15.6,2.7) {$\underline{\mathrm{LN}}\to\underline W_b^{\mathrm{out}}$};
\node[op] (bnd) at (15.6,1.95) {RMS calibration};
\node[align=center, lab, text=rcv!70!black] (kv) at (15.6,0.95) {\tikz\pic{kvgrid=rcv};\\K, V of $\mathcal{R}$ layers\\$\ell\in\mathcal{R}_b$};
\node[op] (rrope) at (15.6,-0.2) {receiver RoPE (K)};
\node[agentB, font=\footnotesize, inner sep=2pt] (bb) at (15.6,-0.9) {into $\mathcal{R}$};
\draw[kvo] (tap) -| (oln.north); \fill[swt!60!black] (tap) circle (0.045); \draw[kvo] (oln) -- (bnd); \draw[kvo] (bnd) -- (kv); \draw[kvo] (kv) -- (rrope); \draw[kvo] (rrope) -- (bb);
\end{scope}
\end{tikzpicture}
\caption{
\swat{} between a frozen sender agent ($\mathcal{S}$) and receiver agent ($\mathcal{R}$). Left: $\mathcal{S}$ prefills the document $d$ once. \swat{} has $B$ blocks where block $b$ cross-attends to $\mathcal{S}$'s KV-cache for the layer group $\mathcal{S}_b$ and outputs $\mathcal{R}$'s KV-cache for the layer group $\mathcal{R}_b$. Only \swat{} is trained where \underline{underlining} marks learned parameters or modules. Right: block $b$ maps $h_b$ to $h_{b+1}$ through causal self-attention on $h_b$, then causal cross-attention on a linear projection of $\mathcal{S}_b$ (RoPE removed), then an FFN. Positions and the causal mask in the cross-attention module are aligned across the tokenizers as described in \cref{subsec:method_cross_fam_token}. The output of $b$ is linearly projected to $\mathcal{R}_b$ with receiver RoPE applied to the keys.
}
\Description{Left: three towers drawn with abstract depth. Blue Sender Agent (S) with L_S layers shows groups S_0, S_b and S_B separated by vertical dots; a small green SwaT box with B blocks and about 300M parameters, trained in under 6 GPU-hours on under 50M tokens, marked with a flame, shows blocks 0, b and B with b highlighted; orange Receiver Agent (R) with L_R layers shows groups R_0, R_b and R_B. Snowflakes mark both agents as frozen. Arrows connect each sender group to its block and each block to its receiver group. The context enters S as S's tokens and SwaT as R's embeddings; R reads an optional query or nothing. Dashed lines zoom from block b into a green panel on the right: K and V of the sender layers in S_b, K de-RoPE'd, are concatenated and pass through a LayerNorm-plus-linear block into the causal cross-attention of a residual stream with causal self-attention, cross-attention and FFN; The residual stream goes from h_b to h_{b+1}. Underlining identifies learned modules; memory normalization has no learned parameters. An output head of LayerNorm, linear, RMS calibration and receiver RoPE writes the keys and values of the receiver layers in R_b into Agent R.}
\end{figure*}
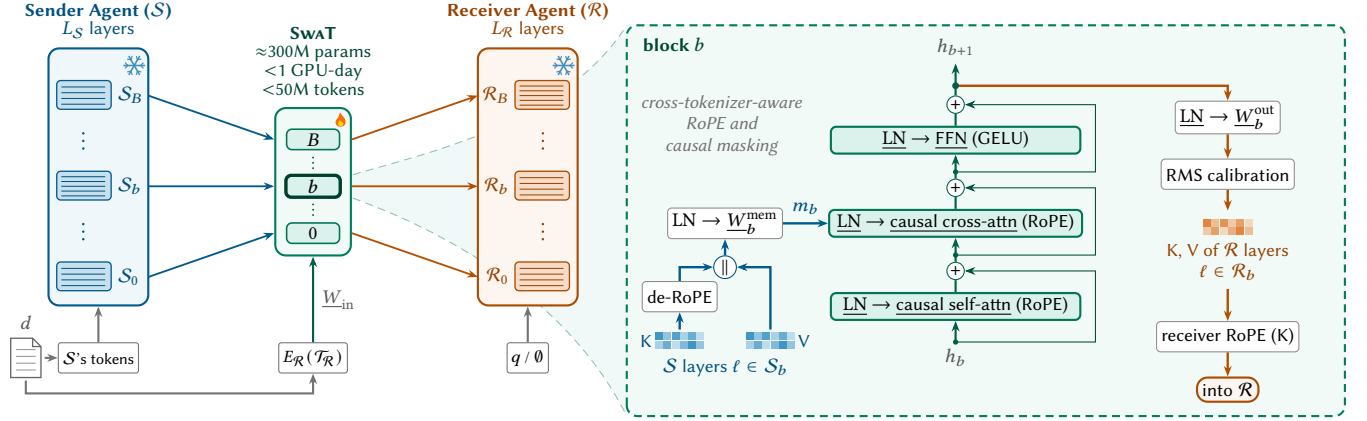

Consider a multi-agent system with $n$ LLM agents $\{\mathcal{A}^{(1)},\ldots,\mathcal{A}^{(n)}\}$. Each agent processes inputs in two stages: \emph{prefill}, which reads an input sequence and computes the agent's internal representations (key-value cache or KV-cache), and \emph{decode}, which uses the KV-cache to generate output tokens autoregressively.

For agent $\mathcal{A}^{(i)}$ and input $x$, the two stages are:
\begin{equation}
    C_x^{(i)} = \mathcal{A}_{\mathrm{prefill}}^{(i)}(x),
    \qquad
    y^{(i)} \sim \mathcal{A}_{\mathrm{decode}}^{(i)}(C_x^{(i)}).
    \label{eq:prefill-decode}
\end{equation}
LLMs by default accumulate their cache to prevent the quadratic cost per generated token ($\mathcal{O}(\vert T_{in}\vert^2*T_{out})$ where $T_{in}$ denotes the input tokens and $T_{out}$ is the number of generated tokens). Thus, as the generated tokens are processed, the cache extends to $C_{x\Vert y^{(i)}}^{(i)}$, where $\Vert$ denotes sequence concatenation. In text-based communication, $\mathcal{A}^{(i)}$ sends $x\Vert y^{(i)}$ to another agent $\mathcal{A}^{(j)}$, which reads it through its own prefill:
\begin{equation}
    C_{x\Vert y^{(i)}}^{(j)} = \mathcal{A}_{\mathrm{prefill}}^{(j)}(x\Vert y^{(i)}),
    \qquad
    y^{(j)} \sim \mathcal{A}_{\mathrm{decode}}^{(j)}(C_{x\Vert y^{(i)}}^{(j)}).
\label{eq:text-communication}
\end{equation}
The information available to $\mathcal{A}^{(j)}$ is therefore mediated by the message $x\Vert y^{(i)}$ that $\mathcal{A}^{(i)}$ produces. We refer to $\mathcal{A}^{(i)}$ as the sender, and $\mathcal{A}^{(j)}$ as the receiver.

Instead of the above, we seek to communicate without agent $\mathcal{A}^{(j)}$ receiving text by using agent $\mathcal{A}^{(i)}$'s cache 
$C^{(i)}$.
Concretely, for any sender--receiver pair, we seek to learn
a mapper $M_{i\rightarrow j}$ that transforms
$C_{x}^{(i)}$ into $\widehat{C}_{x}^{(j)}$ which approximates the receiver’s own prefill cache while preserving (and sometimes exceeding) downstream performance:
\begin{equation}
    \widehat{C}_{x}^{(j)}
    =
    M_{i\rightarrow j}\bigl(C_{x}^{(i)}\bigr),
    \qquad
    y^{(j)} \sim
    \mathcal{A}_{\mathrm{decode}}^{(j)}
    \bigl(\widehat{C}_{x}^{(j)}, q\bigr),
\label{eq:cache-transfer}
\end{equation}
where $q$ is the receiver's query and $y^{(j)}$ is its answer. Notice how $\mathcal{A}^{(j)}$ was not given the raw input $x$.

During decoding (such as in \cref{eq:prefill-decode}), an LLM accesses its previously processed context only through its KV-cache, rather than the raw input. The cache therefore serves as the agent's accumulated internal state, conditioning its subsequent actions. Transferring this cache allows another agent to continue from that state. We call this direct transfer of internal state \emph{telepathy} within the \emph{swarm}, giving our method its name: \swat{} (\underline{Swa}rm \underline{T}elepathy).

\subsection{Summarization vs Attention Matching}
\label{subsec:method-t2t-vs-am}

In the preceding formulation, agent $\mathcal{A}^{(i)}$ passes the full sequence $x\Vert y$ to $\mathcal{A}^{(j)}$. In practical multi-agent systems, this sequence contains lengthy intermediate outputs which are unnecessary for the receiver's task. This can be computationally prohibitive and could degrade performance by introducing context pollution \cite{context_pollution_1,context_pollution_2}. An effective solution is to communicate through summarization.

\textbf{Text-to-Text communication (\ttot{})} involves the following steps:

    \textbf{Sender task.}
    Agent $\mathcal{A}^{(i)}$ processes input $x$ and achieves its task with
    generated output $y$:
    \begin{equation}
        C_x^{(i)} = \mathcal{A}_{\mathrm{prefill}}^{(i)}(x),
        \qquad
        y \sim \mathcal{A}_{\mathrm{decode}}^{(i)}(C_x^{(i)}).
    \end{equation}
    At the end of generation, the cache becomes $C_{x\Vert y}^{(i)}$. In our experiments, the document $d$ from the dataset is $x\Vert y$.

    \textbf{Summarization.}
    The sender generates a summary $s$ of its findings,
    conditioned on $C_{x\Vert y}^{(i)}$ and a
    summarization instruction $p_{\mathrm{sum}}$:
    \begin{equation}
        s \sim \mathcal{A}_{\mathrm{decode}}^{(i)}
        \bigl(C_{x\Vert y}^{(i)}, p_{\mathrm{sum}}\bigr).
    \label{eq:t2t-summary}
    \end{equation}

    \textbf{Receiver prefill.}
    $\mathcal{A}^{(j)}$ reads the summary to construct
    its cache:
    \begin{equation}
        C_s^{(j)}
        = \mathcal{A}_{\mathrm{prefill}}^{(j)}(s).
    \label{eq:t2t-rprefil}
    \end{equation}

    \textbf{Receiver task.}
    Given a subsequent input $x^{(j)}$, the receiver achieves
    its task using the prefilled summary:
    \begin{equation}
        y^{(j)} \sim \mathcal{A}_{\mathrm{decode}}^{(j)}
        \bigl(C_s^{(j)}, x^{(j)}\bigr).
    \end{equation}

Text-to-Text communication significantly reduces the amount of context the receiver must process, but introduces an additional autoregressive generation step at the sender. The information conveyed from $x\Vert y$ is also limited to what the sender includes in $s$.

\textbf{Attention Matching (\am{})}~\citep{yoon_attentionmatching}
is an effective KV-cache compaction method that constructs a smaller set of keys and values yet preserves the influence of the original KV-cache. It introduces a hyperparameter $U$ which is the retained cache budget that controls the trade-off between cache size and fidelity.

KV-cache compaction is a natural fit after \swat{}: the sender already holds the accumulated cache, and \swat{} translates it into a representation the receiver can use directly. Compacting the representation produces a message without generating text or requiring the receiver to prefill text, eliminating steps in \cref{eq:t2t-summary,eq:t2t-rprefil}. Specifically, we combine \swat{} with \am{} (\swat{}+\am{}) by translating the sender's cache and compacting it ($\widetilde{C}^{(j)}_{U}$). Then the transferred and compacted cache ($\widetilde{C}^{(j)}_{U}$) is given to the receiver to begin decoding:
\begin{equation}
    \widetilde{C}^{(j)}_{U}
    =
    \mathrm{AM}_{U}\bigl(
        M_{i\rightarrow j}(C_{x\Vert y}^{(i)})
    \bigr),
        \qquad
    y^{(j)} \sim \mathcal{A}_{\mathrm{decode}}^{(j)}
    \bigl(\widetilde{C}^{(j)}_{U}, x^{(j)}\bigr).
\label{eq:swat-am}
\end{equation}

In \ttot{}, the realized summary length (i.e. the number of tokens generated by $\mathcal{A}_{\mathrm{decode}}^{(i)} \bigl(C_{x\Vert y}^{(i)}, p_{\mathrm{sum}}\bigr)$) depends on the sender $\mathcal{A}^{(i)}$, its accumulated context $C_{x\Vert y}^{(i)}$, and the summary prompt $p_{\mathrm{sum}}$. For a comparison at matched budgets, we choose $U$ to match the length of the summary $s$ generated in \cref{eq:t2t-summary}, measured in receiver tokens. This compares \ttot{} and \swat{}+\am{} at comparable cache budgets.

\begin{table*}[t]
\caption{
Zero-shot retention (\%, $\text{score}_\text{\swat{}} / \text{score}_\text{receiver}$) across 17 benchmarks.
}
\label{tab:exp1-retention}
\centering
\begin{tabular}{r@{\hspace{0.5\tabcolsep}}c@{\hspace{0.5\tabcolsep}}lrrrrrrr}
\toprule
\multicolumn{3}{c}{} & \multicolumn{3}{c}{In-domain (9)} & \multicolumn{3}{c}{OOD (8)} & \\
\cmidrule(lr){4-6}\cmidrule(lr){7-9}
\textbf{Sender} & & \textbf{Receiver} & \shortstack{Short \\ Acc} & \shortstack{Context \\ Acc} & \shortstack{Context \\ F1} & \shortstack{Short \\ Acc} & \shortstack{Context \\ Acc} & \shortstack{Context \\ F1} & \textbf{Mean} \\
\midrule
\textbf{Small} & \makebox[2ex]{}\;$\boldsymbol\rightarrow$\;\makebox[2ex]{} & \textbf{Large} & & & & & & & \\
Qwen 3 4B & \makebox[2ex]{\raisebox{-0.4ex}{\includegraphics[height=2ex,draft=false]{assets/logos/qwen.png}}}\;$\rightarrow$\;\makebox[2ex]{\raisebox{-0.4ex}{\includegraphics[height=2ex,draft=false]{assets/logos/llama.png}}} & Llama 3 8B & 100.6\% & 100.2\% & 110.9\% & 88.7\% & 91.6\% & 98.7\% & \textbf{98.5\%} \\
Llama 3 8B & \makebox[2ex]{\raisebox{-0.4ex}{\includegraphics[height=2ex,draft=false]{assets/logos/llama.png}}}\;$\rightarrow$\;\makebox[2ex]{\raisebox{-0.4ex}{\includegraphics[height=2ex,draft=false]{assets/logos/qwen.png}}} & Qwen 3 14B & 89.6\% & 98.2\% & 116.1\% & 86.5\% & 98.9\% & 103.5\% & \textbf{97.5\%} \\
Ministral 3 8B & \makebox[2ex]{\raisebox{-0.4ex}{\includegraphics[height=2ex,draft=false]{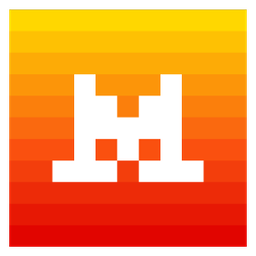}}}\;$\rightarrow$\;\makebox[2ex]{\raisebox{-0.4ex}{\includegraphics[height=2ex,draft=false]{assets/logos/llama.png}}} & Llama 3.2 11B & 104.2\% & 97.5\% & 108.5\% & 89.7\% & 92.8\% & 99.0\% & \textbf{99.4\%} \\
Gemma 3 4B & \makebox[2ex]{\raisebox{-0.4ex}{\includegraphics[height=2ex,draft=false]{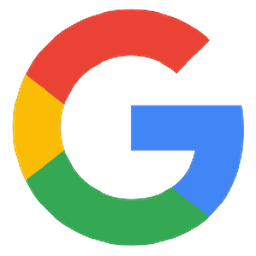}}}\;$\rightarrow$\;\makebox[2ex]{\raisebox{-0.4ex}{\includegraphics[height=2ex,draft=false]{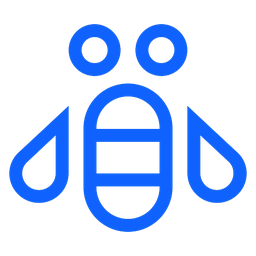}}} & Granite 3.3 8B & 95.4\% & 93.6\% & 159.8\% & 82.0\% & 95.0\% & 131.4\% & \textbf{108.5\%} \\
Qwen 3 4B & \makebox[2ex]{\raisebox{-0.4ex}{\includegraphics[height=2ex,draft=false]{assets/logos/qwen.png}}}\;$\rightarrow$\;\makebox[2ex]{\raisebox{-0.4ex}{\includegraphics[height=2ex,draft=false]{assets/logos/gemma.png}}} & Gemma 2 9B & 94.0\% & 92.4\% & 104.2\% & 83.4\% & 94.5\% & 96.3\% & \textbf{94.2\%} \\
Gemma 2 9B & \makebox[2ex]{\raisebox{-0.4ex}{\includegraphics[height=2ex,draft=false]{assets/logos/gemma.png}}}\;$\rightarrow$\;\makebox[2ex]{\raisebox{-0.4ex}{\includegraphics[height=2ex,draft=false]{assets/logos/ministral.png}}} & Ministral 3 14B & 101.9\% & 95.4\% & 110.0\% & 104.2\% & 95.3\% & 106.9\% & \textbf{102.8\%} \\
Gemma 3 4B & \makebox[2ex]{\raisebox{-0.4ex}{\includegraphics[height=2ex,draft=false]{assets/logos/gemma.png}}}\;$\rightarrow$\;\makebox[2ex]{\raisebox{-0.4ex}{\includegraphics[height=2ex,draft=false]{assets/logos/llama.png}}} & Llama 3 8B & 90.2\% & 101.6\% & 106.9\% & 73.3\% & 91.9\% & 99.0\% & \textbf{92.2\%} \\
\addlinespace
\textbf{Large} & \makebox[2ex]{}\;$\boldsymbol\rightarrow$\;\makebox[2ex]{} & \textbf{Small} & & & & & & & \\
Gemma 3 12B & \makebox[2ex]{\raisebox{-0.4ex}{\includegraphics[height=2ex,draft=false]{assets/logos/gemma.png}}}\;$\rightarrow$\;\makebox[2ex]{\raisebox{-0.4ex}{\includegraphics[height=2ex,draft=false]{assets/logos/qwen.png}}} & Qwen 3 4B & 104.0\% & 100.1\% & 120.5\% & 97.1\% & 97.7\% & 105.8\% & \textbf{104.6\%} \\
Qwen 3 32B & \makebox[2ex]{\raisebox{-0.4ex}{\includegraphics[height=2ex,draft=false]{assets/logos/qwen.png}}}\;$\rightarrow$\;\makebox[2ex]{\raisebox{-0.4ex}{\includegraphics[height=2ex,draft=false]{assets/logos/llama.png}}} & Llama 3 8B & 112.9\% & 104.0\% & 98.4\% & 102.5\% & 95.4\% & 97.1\% & \textbf{103.0\%} \\
Granite 3.3 8B & \makebox[2ex]{\raisebox{-0.4ex}{\includegraphics[height=2ex,draft=false]{assets/logos/granite.png}}}\;$\rightarrow$\;\makebox[2ex]{\raisebox{-0.4ex}{\includegraphics[height=2ex,draft=false]{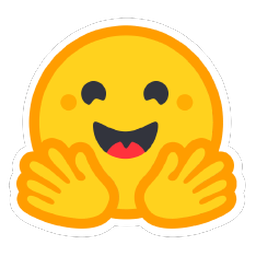}}} & SmolLM3 3B & 112.4\% & 103.9\% & 253.6\% & 113.4\% & 114.7\% & 162.1\% & \textbf{143.2\%} \\
\bottomrule
\end{tabular}
\end{table*}

\section{Method}
\label{sec:method}

For a sender--receiver pair, we train \swat{}, which is a transformer mapper $\underline M_{i\rightarrow j}$, keeping both LLMs frozen. We underline trainable parameters or modules. \swat{}'s architecture is controlled by hyperparameters block count $B$, hidden width $D$, attention-head count $H$, and feed-forward width $F$.

For agent $\mathcal U\in\{\mathcal S,\mathcal R\}$, let $L_{\mathcal U}$, $H_{\mathcal U}$, and $\delta_{\mathcal U}$ denote its KV-layer count, KV-head count, and head dimension; let $D_{\mathcal R}$ be the receiver embedding dimension. These dimensions are fixed by the frozen agents. We choose $B\ll\min(L_{\mathcal S},L_{\mathcal R})$.

\paragraph{Layer groups.}
We partition each agent's KV layers into $B$ contiguous groups. For block $b\in\{0,\ldots,B-1\}$,
\begin{equation}
\mathcal U_b=\left\{\ell:
\left\lfloor\frac{bL_{\mathcal U}}{B}\right\rfloor
\leq\ell<
\left\lfloor\frac{(b+1)L_{\mathcal U}}{B}\right\rfloor
\right\},
\quad \mathcal U\in\{\mathcal S,\mathcal R\}.
\end{equation}
Here, $|\mathcal U_b|$ denotes the number of layers in group $\mathcal U_b$. Block $b$ reads cache from $\mathcal S_b$ and predicts cache for $\mathcal R_b$, allowing each receiver layer in its group to draw on multiple sender layers.

\paragraph{Mapper inputs.}
Let $S$ and $R$ be the sender and receiver token counts for document $d$, respectively, and represent token vectors as columns. We tokenize $d$ with the receiver's tokenizer, obtaining $\mathcal T_{\mathcal R}=(r_0,\ldots,r_{R-1})$, and initialize the hidden states as
$h_0=\underline W_{\mathrm{in}}E_{\mathcal R}(\mathcal T_{\mathcal R})$,
where $\underline W_{\mathrm{in}}\in\mathbb R^{D\times D_{\mathcal R}}$ acts token-wise and $h_0\in\mathbb R^{D\times R}$. The frozen embedding table lookup $E_{\mathcal R}(\mathcal T_{\mathcal R})$ requires 0 \textsc{FLOP}s. For block $b$, we remove sender RoPE from its cached keys and concatenate these keys and values across heads and layers in $\mathcal S_b$ at each sender token. Applying $\underline W_b^{\mathrm{mem}}\in
\mathbb R^{D\times(2|\mathcal S_b|H_{\mathcal S}\delta_{\mathcal S})}$
produces memory $m_b\in\mathbb R^{D\times S}$.

\paragraph{SwaT blocks.}
Block $b$ updates $h_b$ through causal self-attention, causal cross-attention to $m_b$, and a feed-forward network, producing $h_{b+1}$. Each sublayer has its own $\underline{\mathrm{LN}}$ and residual connection. Both attention sublayers learn projections
$\underline W_b^Q,\underline W_b^K,\underline W_b^V,\underline W_b^O\in\mathbb R^{D\times D}$,
with separate weights for self- and cross-attention and $H$ heads of dimension $D/H$. Cross-attention uses $m_b$ as its key/value memory.
The FFN uses $\underline W_b^{(1)}\in\mathbb R^{F\times D}$ and $\underline W_b^{(2)}\in\mathbb R^{D\times F}$ with GELU.
Self-attention uses integer receiver positions; cross-attention uses the character-based RoPE alignment and causal mask described in \cref{subsec:method_cross_fam_token}.

\paragraph{Cache outputs.}
Each block applies its own $\underline{\mathrm{LN}}$ and output projection
$\underline W_b^{\mathrm{out}}\in
\mathbb R^{(2|\mathcal R_b|H_{\mathcal R}\delta_{\mathcal R})\times D}$
to $h_{b+1}$, predicting keys and values for all layers in $\mathcal R_b$. Predicted keys and values are separately rescaled per layer and head to match fixed RMS statistics calibrated from receiver caches on training contexts. We then apply receiver RoPE to the predicted keys at integer receiver positions. Assembling all blocks' outputs yields $\widehat C_d^{(j)}$, with one cache entry per receiver token in every receiver KV layer. The receiver processes $q$ on top of this cache and generates its answer.

\paragraph{Sliding-window layers.}
Some models such as Gemma~2/3 utilize sliding-window layers that only attend to recent tokens and store the cache of only those. To handle a sliding-window layer $l_i$, if $l_i \in \mathcal S_b$ then we force $l_i$ to record all past KV-cache even outside of its window and make that an input to \swat{} ($l_i$'s attention equation is not changed). If $l_i \in \mathcal R_b$, we make \swat{} predict the KV-cache for all tokens in $l_i$, and afterwards $l_i$ acts normally and drops KV-cache for old tokens that it does not need.

\begin{figure*}[h]
  \centering
  \includegraphics[draft=false,width=\textwidth]{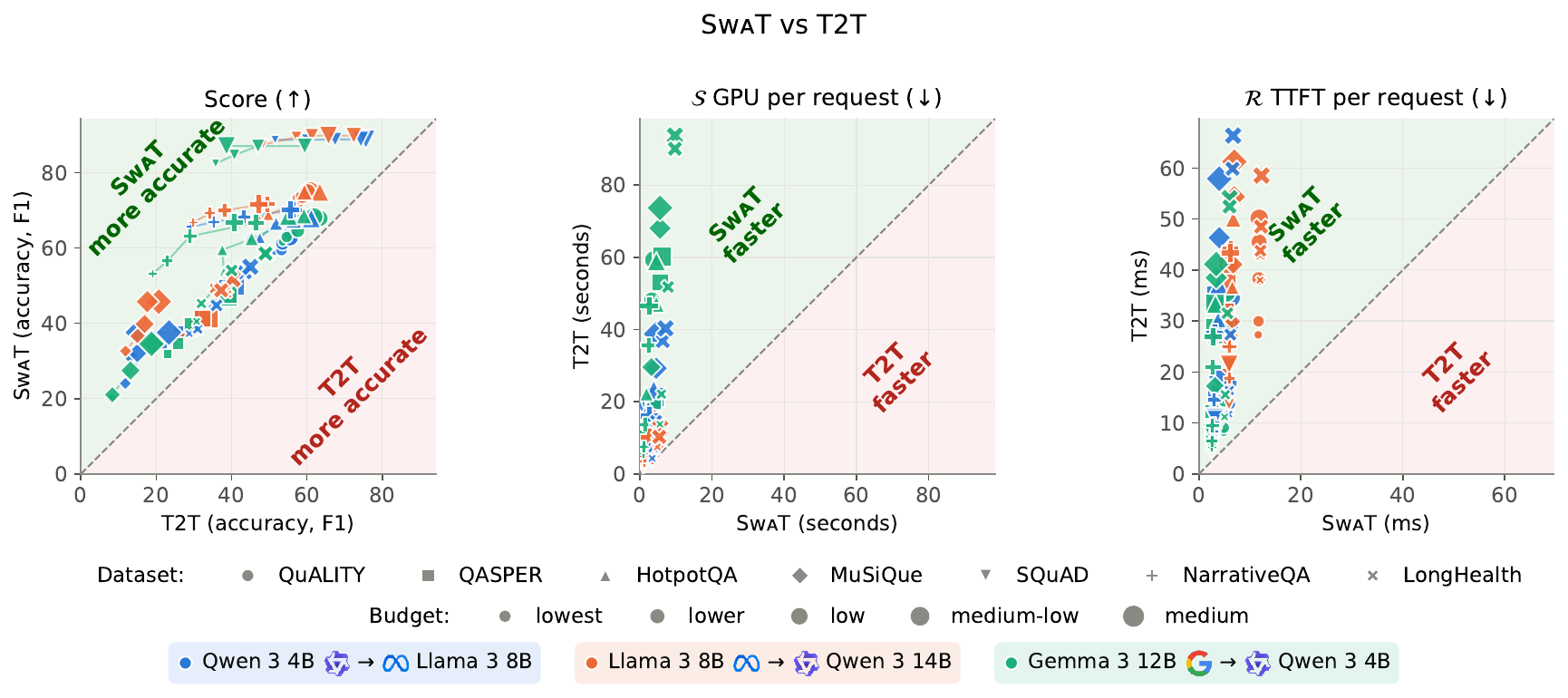}
  \caption{Comparing \swat{} (ours) with \am{} \cite{yoon_attentionmatching} against text-to-text communication (\ttot{}) across three model--pairs, five budgets, and seven datasets. \swat{}+\am{} exceeds \ttot{} in all metrics (task score, sender GPU time, receiver GPU TTFT).}
  \Description{Description}
  \label{fig:exp2_3scatters}
\end{figure*}

\subsection{Training Objectives}

We train \swat{} with two objectives, keeping both agents frozen.
Each training example pairs a document $d$ with a task prompt $q$
and a target continuation $y$, an answer or reasoning trace.
Let $C^{\mathcal S}_d$ and $C^{\mathcal R}_d$ denote the sender's and
receiver's prefill caches of $d$, so that \swat{} predicts the receiver
cache as $\underline M_{i\rightarrow j}(C^{\mathcal S}_d,\mathcal T_{\mathcal R})$.
The first objective is to reconstruct the receiver's native cache using
RMS-normalized mean squared error:
\begin{equation}
\mathcal L_{\mathrm{KV}}
=
\operatorname{MSE}\!\left(
\frac{\underline M_{i\rightarrow j}(C^{\mathcal S}_d,\mathcal T_{\mathcal R})}{\sigma},
\frac{C^{\mathcal R}_d}{\sigma}
\right),
\label{eq:loss-mse}
\end{equation}
where $\sigma$ is the RMS of $C^{\mathcal R}_d$ for
keys and values in each layer-head.
Keys are compared before RoPE.

The second objective is to optimize the transferred cache for downstream
generation using the next-token prediction loss
\begin{equation}
\mathcal L_{\mathrm{NLL}}
=
-\frac{1}{|y|}\sum_{t=1}^{|y|}
\log p_{\mathcal R}\!\left(
y_t\mid\underline M_{i\rightarrow j}(C^{\mathcal S}_d,\mathcal T_{\mathcal R}),q,y_{<t}
\right).
\label{eq:loss-ce}
\end{equation}
Gradients pass through the frozen receiver to update only
$\underline M_{i\rightarrow j}$.

\subsection{Cross-Family Tokenizers}
\label{subsec:method_cross_fam_token}

Our alignment strategy preserves both agents' native tokenizations. Sender and receiver tokens interact only through causal cross-attention, which bridges their representations despite differing sequence lengths and token boundaries. We align positional encodings and construct the causal attention mask by relying on the token character spans, which are defined over the shared characters.

Let document $d$ have $D$ characters $d=(c_1,\ldots,c_D)$ with sender tokens
$\mathcal{T}_{\mathcal S}=(s_0,\ldots,s_{S-1})$ and receiver tokens
$\mathcal{T}_{\mathcal R}=(r_0,\ldots,r_{R-1})$.
For each sender token $s_i$ and receiver token $r_j$ call their character spans $[a_i^{\mathcal S},b_i^{\mathcal S})$ and
$[a_j^{\mathcal R},b_j^{\mathcal R})$.

\paragraph{RoPE alignment.}
For sender token $s_i$, call its midpoint
$m_i=(a_i^{\mathcal S}+b_i^{\mathcal S})/2$, and
choose the largest $j$ with $a_j^{\mathcal R}\leq m_i$
(or $j=0$ if none exists). Then, its cross-attention RoPE position is
\begin{equation}
    p_i=\max\!\left(0,\,
    j+\operatorname{clip}_{[0,1]}\!\left(
    \frac{m_i-a_j^{\mathcal R}}
    {b_j^{\mathcal R}-a_j^{\mathcal R}}
    \right)-\frac12\right).
\label{eq:rope_alignment}
\end{equation}
Receiver tokens retain integer positions. Tokenizer templates are linearly interpolated. Notice how if both tokenizers are the same, $\mathcal{T}_\mathcal{S} = \mathcal{T}_\mathcal{R}$ and their character spans match, so \cref{eq:rope_alignment} reduces to $p_i = i$. This alignment strategy avoids prior token alignment methods from \citet{kv_c2c} where a mismatch between the tokenizers ends up dropping extra tokens causing the model to possibly lose important pieces of information.

\paragraph{Causal masking.}
Receiver token $r_j$ may attend to sender token $s_i$ only
if $b_i^{\mathcal S}\leq b_j^{\mathcal R}$, restricting
sender-cache access to the text prefix ending at $r_j$.

\section{Experiments}
\label{sec:experiments}

Our experiments address three questions:
\textbf{(1)} how much of the receiver's own performance survives when it reads
\swat{}'s transferred cache instead of the text?
We measure this across 17 datasets and 10 cross-family sender--receiver pairs
in both small-to-large and large-to-small
directions (\cref{subsec:exp-retention,tab:exp1-retention}).
\textbf{(2)} how does cache transfer compare with communicating through text?
We compare \swat{}, compressed with Attention Matching (AM), against \ttot{}, at matched communication budgets across 7 datasets, 3 cross-family sender--receiver pairs, and 5 different budgets (\cref{subsec:exp-t2t,fig:exp2_3scatters,tab:exp2-mean_se}).
\textbf{(3)} can the transferred cache carry capabilities the receiver lacks?
Although \swat{} is trained only on English text, we test whether it
transfers visual understanding from vision-language senders to text-only
receivers (\cref{subsec:exp-vision}), and whether it transfers specialized text knowledge and capabilities not seen during training (\cref{subsec:exp-specialized,apdx-sec:exp-long-context}).

\begin{table*}[t]
\caption{
Performance of \swat{} vs \ttot{} over three model--pairs, five budgets, and the mean over seven datasets (QuALITY, QASPER, HotpotQA, MuSiQue, SQuAD, NarrativeQA, LongHealth). All dataset samples are $(d, q)$: document $d$ is given to the sender agent ($\mathcal{S}$) and question $q$ to the receiver agent ($\mathcal{R}$). \ttot{}: $\mathcal{S}$ summarizes $d$, $\mathcal{R}$ prefills the summary along with $q$ and proceeds to answer. \swat{}+AM: the KV-cache of $d$ from $\mathcal{S}$ is transferred (\swat{}) and compacted (AM) then given to $\mathcal{R}$, which prefills only $q$ and proceeds to answer. Budgets are set by the \ttot{} summary from $\mathcal{S}$ under five different summary prompts, and \swat{}+AM is compressed to a similar budget. Metrics: task score (accuracy or F1), tokens communicated between the agents (Tokens), GPU time of $\mathcal{S}$ per request ($\mathcal{S}$ GPU), and time to first token of $\mathcal{R}$ per request ($\mathcal{R}$ TTFT), both in milliseconds. Change is reported $\pm$ bootstrap standard error. Green: \swat{} is statistically better than \ttot{} (95\% bootstrap CI).
}
\label{tab:exp2-mean_se}
\centering
\begingroup
\definecolor{expTwoSection}{HTML}{ECECEC}
\definecolor{expTwoBetter}{HTML}{D9EFD8}
\begin{tabular}{llrrrrrrrrr}
\toprule
 & & \multicolumn{3}{c}{Qwen3-4B $\rightarrow$ Llama3-8B} & \multicolumn{3}{c}{Llama3-8B $\rightarrow$ Qwen3-14B} & \multicolumn{3}{c}{Gemma3-12B $\rightarrow$ Qwen3-4B} \\
\cmidrule(lr){3-5}\cmidrule(lr){6-8}\cmidrule(lr){9-11}
Metric & Budget & \ttot{} & \swat{} & Change & \ttot{} & \swat{} & Change & \ttot{} & \swat{} & Change \\
\midrule
\rowcolor{expTwoSection}
 & \textit{Lowest} & \multicolumn{3}{c}{\textit{$\approx$16.2$\,\times$ compression}} & \multicolumn{3}{c}{\textit{$\approx$12.7$\,\times$}} & \multicolumn{3}{c}{\textit{$\approx$17.2$\,\times$}} \\
Score &  & 34.2 & \textbf{53.1} & \cellcolor{expTwoBetter}+18.9$\pm$4.6 & 37.2 & \textbf{58.6} & \cellcolor{expTwoBetter}+21.4$\pm$4.5 & 29.8 & \textbf{49.0} & \cellcolor{expTwoBetter}+19.3$\pm$5.5 \\
Tokens &  & 117 & \textbf{113} & $-$3.1$\pm$2.1\% & 161 & \textbf{149} & \cellcolor{expTwoBetter}$-$7.4$\pm$3.3\% & 114 & \textbf{110} & $-$3.8$\pm$2.2\% \\
$\mathcal{S}$ GPU (ms) &  & 3,608 & \textbf{1,527} & \cellcolor{expTwoBetter}$\times$2.4$\pm$0.4 & 3,675 & \textbf{1,897} & \cellcolor{expTwoBetter}$\times$1.9$\pm$0.2 & 7,329 & \textbf{2,094} & \cellcolor{expTwoBetter}$\times$3.5$\pm$0.6 \\
$\mathcal{R}$ TTFT (ms) &  & 9.2 & \textbf{3.8} & \cellcolor{expTwoBetter}$\times$2.4$\pm$0.2 & 21.8 & \textbf{7.3} & \cellcolor{expTwoBetter}$\times$3.0$\pm$0.3 & 6.7 & \textbf{3.2} & \cellcolor{expTwoBetter}$\times$2.1$\pm$0.1 \\
\addlinespace
\rowcolor{expTwoSection}
 & \textit{Lower} & \multicolumn{3}{c}{\textit{$\approx$11.0$\,\times$ compression}} & \multicolumn{3}{c}{\textit{$\approx$9.7$\,\times$}} & \multicolumn{3}{c}{\textit{$\approx$12.5$\,\times$}} \\
Score &  & 38.1 & \textbf{55.1} & \cellcolor{expTwoBetter}+17.0$\pm$3.7 & 39.9 & \textbf{60.7} & \cellcolor{expTwoBetter}+20.8$\pm$3.7 & 31.7 & \textbf{52.1} & \cellcolor{expTwoBetter}+20.4$\pm$5.0 \\
Tokens &  & 181 & \textbf{166} & \cellcolor{expTwoBetter}$-$8.2$\pm$4.2\% & 211 & \textbf{196} & \cellcolor{expTwoBetter}$-$7.2$\pm$3.6\% & 159 & \textbf{152} & $-$4.3$\pm$2.3\% \\
$\mathcal{S}$ GPU (ms) &  & 5,387 & \textbf{1,661} & \cellcolor{expTwoBetter}$\times$3.2$\pm$0.5 & 4,637 & \textbf{2,049} & \cellcolor{expTwoBetter}$\times$2.3$\pm$0.2 & 10,229 & \textbf{2,158} & \cellcolor{expTwoBetter}$\times$4.7$\pm$0.7 \\
$\mathcal{R}$ TTFT (ms) &  & 12.2 & \textbf{3.9} & \cellcolor{expTwoBetter}$\times$3.1$\pm$0.3 & 25.4 & \textbf{7.4} & \cellcolor{expTwoBetter}$\times$3.4$\pm$0.3 & 8.4 & \textbf{3.2} & \cellcolor{expTwoBetter}$\times$2.6$\pm$0.2 \\
\addlinespace
\rowcolor{expTwoSection}
 & \textit{Low} & \multicolumn{3}{c}{\textit{$\approx$7.5$\,\times$ compression}} & \multicolumn{3}{c}{\textit{$\approx$7.9$\,\times$}} & \multicolumn{3}{c}{\textit{$\approx$6.9$\,\times$}} \\
Score &  & 42.7 & \textbf{57.7} & \cellcolor{expTwoBetter}+14.9$\pm$2.6 & 42.9 & \textbf{61.8} & \cellcolor{expTwoBetter}+18.9$\pm$3.4 & 37.4 & \textbf{56.9} & \cellcolor{expTwoBetter}+19.6$\pm$4.6 \\
Tokens &  & 271 & \textbf{246} & \cellcolor{expTwoBetter}$-$9.1$\pm$3.2\% & 261 & \textbf{241} & \cellcolor{expTwoBetter}$-$7.9$\pm$2.7\% & 297 & \textbf{274} & \cellcolor{expTwoBetter}$-$7.7$\pm$2.8\% \\
$\mathcal{S}$ GPU (ms) &  & 7,826 & \textbf{1,909} & \cellcolor{expTwoBetter}$\times$4.1$\pm$0.6 & 6,214 & \textbf{2,265} & \cellcolor{expTwoBetter}$\times$2.7$\pm$0.3 & 19,402 & \textbf{2,664} & \cellcolor{expTwoBetter}$\times$7.3$\pm$1.0 \\
$\mathcal{R}$ TTFT (ms) &  & 16.1 & \textbf{4.0} & \cellcolor{expTwoBetter}$\times$4.1$\pm$0.4 & 31.6 & \textbf{7.4} & \cellcolor{expTwoBetter}$\times$4.2$\pm$0.5 & 13.2 & \textbf{3.4} & \cellcolor{expTwoBetter}$\times$3.9$\pm$0.5 \\
\addlinespace
\rowcolor{expTwoSection}
 & \textit{Medium-low} & \multicolumn{3}{c}{\textit{$\approx$4.1$\,\times$ compression}} & \multicolumn{3}{c}{\textit{$\approx$5.5$\,\times$}} & \multicolumn{3}{c}{\textit{$\approx$3.5$\,\times$}} \\
Score &  & 48.6 & \textbf{61.4} & \cellcolor{expTwoBetter}+12.8$\pm$2.1 & 46.9 & \textbf{63.9} & \cellcolor{expTwoBetter}+16.9$\pm$2.6 & 48.0 & \textbf{61.4} & \cellcolor{expTwoBetter}+13.4$\pm$3.1 \\
Tokens &  & 571 & \textbf{448} & \cellcolor{expTwoBetter}$-$21.5$\pm$5.5\% & 386 & \textbf{341} & \cellcolor{expTwoBetter}$-$11.6$\pm$6.9\% & 676 & \textbf{535} & \cellcolor{expTwoBetter}$-$20.8$\pm$6.0\% \\
$\mathcal{S}$ GPU (ms) &  & 17,170 & \textbf{2,912} & \cellcolor{expTwoBetter}$\times$5.9$\pm$0.5 & 8,873 & \textbf{2,874} & \cellcolor{expTwoBetter}$\times$3.1$\pm$0.3 & 42,526 & \textbf{4,019} & \cellcolor{expTwoBetter}$\times$10.6$\pm$1.1 \\
$\mathcal{R}$ TTFT (ms) &  & 30.3 & \textbf{4.1} & \cellcolor{expTwoBetter}$\times$7.4$\pm$1.1 & 41.5 & \textbf{7.6} & \cellcolor{expTwoBetter}$\times$5.5$\pm$0.7 & 25.9 & \textbf{3.6} & \cellcolor{expTwoBetter}$\times$7.2$\pm$1.1 \\
\addlinespace
\rowcolor{expTwoSection}
 & \textit{Medium} & \multicolumn{3}{c}{\textit{$\approx$3.6$\,\times$ compression}} & \multicolumn{3}{c}{\textit{$\approx$5.4$\,\times$}} & \multicolumn{3}{c}{\textit{$\approx$3.3$\,\times$}} \\
Score &  & 51.1 & \textbf{61.8} & \cellcolor{expTwoBetter}+10.7$\pm$1.6 & 47.4 & \textbf{64.3} & \cellcolor{expTwoBetter}+16.9$\pm$2.9 & 43.3 & \textbf{61.6} & \cellcolor{expTwoBetter}+18.3$\pm$5.4 \\
Tokens &  & 656 & \textbf{511} & \cellcolor{expTwoBetter}$-$22.1$\pm$6.4\% & 412 & \textbf{353} & \cellcolor{expTwoBetter}$-$14.2$\pm$6.8\% & 779 & \textbf{572} & \cellcolor{expTwoBetter}$-$26.5$\pm$6.6\% \\
$\mathcal{S}$ GPU (ms) &  & 19,665 & \textbf{3,258} & \cellcolor{expTwoBetter}$\times$6.0$\pm$0.6 & 9,734 & \textbf{2,895} & \cellcolor{expTwoBetter}$\times$3.4$\pm$0.3 & 48,709 & \textbf{4,155} & \cellcolor{expTwoBetter}$\times$11.7$\pm$1.4 \\
$\mathcal{R}$ TTFT (ms) &  & 34.0 & \textbf{4.2} & \cellcolor{expTwoBetter}$\times$8.1$\pm$1.3 & 45.0 & \textbf{7.6} & \cellcolor{expTwoBetter}$\times$5.9$\pm$0.7 & 29.4 & \textbf{3.6} & \cellcolor{expTwoBetter}$\times$8.1$\pm$1.3 \\
\addlinespace
\rowcolor{expTwoSection}
 & \textit{Overall} & \multicolumn{3}{c}{} & \multicolumn{3}{c}{} & \multicolumn{3}{c}{} \\
Score &  & 42.9 & \textbf{57.8} & \cellcolor{expTwoBetter}+14.9$\pm$2.5 & 42.9 & \textbf{61.9} & \cellcolor{expTwoBetter}+19.0$\pm$3.1 & 38.0 & \textbf{56.2} & \cellcolor{expTwoBetter}+18.2$\pm$4.6 \\
Tokens &  & 293 & \textbf{254} & \cellcolor{expTwoBetter}$-$13.1$\pm$3.8\% & 269 & \textbf{243} & \cellcolor{expTwoBetter}$-$9.7$\pm$4.3\% & 309 & \textbf{269} & \cellcolor{expTwoBetter}$-$13.1$\pm$3.4\% \\
$\mathcal{S}$ GPU (ms) &  & 8,752 & \textbf{2,150} & \cellcolor{expTwoBetter}$\times$4.1$\pm$0.5 & 6,198 & \textbf{2,360} & \cellcolor{expTwoBetter}$\times$2.6$\pm$0.3 & 19,761 & \textbf{2,888} & \cellcolor{expTwoBetter}$\times$6.8$\pm$0.8 \\
$\mathcal{R}$ TTFT (ms) &  & 18.0 & \textbf{4.0} & \cellcolor{expTwoBetter}$\times$4.5$\pm$0.5 & 31.8 & \textbf{7.5} & \cellcolor{expTwoBetter}$\times$4.3$\pm$0.4 & 14.1 & \textbf{3.4} & \cellcolor{expTwoBetter}$\times$4.2$\pm$0.5 \\
\bottomrule
\end{tabular}
\endgroup
\end{table*}

\subsection{Setup}
\label{subsec:exp-setup}

\paragraph{Models.}
We train \swat{} across sender--receiver pairs drawn from six model
families: Qwen~3~\cite{family_qwen3}, Llama~3~\cite{family_llama3},
Gemma~2/3~\cite{family_gemma2,family_gemma3}, Ministral~3~\cite{family_ministral3},
Granite~3~\cite{family_granite}, and SmolLM3~\cite{family_smollm3},
ranging from 3B to 32B parameters.
Because the number of possible sender--receiver pairs grows combinatorially
with the number of models, we evaluate 10 pairs chosen so that every pair
crosses model families.
Seven pairs transfer from a smaller sender to a larger receiver and three
from a larger sender to a smaller one.

\paragraph{Training data.}
The MSE loss (\cref{eq:loss-mse}) trains for 5k steps with samples drawn from PG-19 and WikiText-103 and the training splits of QASPER, QuALITY, HotpotQA, and MuSiQue.
The NLL loss (\cref{eq:loss-ce}) adds the training splits of ARC-Challenge, HellaSwag, WinoGrande,
MMLU (auxiliary), and GSM8K with generated reasoning traces.
All training data is English text, totalling ~44M unique tokens.

\paragraph{Optimization.}
Each \swat{} has $B\in\{7,8,9\}$ blocks with $D=1024$, $H=16$, and $F=4096$,
giving approximately 300M trainable parameters.
We train with AdamW, weight decay $10^{-3}$, gradient
clipping at~1.0, all in bf16.
We first minimize \cref{eq:loss-mse} for 5{,}000 steps with a
cosine learning rate peaking at $3\times10^{-4}$,
then minimize \cref{eq:loss-ce} for 24{,}000 steps at a
constant learning rate of $3\times10^{-6}$.

\paragraph{Evaluation.}
\label{par:exp-setup-eval}
We evaluate on 17 LLM benchmarks: 9 whose training splits \swat{} saw
(in-domain) and 8 held out entirely (out-of-domain).
For short task benchmarks (ARC-Challenge, HellaSwag, WinoGrande, MMLU, GSM8K, MedQA,
OpenBookQA, MATH-500), each sample is a short question. Thus, we consider $d$ to be the task description, question, and multiple choices (if any), while the query $q$ is empty ($\emptyset$).
For long context benchmarks (QuALITY, QASPER,
HotpotQA, MuSiQue, BoolQ, RACE, LongHealth, SQuAD, NarrativeQA), we have a long context followed by a question. In these tasks, we consider $d$ to be the long context and the query $q$ to be the question and multiple choices (if any).
We report accuracy for multiple-choice and math tasks and token F1 for free-form answers, all with greedy decoding.
We additionally evaluate the vision experiments on 5 VLM benchmarks: VQAv2,
TallyQA, OK-VQA, GQA, and VizWiz. As \swat{} was trained only on English text, it has of course never seen these VLM datasets.

\begin{table}[t]
\caption{
Test accuracy (\%) of 3 VLM $\rightarrow$ text-only LLM pairs on 5 VQA benchmarks (TQA$=$TallyQA, VW$=$VizWiz). $\pm$ is bootstrap standard error. \emph{Sender}: VLM answers from the image
and question.
\emph{\ttot{}}: the receiver LLM answers from the sender VLM's caption. \emph{\swat{}}: the sender VLM's KV-cache is transferred
to the text-only receiver. \swat{} is significantly better than results marked with $^*$ (95\% paired bootstrap).
}
\label{tab:vision-subset}
\centering
\begin{tabular}{lrrrrrr}
\toprule
Method & VQAv2 & TQA & OKVQA & GQA & VW & Avg. \\
\midrule
\multicolumn{7}{l}{\makebox[2ex]{\raisebox{-0.4ex}{\includegraphics[height=2ex,draft=false]{assets/logos/gemma.png}}} \textbf{Gemma 3 4B} $\rightarrow$ \makebox[2ex]{\raisebox{-0.4ex}{\includegraphics[height=2ex,draft=false]{assets/logos/granite.png}}} \textbf{Granite 3.3 8B}} \\
Sender & 45.3$^*$ & 53.4$^*$ & 22.0$^*$ & 31.4$^*$ & \textbf{19.0} & 34.2$\pm$.4$^*$ \\
\ttot{} & 33.8$^*$ & 32.0$^*$ & 15.7$^*$ & 16.3$^*$ & 7.7$^*$ & 21.1$\pm$.4$^*$ \\
\swat{} & \textbf{59.5} & \textbf{61.8} & \textbf{28.7} & \textbf{39.5} & 18.3 & \textbf{41.6$\pm$.4} \\
\multicolumn{7}{l}{\makebox[2ex]{\raisebox{-0.4ex}{\includegraphics[height=2ex,draft=false]{assets/logos/gemma.png}}} \textbf{Gemma 3 4B} $\rightarrow$ \makebox[2ex]{\raisebox{-0.4ex}{\includegraphics[height=2ex,draft=false]{assets/logos/llama.png}}} \textbf{Llama 3 8B}} \\
Sender & 45.3$^*$ & 53.4$^*$ & 22.0$^*$ & 31.4$^*$ & \textbf{19.0} & 34.2$\pm$.4$^*$ \\
\ttot{} & 16.4$^*$ & 12.3$^*$ & 18.3$^*$ & 10.9$^*$ & 8.4$^*$ & 13.3$\pm$.3$^*$ \\
\swat{} & \textbf{57.5} & \textbf{56.4} & \textbf{29.7} & \textbf{35.5} & 18.7 & \textbf{39.6$\pm$.4} \\
\multicolumn{7}{l}{\makebox[2ex]{\raisebox{-0.4ex}{\includegraphics[height=2ex,draft=false]{assets/logos/gemma.png}}} \textbf{Gemma 3 12B} $\rightarrow$ \makebox[2ex]{\raisebox{-0.4ex}{\includegraphics[height=2ex,draft=false]{assets/logos/qwen.png}}} \textbf{Qwen 3 4B}} \\
Sender & 44.5$^*$ & 22.4$^*$ & \textbf{24.4} & 28.3$^*$ & \textbf{20.0} & 27.9$\pm$.4$^*$ \\
\ttot{} & 47.0$^*$ & 37.6$^*$ & 19.1$^*$ & 29.7$^*$ & 12.2$^*$ & 29.1$\pm$.4$^*$ \\
\swat{} & \textbf{51.7} & \textbf{49.4} & 21.8 & \textbf{33.0} & 15.1 & \textbf{34.2$\pm$.4} \\
\bottomrule
\end{tabular}
\end{table}

\begin{table}[t]
\caption{
Zero-shot specialized knowledge transfer from Qwen 3 4B to Llama 3 8B on CMMLU (Chinese MMLU; accuracy, \%). Although our training data contained only English text, \swat{} transfers most of Qwen's advantage on foreign text never seen during training. Change is relative to Llama 3 8B.
}
\label{tab:chinese-knowledge}
\centering
\begingroup
\definecolor{chinesegain}{HTML}{228B45}
\definecolor{chineseloss}{HTML}{B64040}
\begin{tabular}{lcccc}
\toprule
\multirow{2}{*}{CMMLU subject} & Qwen & Llama & \multicolumn{2}{c}{\swat{} Q$\rightarrow$L} \\
\cmidrule(lr){4-5}
 & 3 4B & 3 8B & Acc. & Change \\
\midrule
Arts & 90.0 & 55.6 & 88.8 & \textcolor{chinesegain}{+59.6\%} \\
Security study & 83.7 & 55.6 & 77.0 & \textcolor{chinesegain}{+38.7\%} \\
Concept. physics & 83.7 & 45.6 & 73.5 & \textcolor{chinesegain}{+61.2\%} \\
Prof. psychology & 82.8 & 56.0 & 77.6 & \textcolor{chinesegain}{+38.5\%} \\
\addlinespace
All 67 subjects & 69.7 & 51.2 & 65.1 & \textcolor{chinesegain}{+27.1\%} \\
\bottomrule
\end{tabular}
\endgroup
\end{table}

\subsection{Retention Across Model Pairs}
\label{subsec:exp-retention}

In \cref{tab:exp1-retention}, we evaluate the retention capabilities of the transferred cache from \swat{} compared to if the model perceives the raw text directly. Concretely, for each sender--receiver pair, we calculate the score of the receiver solving the task $(d,q)$ normally by itself (the denominator) and compare it with the score of the sender prefilling $d$ and using \swat{} to transfer its KV-cache to the receiver which only sees the raw value of $q$ (where $d$, $q$ are dataset specific and defined in \cref{par:exp-setup-eval}). Across the 10 model--pairs evaluated on 17 benchmarks, the lowest mean retention is $>90\%$ even in the ‘small to large’ pairing where the large model has to solve the task using the smaller model's KV-cache compared to using its own KV-cache. In the `large to small' pairs, the lowest mean retention is $>100\%$, which is logically plausible as the receiver utilizes the larger model's KV-cache to solve the task instead of relying on its own (smaller model) KV-cache. \Cref{tab:exp1-retention} indicates that across our tested benchmarks, a receiver agent viewing the task and solving it through the \swat{} translated KV-cache of the sender agent achieves, on average, performance comparable to that of the receiver agent viewing the task itself (in the `small to large' case), or performance that even slightly exceeds that performance (in the `large to small' case). Additional details are provided in \cref{apdx-sec:retension}.

\subsection{Cache Transfer versus Text Communication}
\label{subsec:exp-t2t}

We compare the effectiveness of using \swat{} with \am{} against text-to-text communication (\ttot{}) as described in \cref{subsec:method-t2t-vs-am}. We extensively compare the two methods across three model--pairs, five budgets, and seven benchmarks. We provide results averaged over the seven benchmarks in \cref{tab:exp2-mean_se} (arithmetic mean for score, geometric mean for other metrics). We additionally report the overall mean across all budgets and datasets, the bootstrapped standard error resampling datasets and their samples, as well as reporting when \swat{}+\am{} is significantly better than \ttot{} (95\% bootstrap CI). In addition, we report individual results not averaged per dataset in \cref{fig:exp2_3scatters}. Additional details are provided in \cref{apdx-sec:exp2}.

\begin{table*}[t]
\caption{
Receiver re-prefill vs \swat{} FLOP count and latency.
}
\label{tab:compute-flop-speed}
\centering
\begin{tabular}{r|rrrrrr|rrrrrr}
\toprule
 & \multicolumn{6}{c|}{Qwen 3 4B \makebox[2ex]{\raisebox{-0.4ex}{\includegraphics[height=2ex,draft=false]{assets/logos/qwen.png}}}\;$\rightarrow$\;\makebox[2ex]{\raisebox{-0.4ex}{\includegraphics[height=2ex,draft=false]{assets/logos/llama.png}}} Llama 3 8B} & \multicolumn{6}{c}{Llama 3 8B \makebox[2ex]{\raisebox{-0.4ex}{\includegraphics[height=2ex,draft=false]{assets/logos/llama.png}}}\;$\rightarrow$\;\makebox[2ex]{\raisebox{-0.4ex}{\includegraphics[height=2ex,draft=false]{assets/logos/qwen.png}}} Qwen 3 14B} \\
\cmidrule(lr){2-7}\cmidrule(lr){8-13}
 & \multicolumn{3}{c}{FLOPs (TFLOPs)} & \multicolumn{3}{c|}{Latency (ms)} & \multicolumn{3}{c}{FLOPs (TFLOPs)} & \multicolumn{3}{c}{Latency (ms)} \\
\cmidrule(lr){2-4}\cmidrule(lr){5-7}\cmidrule(lr){8-10}\cmidrule(lr){11-13}
Seq Len & \multicolumn{1}{c}{Re-prefill} & \multicolumn{1}{c}{\swat{}} & \multicolumn{1}{c}{Ratio} & \multicolumn{1}{c}{Re-prefill} & \multicolumn{1}{c}{\swat{}} & \multicolumn{1}{c|}{Speedup} & \multicolumn{1}{c}{Re-prefill} & \multicolumn{1}{c}{\swat{}} & \multicolumn{1}{c}{Ratio} & \multicolumn{1}{c}{Re-prefill} & \multicolumn{1}{c}{\swat{}} & \multicolumn{1}{c}{Speedup} \\
\midrule
64 & 0.89 & 0.04 & $\times$24.7 & 26.0 & 2.1 & $\times$12.3 & 1.69 & 0.04 & $\times$45.3 & 43.5 & 2.2 & $\times$20.2 \\
256 & 3.59 & 0.15 & $\times$24.4 & 31.7 & 2.6 & $\times$12.0 & 6.79 & 0.15 & $\times$44.7 & 47.8 & 2.7 & $\times$18.0 \\
1K & 14.57 & 0.63 & $\times$23.2 & 52.9 & 5.7 & $\times$\phantom{0}9.3 & 27.49 & 0.65 & $\times$42.5 & 95.8 & 5.7 & $\times$16.7 \\
4K & 61.57 & 3.13 & $\times$19.7 & 209.9 & 21.7 & $\times$\phantom{0}9.7 & 115.11 & 3.20 & $\times$35.9 & 394.1 & 22.3 & $\times$17.7 \\
16K & 299.07 & 22.40 & $\times$13.3 & 1,054.5 & 131.4 & $\times$\phantom{0}8.0 & 542.89 & 22.71 & $\times$23.9 & 1,912.0 & 134.9 & $\times$14.2 \\
32K & 738.88 & 71.19 & $\times$10.4 & 2,763.2 & 391.6 & $\times$\phantom{0}7.1 & 1,305.68 & 71.81 & $\times$18.2 & 4,799.9 & 399.0 & $\times$12.0 \\
64K & 2,040.71 & 247.94 & $\times$\phantom{0}8.2 & 7,966.4 & 1,296.3 & $\times$\phantom{0}6.1 & 3,490.98 & 249.18 & $\times$14.0 & 13,465.7 & 1,314.9 & $\times$10.2 \\
\bottomrule
\end{tabular}
\end{table*}

\subsection{Zero-Shot Capabilities}

\swat{} was only trained on English text, yet we uncover additional capabilities transferred to the receiver model outside of English text. Additional details and capabilities are provided in the supplementary material.

\subsubsection{Transferring Vision to Text-Only Receivers}
\label{subsec:exp-vision}

Despite \swat{} only training on English text, we want to know whether \swat{} can provide the receiver with something it never had: vision. We use Gemma~3 VLMs as senders and text-only LLMs as receivers. We take multiple VLM benchmarks and sample image--question pairs $(I, q)$ from each. The sender looks at the image and writes a short caption $P$. In \ttot{}, the receiver reads $P$ as text and attempts to answer $q$. With \swat{}, the sender prefills $I \Vert P$ (where the VLM enriches the KV-cache of $P$ with visual information $I$) and we take the KV-cache of $P$ and transfer it using \swat{} to the receiver which then attempts to answer $q$. We report these results in \cref{tab:vision-subset}. \swat{} beats \ttot{} and even beats the sender VLM answering by itself. The trend against \ttot{} holds for many more benchmarks while the trend against the vanilla sender VLM does not hold for some benchmarks especially ones that require detailed image information. Additional details are provided in \cref{apdx-sec:vision}.

\subsubsection{Transferring Specialized Knowledge}
\label{subsec:exp-specialized}

We next ask whether \swat{} can transfer from the sender specialized knowledge that the receiver lacks. Specifically we focus on languages as they occupy their own tokens that aren't in \swat{}'s English-only training data. Qwen~3~4B is much stronger in Chinese than Llama~3~8B. On all 67 subjects of CMMLU it scores approximately 20 points more than Llama. We let Qwen read each question and pass its cache to Llama through \swat{}. We report these results in \cref{tab:chinese-knowledge}. This makes Llama close three quarters of the gap to Qwen.

\section{Comparisons}

\paragraph{Compute.}
A self-evident but important aspect of \swat{} is that it is designed to require much less compute (both in terms of FLOPs and wall-clock time) to transfer the KV-cache from sender to receiver than it would take the receiver to prefill the original text. In \cref{tab:compute-flop-speed} we compare this for sequence lengths ranging from $64$ to $64{,}000$. We point out that in the worst case across all sequence lengths, the FLOP count of \swat{} is at least $\times8$ and $\times14$ less than what is required for a re-prefill by the receiver across the two pairs. In terms of latency, \swat{} is approximately $\times10$ and $\times16$ faster for short context lengths ($<1$K tokens) across the two pairs. For extremely long context lengths up to $64$K tokens, the two pairs are still $\times6$ and $\times10$ faster respectively. We note that the re-prefill FLOP count and latency significantly increase when going from the 8B receiver to the 14B receiver while \swat{} costs roughly the same across the two.

\paragraph{Other methods.}
Although other KV-cache transfer methods do not target the same capabilities as \swat{} (as highlighted in \cref{tab:cross-model-kv-comparison}), we implement two recent cross-model and cross-context KV-cache transfer methods (referred to as `Closed Form' \cite{kv_nvidia} and `Dense Latent' \cite{kv_dense_latent}). Since neither method was designed for cross-family KV-cache transfer, we augment both with cross-tokenizer-aware functionality to work across tokenizers. We report the results of the two methods compared to \swat{} on three model--pairs averaged across 17 benchmarks in \cref{tab:abl-dense}. Each method and model--pair was trained and evaluated three times with standard error displayed. We note that \cite{kv_nvidia} was given different calibration data across the three seeds to evaluate variance; the two stochastic gradient descent methods were simply given different seeds. 

\begin{table}[t]
\caption{
Mean (\%, F1) over 17 benchmarks ($\pm$ SE over 3 training seeds). Other methods are provided
cross-tokenizer-aware alignment to function across model families. \swat{} exceeds others by $\geq$20 points in the cross-generation/family cases.
}
\label{tab:abl-dense}
\centering
\begin{tabular}{llccc}
\toprule
\begin{tabular}[c]{@{}l@{}}Sender \\ \hspace{0.6em}$\rightarrow$ Receiver\end{tabular} & Mode & \shortstack{Closed \\ form~\cite{kv_nvidia}} & \shortstack{Dense \\ latent~\cite{kv_dense_latent}} & \shortstack{\swat{} \\ (ours)} \\
\midrule
\begin{tabular}[c]{@{}l@{}}Qwen 3 4B \\ \hspace{0.6em}$\rightarrow$ Qwen 3 8B\end{tabular} & \begin{tabular}[c]{@{}l@{}}Same \\ Family\end{tabular} & 59.9\,{\scriptsize$\pm$0.2} & \textbf{73.5}\,{\scriptsize$\pm$0.1} & 72.3\,{\scriptsize$\pm$0.2} \\
\midrule
\begin{tabular}[c]{@{}l@{}}Qwen 3 4B \\ \hspace{0.6em}$\rightarrow$ Qwen 2 7B\end{tabular} & \begin{tabular}[c]{@{}l@{}}Cross \\ Gen.\end{tabular} & 41.0\,{\scriptsize$\pm$0.5} & 47.3\,{\scriptsize$\pm$0.2} & \textbf{68.1}\,{\scriptsize$\pm$0.1} \\
\midrule
\begin{tabular}[c]{@{}l@{}}Qwen 3 4B \\ \hspace{0.6em}$\rightarrow$ Llama 3 8B\end{tabular} & \begin{tabular}[c]{@{}l@{}}Cross \\ Family\end{tabular} & 30.5\,{\scriptsize$\pm$2.2} & 43.5\,{\scriptsize$\pm$0.3} & \textbf{63.5}\,{\scriptsize$\pm$0.1} \\
\bottomrule
\end{tabular}
\end{table}

\section{Discussion}
\textbf{Limitations.}
\swat{} has several limitations. First, each sender--receiver pair requires a trained mapper. For a multi-agent system with $n$ families, we would need $n(n-1)$ trained mappers, which quickly becomes infeasible. Future work might investigate having a shared latent space to train mappers in and out of it for each agent which would total to only $2n$ mappers being needed. Another limitation is our focus on individual message handoffs instead of repeated transfers or aggregating multiple messages together, both of which introduce challenges as KV-cache degrades and isn't as composable as text. Future work might integrate \swat{} with recent KV-cache composition techniques such as \citet{kvlink}.

\textbf{Ethical considerations.}
\swat{} aims to improve the capabilities of multi-agent systems. The current implementation requires the sender's text to be available; thus, text-based monitoring is possible. However, the transferred cache may contain information beyond what is explicit in the text, which highlights the importance of future work investigating the auditability and safety of the benefits of KV-cache transfer while still utilizing the benefits.

\section{Related Work}

\textbf{Text communication in multi-agent systems.}
LLM multi-agent systems have progressed rapidly solving novel challenges \cite{text_mas_rw1,text_mas_rw2,mas_text1,mas_text2,mas_text3}. In these systems, communication is primarily in natural language and can occur in various ways to improve the performance of the system such as through debate \cite{mas_text7,mas_text8,mas_text2}, ensembling \cite{mas_text3,mas_text1}, and even fine-tuning on communication \cite{vighnesh_finetune}.

\textbf{Cross-model KV-cache transfer.}
Many recent works investigate whether KV-cache can be transferred between models. Certain recent works re-use KV-cache within variations of the same model \cite{kv_droid,kv_icarus}. Other works extend KV-cache transfer across different models within the same family without requiring the receiver to prefill \cite{kv_dense_latent,kv_nvidia,kv_google,kv_cachebridge}. Other recent works translate KV-cache across families but require the receiver to prefill text \cite{kv_lcf,kv_c2c,kv_xkv,kv_mot}.

\section{Conclusion}
We introduce \swat{}, a cross-family KV-cache transfer method between LLMs without receiver prefill. Evaluation across many model--pairs and datasets shows the effectiveness of \swat{} as a communication protocol in multi-agent systems, achieving more accurate and faster communication compared to text-to-text communication.

\clearpage

% Future work: KVLink, larger training beyond 50 M tokens? training with RL? Various training techniques can push the training token budget well below our 50M token limit as well as below our 1 GPU-day training runs. In our experiments we focus on demonstraining \swat{} across various families in a sender receiver settings, future work could expand on this setting by experimenting with multi-agent systems involving three or more agents simultaneously communicating using \swat{} to solve a single problem.

\bibliographystyle{ACM-Reference-Format} 
\bibliography{main}

\appendix
\crefalias{section}{appendix}
\crefname{appendix}{Appendix}{Appendices}

\clearpage
\clearpage
\clearpage

\section*{Appendix}

\section{Code}
\label{subsec:code_repo}

We provide the code repository at:
\begin{center}
\url{https://github.com/Ar-Kareem/SwaT}
\end{center}

\section{AI use}

AI-assisted technologies were used for polishing the text, helping with \LaTeX \ format, and for writing code and plotting scripts. AI was not used in the creation of hypotheses or methodologies.

\section{Additional retention data}
\label{apdx-sec:retension}

In \cref{tab:exp1-retention-in-domain,tab:exp1-retention-ood} we show the breakdown of all pairs on all datasets of the aggregate \cref{tab:exp1-retention}. We note that MATH-500 is a highly out-of-distribution task that requires answers formatted as \LaTeX \ expressions while the training data explicitly trains on GSM8K where final answers are numerical values. 

\begin{table*}[t]
\caption{
In-domain (9): per-benchmark receiver score with its own context, \swat{} score, and zero-shot retention (italic; \swat{} over receiver). Scores in percent. HSwag = HellaSwag; WGrande = WinoGrande; Hotpot = HotpotQA.
}
\label{tab:exp1-retention-in-domain}
\centering
\begin{tabular}{llrrrrrrrrr}
\toprule
\multicolumn{2}{c}{} & \multicolumn{5}{c}{Short} & \multicolumn{4}{c}{Context} \\
\cmidrule(lr){3-7}\cmidrule(lr){8-11}
Pair & Row & \shortstack{\strut ARC-C \\ \strut Acc} & \shortstack{\strut HSwag \\ \strut Acc} & \shortstack{\strut WGrande \\ \strut Acc} & \shortstack{\strut MMLU \\ \strut Acc} & \shortstack{\strut GSM8K \\ \strut Acc} & \shortstack{\strut QuALITY \\ \strut Acc} & \shortstack{\strut QASPER \\ \strut F1} & \shortstack{\strut Hotpot \\ \strut F1} & \shortstack{\strut MuSiQue \\ \strut F1} \\
\midrule
\multicolumn{11}{l}{\textbf{Small} $\boldsymbol\rightarrow$ \textbf{Large}} \\
\makebox[2ex]{\raisebox{-0.4ex}{\includegraphics[height=2ex,draft=false]{assets/logos/qwen.png}}} Qwen 3 4B & Receiver & 80.6 & 77.2 & 59.2 & 62.2 & 75.2 & 64.1 & 49.0 & 67.4 & 34.6 \\
$\rightarrow$ \makebox[2ex]{\raisebox{-0.4ex}{\includegraphics[height=2ex,draft=false]{assets/logos/llama.png}}} Llama 3 8B & \swat{} & 85.2 & 83.6 & 58.8 & 62.0 & 67.6 & 64.2 & 51.1 & 70.6 & 42.6 \\
 & \textit{Retention (\%)} & \textit{105.7} & \textit{108.3} & \textit{99.3} & \textit{99.7} & \textit{89.9} & \textit{100.2} & \textit{104.4} & \textit{104.8} & \textit{123.4} \\
\midrule
\makebox[2ex]{\raisebox{-0.4ex}{\includegraphics[height=2ex,draft=false]{assets/logos/llama.png}}} Llama 3 8B & Receiver & 92.6 & 91.4 & 69.4 & 73.2 & 89.8 & 80.1 & 49.1 & 73.1 & 43.4 \\
$\rightarrow$ \makebox[2ex]{\raisebox{-0.4ex}{\includegraphics[height=2ex,draft=false]{assets/logos/qwen.png}}} Qwen 3 14B & \swat{} & 78.6 & 80.6 & 65.6 & 60.0 & 88.6 & 78.7 & 56.3 & 78.2 & 55.0 \\
 & \textit{Retention (\%)} & \textit{84.9} & \textit{88.2} & \textit{94.5} & \textit{82.0} & \textit{98.7} & \textit{98.2} & \textit{114.7} & \textit{106.9} & \textit{126.8} \\
\midrule
\makebox[2ex]{\raisebox{-0.4ex}{\includegraphics[height=2ex,draft=false]{assets/logos/ministral.png}}} Ministral 3 8B & Receiver & 80.0 & 80.6 & 60.6 & 66.4 & 79.8 & 72.9 & 49.3 & 71.3 & 38.6 \\
$\rightarrow$ \makebox[2ex]{\raisebox{-0.4ex}{\includegraphics[height=2ex,draft=false]{assets/logos/llama.png}}} Llama 3.2 11B & \swat{} & 88.8 & 86.6 & 64.2 & 68.4 & 74.8 & 71.0 & 49.4 & 71.9 & 48.0 \\
 & \textit{Retention (\%)} & \textit{111.0} & \textit{107.4} & \textit{105.9} & \textit{103.0} & \textit{93.7} & \textit{97.5} & \textit{100.3} & \textit{100.9} & \textit{124.5} \\
\midrule
\makebox[2ex]{\raisebox{-0.4ex}{\includegraphics[height=2ex,draft=false]{assets/logos/gemma.png}}} Gemma 3 4B & Receiver & 80.4 & 77.2 & 62.6 & 61.0 & 77.4 & 67.2 & 33.3 & 47.9 & 24.4 \\
$\rightarrow$ \makebox[2ex]{\raisebox{-0.4ex}{\includegraphics[height=2ex,draft=false]{assets/logos/granite.png}}} Granite 3.3 8B & \swat{} & 73.4 & 77.0 & 63.2 & 52.0 & 77.4 & 62.9 & 53.7 & 67.3 & 43.2 \\
 & \textit{Retention (\%)} & \textit{91.3} & \textit{99.7} & \textit{101.0} & \textit{85.2} & \textit{100.0} & \textit{93.6} & \textit{161.6} & \textit{140.6} & \textit{177.2} \\
\midrule
\makebox[2ex]{\raisebox{-0.4ex}{\includegraphics[height=2ex,draft=false]{assets/logos/qwen.png}}} Qwen 3 4B & Receiver & 90.6 & 88.2 & 69.4 & 67.8 & 80.0 & 75.1 & 45.7 & 73.4 & 46.5 \\
$\rightarrow$ \makebox[2ex]{\raisebox{-0.4ex}{\includegraphics[height=2ex,draft=false]{assets/logos/gemma.png}}} Gemma 2 9B & \swat{} & 85.2 & 85.0 & 63.0 & 64.2 & 75.2 & 69.4 & 49.6 & 71.8 & 49.4 \\
 & \textit{Retention (\%)} & \textit{94.0} & \textit{96.4} & \textit{90.8} & \textit{94.7} & \textit{94.0} & \textit{92.4} & \textit{108.5} & \textit{97.8} & \textit{106.3} \\
\midrule
\makebox[2ex]{\raisebox{-0.4ex}{\includegraphics[height=2ex,draft=false]{assets/logos/gemma.png}}} Gemma 2 9B & Receiver & 92.0 & 84.2 & 66.8 & 72.4 & 82.8 & 80.0 & 42.2 & 75.1 & 51.3 \\
$\rightarrow$ \makebox[2ex]{\raisebox{-0.4ex}{\includegraphics[height=2ex,draft=false]{assets/logos/ministral.png}}} Ministral 3 14B & \swat{} & 88.2 & 91.8 & 70.4 & 66.8 & 88.4 & 76.3 & 51.5 & 75.4 & 55.3 \\
 & \textit{Retention (\%)} & \textit{95.9} & \textit{109.0} & \textit{105.4} & \textit{92.3} & \textit{106.8} & \textit{95.4} & \textit{121.9} & \textit{100.4} & \textit{107.7} \\
\midrule
\makebox[2ex]{\raisebox{-0.4ex}{\includegraphics[height=2ex,draft=false]{assets/logos/gemma.png}}} Gemma 3 4B & Receiver & 80.6 & 77.2 & 59.2 & 62.2 & 75.2 & 64.1 & 49.0 & 67.4 & 34.6 \\
$\rightarrow$ \makebox[2ex]{\raisebox{-0.4ex}{\includegraphics[height=2ex,draft=false]{assets/logos/llama.png}}} Llama 3 8B & \swat{} & 70.2 & 76.0 & 54.4 & 51.0 & 69.0 & 65.1 & 50.7 & 69.6 & 39.3 \\
 & \textit{Retention (\%)} & \textit{87.1} & \textit{98.4} & \textit{91.9} & \textit{82.0} & \textit{91.8} & \textit{101.6} & \textit{103.6} & \textit{103.4} & \textit{113.8} \\
\midrule
\multicolumn{11}{l}{\textbf{Large} $\boldsymbol\rightarrow$ \textbf{Small}} \\
\makebox[2ex]{\raisebox{-0.4ex}{\includegraphics[height=2ex,draft=false]{assets/logos/gemma.png}}} Gemma 3 12B & Receiver & 87.4 & 84.6 & 59.8 & 63.6 & 85.0 & 70.2 & 47.0 & 64.1 & 31.8 \\
$\rightarrow$ \makebox[2ex]{\raisebox{-0.4ex}{\includegraphics[height=2ex,draft=false]{assets/logos/qwen.png}}} Qwen 3 4B & \swat{} & 87.8 & 90.0 & 66.4 & 66.8 & 82.6 & 70.3 & 53.2 & 71.4 & 43.6 \\
 & \textit{Retention (\%)} & \textit{100.5} & \textit{106.4} & \textit{111.0} & \textit{105.0} & \textit{97.2} & \textit{100.1} & \textit{113.3} & \textit{111.4} & \textit{136.8} \\
\midrule
\makebox[2ex]{\raisebox{-0.4ex}{\includegraphics[height=2ex,draft=false]{assets/logos/qwen.png}}} Qwen 3 32B & Receiver & 80.6 & 77.2 & 59.2 & 62.2 & 75.2 & 64.1 & 49.0 & 67.4 & 34.6 \\
$\rightarrow$ \makebox[2ex]{\raisebox{-0.4ex}{\includegraphics[height=2ex,draft=false]{assets/logos/llama.png}}} Llama 3 8B & \swat{} & 94.8 & 92.0 & 68.2 & 77.0 & 66.8 & 66.7 & 48.4 & 67.3 & 33.4 \\
 & \textit{Retention (\%)} & \textit{117.6} & \textit{119.2} & \textit{115.2} & \textit{123.8} & \textit{88.8} & \textit{104.0} & \textit{98.8} & \textit{99.9} & \textit{96.6} \\
\midrule
\makebox[2ex]{\raisebox{-0.4ex}{\includegraphics[height=2ex,draft=false]{assets/logos/granite.png}}} Granite 3.3 8B & Receiver & 78.2 & 71.8 & 54.4 & 53.6 & 73.8 & 60.9 & 28.4 & 30.7 & 11.5 \\
$\rightarrow$ \makebox[2ex]{\raisebox{-0.4ex}{\includegraphics[height=2ex,draft=false]{assets/logos/smollm3.png}}} SmolLM3 3B & \swat{} & 82.8 & 85.6 & 68.2 & 59.2 & 74.6 & 63.3 & 47.8 & 65.9 & 43.3 \\
 & \textit{Retention (\%)} & \textit{105.9} & \textit{119.2} & \textit{125.4} & \textit{110.4} & \textit{101.1} & \textit{103.9} & \textit{168.4} & \textit{214.6} & \textit{377.8} \\
\bottomrule
\end{tabular}
\end{table*}

\begin{table*}[t]
\caption{
OOD (8): per-benchmark receiver score with its own context, \swat{} score, and zero-shot retention (italic; \swat{} over receiver). Scores in percent.
}
\label{tab:exp1-retention-ood}
\centering
\begin{tabular}{llrrrrrrrr}
\toprule
\multicolumn{2}{c}{} & \multicolumn{3}{c}{Short} & \multicolumn{5}{c}{Context} \\
\cmidrule(lr){3-5}\cmidrule(lr){6-10}
Pair & Row & \shortstack{\strut MedQA \\ \strut Acc} & \shortstack{\strut OpenBookQA \\ \strut Acc} & \shortstack{\strut MATH-500 \\ \strut Acc} & \shortstack{\strut BoolQ \\ \strut Acc} & \shortstack{\strut RACE \\ \strut Acc} & \shortstack{\strut LongHealth \\ \strut Acc} & \shortstack{\strut SQuAD \\ \strut F1} & \shortstack{\strut NarrQA \\ \strut F1} \\
\midrule
\multicolumn{10}{l}{\textbf{Small} $\boldsymbol\rightarrow$ \textbf{Large}} \\
\makebox[2ex]{\raisebox{-0.4ex}{\includegraphics[height=2ex,draft=false]{assets/logos/qwen.png}}} Qwen 3 4B & Receiver & 60.0 & 76.0 & 28.4 & 83.7 & 81.3 & 66.8 & 88.9 & 74.2 \\
$\rightarrow$ \makebox[2ex]{\raisebox{-0.4ex}{\includegraphics[height=2ex,draft=false]{assets/logos/llama.png}}} Llama 3 8B & \swat{} & 54.8 & 79.2 & 20.0 & 79.8 & 76.4 & 57.0 & 87.5 & 73.4 \\
 & \textit{Retention (\%)} & \textit{91.3} & \textit{104.2} & \textit{70.4} & \textit{95.3} & \textit{94.0} & \textit{85.4} & \textit{98.4} & \textit{98.9} \\
\midrule
\makebox[2ex]{\raisebox{-0.4ex}{\includegraphics[height=2ex,draft=false]{assets/logos/llama.png}}} Llama 3 8B & Receiver & 64.4 & 90.2 & 76.6 & 89.1 & 89.4 & 72.0 & 86.9 & 72.7 \\
$\rightarrow$ \makebox[2ex]{\raisebox{-0.4ex}{\includegraphics[height=2ex,draft=false]{assets/logos/qwen.png}}} Qwen 3 14B & \swat{} & 54.4 & 75.6 & 69.8 & 88.9 & 88.0 & 71.0 & 90.5 & 74.8 \\
 & \textit{Retention (\%)} & \textit{84.5} & \textit{83.8} & \textit{91.1} & \textit{99.7} & \textit{98.4} & \textit{98.6} & \textit{104.1} & \textit{102.9} \\
\midrule
\makebox[2ex]{\raisebox{-0.4ex}{\includegraphics[height=2ex,draft=false]{assets/logos/ministral.png}}} Ministral 3 8B & Receiver & 59.8 & 81.6 & 49.0 & 85.0 & 83.4 & 71.5 & 88.3 & 75.7 \\
$\rightarrow$ \makebox[2ex]{\raisebox{-0.4ex}{\includegraphics[height=2ex,draft=false]{assets/logos/llama.png}}} Llama 3.2 11B & \swat{} & 59.6 & 83.0 & 33.2 & 82.4 & 78.3 & 62.7 & 89.2 & 73.5 \\
 & \textit{Retention (\%)} & \textit{99.7} & \textit{101.7} & \textit{67.8} & \textit{96.9} & \textit{93.9} & \textit{87.8} & \textit{101.0} & \textit{97.1} \\
\midrule
\makebox[2ex]{\raisebox{-0.4ex}{\includegraphics[height=2ex,draft=false]{assets/logos/gemma.png}}} Gemma 3 4B & Receiver & 51.6 & 78.4 & 61.0 & 81.9 & 79.0 & 60.2 & 68.5 & 50.0 \\
$\rightarrow$ \makebox[2ex]{\raisebox{-0.4ex}{\includegraphics[height=2ex,draft=false]{assets/logos/granite.png}}} Granite 3.3 8B & \swat{} & 44.0 & 72.2 & 41.8 & 82.1 & 77.2 & 52.5 & 86.1 & 68.5 \\
 & \textit{Retention (\%)} & \textit{85.3} & \textit{92.1} & \textit{68.5} & \textit{100.2} & \textit{97.7} & \textit{87.1} & \textit{125.7} & \textit{137.2} \\
\midrule
\makebox[2ex]{\raisebox{-0.4ex}{\includegraphics[height=2ex,draft=false]{assets/logos/qwen.png}}} Qwen 3 4B & Receiver & 57.2 & 88.2 & 51.4 & 89.1 & 87.6 & 68.2 & 88.6 & 78.7 \\
$\rightarrow$ \makebox[2ex]{\raisebox{-0.4ex}{\includegraphics[height=2ex,draft=false]{assets/logos/gemma.png}}} Gemma 2 9B & \swat{} & 52.4 & 79.8 & 35.0 & 86.3 & 83.8 & 62.0 & 88.7 & 72.7 \\
 & \textit{Retention (\%)} & \textit{91.6} & \textit{90.5} & \textit{68.1} & \textit{96.8} & \textit{95.7} & \textit{90.8} & \textit{100.2} & \textit{92.3} \\
\midrule
\makebox[2ex]{\raisebox{-0.4ex}{\includegraphics[height=2ex,draft=false]{assets/logos/gemma.png}}} Gemma 2 9B & Receiver & 62.4 & 86.2 & 55.2 & 85.6 & 87.7 & 73.0 & 83.6 & 71.6 \\
$\rightarrow$ \makebox[2ex]{\raisebox{-0.4ex}{\includegraphics[height=2ex,draft=false]{assets/logos/ministral.png}}} Ministral 3 14B & \swat{} & 56.4 & 86.0 & 67.6 & 82.2 & 84.6 & 68.2 & 90.7 & 75.4 \\
 & \textit{Retention (\%)} & \textit{90.4} & \textit{99.8} & \textit{122.5} & \textit{96.0} & \textit{96.5} & \textit{93.5} & \textit{108.6} & \textit{105.2} \\
\midrule
\makebox[2ex]{\raisebox{-0.4ex}{\includegraphics[height=2ex,draft=false]{assets/logos/gemma.png}}} Gemma 3 4B & Receiver & 60.0 & 76.0 & 28.4 & 83.7 & 81.3 & 66.8 & 88.9 & 74.2 \\
$\rightarrow$ \makebox[2ex]{\raisebox{-0.4ex}{\includegraphics[height=2ex,draft=false]{assets/logos/llama.png}}} Llama 3 8B & \swat{} & 43.0 & 69.8 & 16.0 & 80.0 & 77.5 & 56.8 & 87.8 & 73.7 \\
 & \textit{Retention (\%)} & \textit{71.7} & \textit{91.8} & \textit{56.3} & \textit{95.6} & \textit{95.2} & \textit{85.0} & \textit{98.7} & \textit{99.2} \\
\midrule
\multicolumn{10}{l}{\textbf{Large} $\boldsymbol\rightarrow$ \textbf{Small}} \\
\makebox[2ex]{\raisebox{-0.4ex}{\includegraphics[height=2ex,draft=false]{assets/logos/gemma.png}}} Gemma 3 12B & Receiver & 55.0 & 80.6 & 71.0 & 84.3 & 83.9 & 68.0 & 80.7 & 68.0 \\
$\rightarrow$ \makebox[2ex]{\raisebox{-0.4ex}{\includegraphics[height=2ex,draft=false]{assets/logos/qwen.png}}} Qwen 3 4B & \swat{} & 54.4 & 82.2 & 64.2 & 85.2 & 82.5 & 63.7 & 88.0 & 69.8 \\
 & \textit{Retention (\%)} & \textit{98.9} & \textit{102.0} & \textit{90.4} & \textit{101.2} & \textit{98.3} & \textit{93.7} & \textit{109.0} & \textit{102.6} \\
\midrule
\makebox[2ex]{\raisebox{-0.4ex}{\includegraphics[height=2ex,draft=false]{assets/logos/qwen.png}}} Qwen 3 32B & Receiver & 60.0 & 76.0 & 28.4 & 83.7 & 81.3 & 66.8 & 88.9 & 74.2 \\
$\rightarrow$ \makebox[2ex]{\raisebox{-0.4ex}{\includegraphics[height=2ex,draft=false]{assets/logos/llama.png}}} Llama 3 8B & \swat{} & 71.0 & 92.4 & 19.2 & 81.9 & 78.3 & 61.5 & 86.8 & 71.6 \\
 & \textit{Retention (\%)} & \textit{118.3} & \textit{121.6} & \textit{67.6} & \textit{97.8} & \textit{96.2} & \textit{92.1} & \textit{97.7} & \textit{96.4} \\
\midrule
\makebox[2ex]{\raisebox{-0.4ex}{\includegraphics[height=2ex,draft=false]{assets/logos/granite.png}}} Granite 3.3 8B & Receiver & 29.6 & 74.6 & 61.8 & 83.5 & 74.1 & 39.2 & 50.6 & 44.0 \\
$\rightarrow$ \makebox[2ex]{\raisebox{-0.4ex}{\includegraphics[height=2ex,draft=false]{assets/logos/smollm3.png}}} SmolLM3 3B & \swat{} & 48.2 & 77.4 & 45.4 & 81.3 & 75.1 & 57.0 & 84.0 & 69.6 \\
 & \textit{Retention (\%)} & \textit{162.8} & \textit{103.8} & \textit{73.5} & \textit{97.4} & \textit{101.3} & \textit{145.2} & \textit{166.0} & \textit{158.3} \\
\bottomrule
\end{tabular}
\end{table*}

\clearpage
\clearpage
\clearpage

\section{Additional \swat{}+\am{} details}
\label{apdx-sec:exp2}

We provide a comparison of \swat{}+\am{} vs \ttot{} in all seven datasets in \cref{tab:exp2-appendix-hotpotqa,tab:exp2-appendix-longhealth,tab:exp2-appendix-musique,tab:exp2-appendix-narrativeqa,tab:exp2-appendix-qasper,tab:exp2-appendix-quality,tab:exp2-appendix-squad}. We also provide the mean over the datasets in \cref{tab:exp2-geomean} where we report both the geometric mean of all metrics and a row for the arithmetic mean over the score.

\begin{table*}[t]
\caption{
Experiment 2, HotpotQA. \ttot{}: sender summarizes to receiver. \swat{}+AM: the sender's cache rendered and compressed with Attention Matching. Budgets are determined by the sender agent in \ttot{} given one of five summary prompts and sorted from lowest to highest per pair. Change: score difference in points, relative token difference, and speedup.
}
\label{tab:exp2-appendix-hotpotqa}
\centering
\begingroup
\definecolor{expTwoSection}{HTML}{ECECEC}
\definecolor{expTwoBetter}{HTML}{D9EFD8}
\begin{tabular}{llrrrrrrrrr}
\toprule
 & & \multicolumn{3}{c}{Qwen3-4B $\rightarrow$ Llama3-8B} & \multicolumn{3}{c}{Llama3-8B $\rightarrow$ Qwen3-14B} & \multicolumn{3}{c}{Gemma3-12B $\rightarrow$ Qwen3-4B} \\
\cmidrule(lr){3-5}\cmidrule(lr){6-8}\cmidrule(lr){9-11}
Metric & Budget & \ttot{} & \swat{} & Change & \ttot{} & \swat{} & Change & \ttot{} & \swat{} & Change \\
\midrule
\rowcolor{expTwoSection}
 & \textit{Lowest} & \multicolumn{3}{c}{\textit{$\approx$10$\,\times$ compression ($P_{\mathrm{sum},1}$)}} & \multicolumn{3}{c}{\textit{$\approx$9$\,\times$ ($P_{\mathrm{sum},1}$)}} & \multicolumn{3}{c}{\textit{$\approx$10$\,\times$ ($P_{\mathrm{sum},1}$)}} \\
Score &  & 44.6 & \textbf{62.8} & \cellcolor{expTwoBetter}+18.2 & 47.2 & \textbf{67.9} & \cellcolor{expTwoBetter}+20.6 & 37.9 & \textbf{53.8} & \cellcolor{expTwoBetter}+15.9 \\
Tokens &  & 144 & \textbf{132} & \cellcolor{expTwoBetter}$-$8.2\% & 170 & \textbf{154} & \cellcolor{expTwoBetter}$-$9.6\% & 151 & \textbf{138} & \cellcolor{expTwoBetter}$-$8.3\% \\
$\mathcal{S}$ GPU (ms) &  & 4,731 & \textbf{997} & \cellcolor{expTwoBetter}$\times$4.7 & 3,687 & \textbf{1,438} & \cellcolor{expTwoBetter}$\times$2.6 & 9,452 & \textbf{1,336} & \cellcolor{expTwoBetter}$\times$7.1 \\
$\mathcal{R}$ TTFT (ms) &  & 9.8 & \textbf{3.6} & \cellcolor{expTwoBetter}$\times$2.7 & 21.8 & \textbf{6.5} & \cellcolor{expTwoBetter}$\times$3.4 & 7.2 & \textbf{2.9} & \cellcolor{expTwoBetter}$\times$2.5 \\
\addlinespace
\rowcolor{expTwoSection}
 & \textit{Lower} & \multicolumn{3}{c}{\textit{$\approx$7$\,\times$ compression ($P_{\mathrm{sum},2}$)}} & \multicolumn{3}{c}{\textit{$\approx$6$\,\times$ ($P_{\mathrm{sum},2}$)}} & \multicolumn{3}{c}{\textit{$\approx$7$\,\times$ ($P_{\mathrm{sum},2}$)}} \\
Score &  & 47.5 & \textbf{62.7} & \cellcolor{expTwoBetter}+15.2 & 49.8 & \textbf{68.8} & \cellcolor{expTwoBetter}+19.0 & 37.5 & \textbf{59.5} & \cellcolor{expTwoBetter}+21.9 \\
Tokens &  & 207 & \textbf{188} & \cellcolor{expTwoBetter}$-$9.3\% & 235 & \textbf{231} & \cellcolor{expTwoBetter}$-$1.7\% & 209 & \textbf{198} & \cellcolor{expTwoBetter}$-$5.3\% \\
$\mathcal{S}$ GPU (ms) &  & 6,631 & \textbf{1,090} & \cellcolor{expTwoBetter}$\times$6.1 & 5,162 & \textbf{1,637} & \cellcolor{expTwoBetter}$\times$3.2 & 13,511 & \textbf{1,471} & \cellcolor{expTwoBetter}$\times$9.2 \\
$\mathcal{R}$ TTFT (ms) &  & 12.9 & \textbf{3.6} & \cellcolor{expTwoBetter}$\times$3.5 & 27.3 & \textbf{6.5} & \cellcolor{expTwoBetter}$\times$4.2 & 9.3 & \textbf{3.0} & \cellcolor{expTwoBetter}$\times$3.1 \\
\addlinespace
\rowcolor{expTwoSection}
 & \textit{Low} & \multicolumn{3}{c}{\textit{$\approx$5$\,\times$ compression ($P_{\mathrm{sum},3}$)}} & \multicolumn{3}{c}{\textit{$\approx$5$\,\times$ ($P_{\mathrm{sum},3}$)}} & \multicolumn{3}{c}{\textit{$\approx$5$\,\times$ ($P_{\mathrm{sum},3}$)}} \\
Score &  & 51.8 & \textbf{66.5} & \cellcolor{expTwoBetter}+14.7 & 55.9 & \textbf{71.4} & \cellcolor{expTwoBetter}+15.5 & 45.3 & \textbf{62.4} & \cellcolor{expTwoBetter}+17.1 \\
Tokens &  & 297 & \textbf{263} & \cellcolor{expTwoBetter}$-$11.2\% & 335 & \textbf{277} & \cellcolor{expTwoBetter}$-$17.4\% & 344 & \textbf{277} & \cellcolor{expTwoBetter}$-$19.5\% \\
$\mathcal{S}$ GPU (ms) &  & 9,132 & \textbf{1,315} & \cellcolor{expTwoBetter}$\times$6.9 & 7,403 & \textbf{1,790} & \cellcolor{expTwoBetter}$\times$4.1 & 22,165 & \textbf{1,821} & \cellcolor{expTwoBetter}$\times$12.2 \\
$\mathcal{R}$ TTFT (ms) &  & 17.2 & \textbf{3.7} & \cellcolor{expTwoBetter}$\times$4.6 & 36.6 & \textbf{6.6} & \cellcolor{expTwoBetter}$\times$5.6 & 13.7 & \textbf{3.0} & \cellcolor{expTwoBetter}$\times$4.5 \\
\addlinespace
\rowcolor{expTwoSection}
 & \textit{Medium-low} & \multicolumn{3}{c}{\textit{$\approx$3$\,\times$ compression ($P_{\mathrm{sum},5}$)}} & \multicolumn{3}{c}{\textit{$\approx$3$\,\times$ ($P_{\mathrm{sum},5}$)}} & \multicolumn{3}{c}{\textit{$\approx$2$\,\times$ ($P_{\mathrm{sum},4}$)}} \\
Score &  & 56.5 & \textbf{67.4} & \cellcolor{expTwoBetter}+10.9 & 59.4 & \textbf{74.9} & \cellcolor{expTwoBetter}+15.5 & 59.2 & \textbf{68.5} & \cellcolor{expTwoBetter}+9.3 \\
Tokens &  & 628 & \textbf{439} & \cellcolor{expTwoBetter}$-$30.0\% & 499 & \textbf{461} & \cellcolor{expTwoBetter}$-$7.6\% & 767 & \textbf{692} & \cellcolor{expTwoBetter}$-$9.7\% \\
$\mathcal{S}$ GPU (ms) &  & 18,038 & \textbf{2,325} & \cellcolor{expTwoBetter}$\times$7.8 & 10,604 & \textbf{2,805} & \cellcolor{expTwoBetter}$\times$3.8 & 47,056 & \textbf{4,561} & \cellcolor{expTwoBetter}$\times$10.3 \\
$\mathcal{R}$ TTFT (ms) &  & 30.3 & \textbf{3.8} & \cellcolor{expTwoBetter}$\times$8.0 & 50.0 & \textbf{6.7} & \cellcolor{expTwoBetter}$\times$7.4 & 27.2 & \textbf{3.3} & \cellcolor{expTwoBetter}$\times$8.4 \\
\addlinespace
\rowcolor{expTwoSection}
 & \textit{Medium} & \multicolumn{3}{c}{\textit{$\approx$2$\,\times$ compression ($P_{\mathrm{sum},4}$)}} & \multicolumn{3}{c}{\textit{$\approx$3$\,\times$ ($P_{\mathrm{sum},4}$)}} & \multicolumn{3}{c}{\textit{$\approx$2$\,\times$ ($P_{\mathrm{sum},5}$)}} \\
Score &  & 61.2 & \textbf{67.9} & \cellcolor{expTwoBetter}+6.7 & 63.4 & \textbf{74.9} & \cellcolor{expTwoBetter}+11.4 & 55.1 & \textbf{68.5} & \cellcolor{expTwoBetter}+13.4 \\
Tokens &  & 789 & \textbf{659} & \cellcolor{expTwoBetter}$-$16.5\% & 542 & \textbf{461} & \cellcolor{expTwoBetter}$-$14.9\% & 921 & \textbf{692} & \cellcolor{expTwoBetter}$-$24.8\% \\
$\mathcal{S}$ GPU (ms) &  & 23,356 & \textbf{3,847} & \cellcolor{expTwoBetter}$\times$6.1 & 12,119 & \textbf{2,805} & \cellcolor{expTwoBetter}$\times$4.3 & 58,826 & \textbf{4,561} & \cellcolor{expTwoBetter}$\times$12.9 \\
$\mathcal{R}$ TTFT (ms) &  & 37.2 & \textbf{3.9} & \cellcolor{expTwoBetter}$\times$9.5 & 54.6 & \textbf{6.7} & \cellcolor{expTwoBetter}$\times$8.1 & 33.3 & \textbf{3.3} & \cellcolor{expTwoBetter}$\times$10.2 \\
\bottomrule
\end{tabular}
\endgroup
\end{table*}

\begin{table*}[t]
\caption{
Experiment 2, LongHealth. \ttot{}: sender summarizes to receiver. \swat{}+AM: the sender's cache rendered and compressed with Attention Matching. Budgets are determined by the sender agent in \ttot{} given one of five summary prompts and sorted from lowest to highest per pair. Change: score difference in points, relative token difference, and speedup.
}
\label{tab:exp2-appendix-longhealth}
\centering
\begingroup
\definecolor{expTwoSection}{HTML}{ECECEC}
\definecolor{expTwoBetter}{HTML}{D9EFD8}
\begin{tabular}{llrrrrrrrrr}
\toprule
 & & \multicolumn{3}{c}{Qwen3-4B $\rightarrow$ Llama3-8B} & \multicolumn{3}{c}{Llama3-8B $\rightarrow$ Qwen3-14B} & \multicolumn{3}{c}{Gemma3-12B $\rightarrow$ Qwen3-4B} \\
\cmidrule(lr){3-5}\cmidrule(lr){6-8}\cmidrule(lr){9-11}
Metric & Budget & \ttot{} & \swat{} & Change & \ttot{} & \swat{} & Change & \ttot{} & \swat{} & Change \\
\midrule
\rowcolor{expTwoSection}
 & \textit{Lowest} & \multicolumn{3}{c}{\textit{$\approx$58$\,\times$ compression ($P_{\mathrm{sum},1}$)}} & \multicolumn{3}{c}{\textit{$\approx$26$\,\times$ ($P_{\mathrm{sum},3}$)}} & \multicolumn{3}{c}{\textit{$\approx$39$\,\times$ ($P_{\mathrm{sum},1}$)}} \\
Score &  & 28.7 & \textbf{37.2} & \cellcolor{expTwoBetter}+8.5 & 39.0 & \textbf{47.8} & \cellcolor{expTwoBetter}+8.8 & 30.8 & \textbf{40.5} & \cellcolor{expTwoBetter}+9.8 \\
Tokens &  & 130 & \textbf{129} & \cellcolor{expTwoBetter}$-$0.5\% & 308 & \textbf{303} & \cellcolor{expTwoBetter}$-$1.5\% & 203 & \textbf{202} & \cellcolor{expTwoBetter}$-$0.4\% \\
$\mathcal{S}$ GPU (ms) &  & 4,299 & \textbf{3,410} & \cellcolor{expTwoBetter}$\times$1.3 & 7,481 & \textbf{4,840} & \cellcolor{expTwoBetter}$\times$1.5 & 13,791 & \textbf{5,567} & \cellcolor{expTwoBetter}$\times$2.5 \\
$\mathcal{R}$ TTFT (ms) &  & 11.7 & \textbf{5.9} & \cellcolor{expTwoBetter}$\times$2.0 & 38.0 & \textbf{12.0} & \cellcolor{expTwoBetter}$\times$3.2 & 11.2 & \textbf{5.1} & \cellcolor{expTwoBetter}$\times$2.2 \\
\addlinespace
\rowcolor{expTwoSection}
 & \textit{Lower} & \multicolumn{3}{c}{\textit{$\approx$34$\,\times$ compression ($P_{\mathrm{sum},2}$)}} & \multicolumn{3}{c}{\textit{$\approx$22$\,\times$ ($P_{\mathrm{sum},1}$)}} & \multicolumn{3}{c}{\textit{$\approx$25$\,\times$ ($P_{\mathrm{sum},2}$)}} \\
Score &  & 31.0 & \textbf{38.5} & \cellcolor{expTwoBetter}+7.5 & 35.5 & \textbf{49.2} & \cellcolor{expTwoBetter}+13.8 & 32.0 & \textbf{45.2} & \cellcolor{expTwoBetter}+13.2 \\
Tokens &  & 224 & \textbf{221} & \cellcolor{expTwoBetter}$-$1.7\% & 363 & \textbf{358} & \cellcolor{expTwoBetter}$-$1.3\% & 326 & \textbf{315} & \cellcolor{expTwoBetter}$-$3.3\% \\
$\mathcal{S}$ GPU (ms) &  & 7,272 & \textbf{3,680} & \cellcolor{expTwoBetter}$\times$2.0 & 8,615 & \textbf{5,004} & \cellcolor{expTwoBetter}$\times$1.7 & 22,103 & \textbf{5,975} & \cellcolor{expTwoBetter}$\times$3.7 \\
$\mathcal{R}$ TTFT (ms) &  & 16.4 & \textbf{6.0} & \cellcolor{expTwoBetter}$\times$2.7 & 42.9 & \textbf{12.1} & \cellcolor{expTwoBetter}$\times$3.6 & 15.5 & \textbf{5.2} & \cellcolor{expTwoBetter}$\times$3.0 \\
\addlinespace
\rowcolor{expTwoSection}
 & \textit{Low} & \multicolumn{3}{c}{\textit{$\approx$16$\,\times$ compression ($P_{\mathrm{sum},3}$)}} & \multicolumn{3}{c}{\textit{$\approx$22$\,\times$ ($P_{\mathrm{sum},2}$)}} & \multicolumn{3}{c}{\textit{$\approx$10$\,\times$ ($P_{\mathrm{sum},3}$)}} \\
Score &  & 36.0 & \textbf{44.8} & \cellcolor{expTwoBetter}+8.8 & 38.8 & \textbf{49.2} & \cellcolor{expTwoBetter}+10.5 & 40.0 & \textbf{54.0} & \cellcolor{expTwoBetter}+14.0 \\
Tokens &  & 475 & \textbf{469} & \cellcolor{expTwoBetter}$-$1.3\% & 370 & \textbf{358} & \cellcolor{expTwoBetter}$-$3.1\% & 798 & \textbf{788} & \cellcolor{expTwoBetter}$-$1.3\% \\
$\mathcal{S}$ GPU (ms) &  & 14,697 & \textbf{4,465} & \cellcolor{expTwoBetter}$\times$3.3 & 8,876 & \textbf{5,004} & \cellcolor{expTwoBetter}$\times$1.8 & 51,790 & \textbf{7,802} & \cellcolor{expTwoBetter}$\times$6.6 \\
$\mathcal{R}$ TTFT (ms) &  & 27.3 & \textbf{6.2} & \cellcolor{expTwoBetter}$\times$4.4 & 43.8 & \textbf{12.1} & \cellcolor{expTwoBetter}$\times$3.6 & 31.5 & \textbf{5.6} & \cellcolor{expTwoBetter}$\times$5.6 \\
\addlinespace
\rowcolor{expTwoSection}
 & \textit{Medium-low} & \multicolumn{3}{c}{\textit{$\approx$7$\,\times$ compression ($P_{\mathrm{sum},4}$)}} & \multicolumn{3}{c}{\textit{$\approx$18$\,\times$ ($P_{\mathrm{sum},5}$)}} & \multicolumn{3}{c}{\textit{$\approx$6$\,\times$ ($P_{\mathrm{sum},5}$)}} \\
Score &  & 43.5 & \textbf{53.5} & \cellcolor{expTwoBetter}+10.0 & 37.2 & \textbf{48.8} & \cellcolor{expTwoBetter}+11.5 & 49.0 & \textbf{58.5} & \cellcolor{expTwoBetter}+9.5 \\
Tokens &  & 1,194 & \textbf{1,071} & \cellcolor{expTwoBetter}$-$10.3\% & 439 & \textbf{438} & \cellcolor{expTwoBetter}$-$0.4\% & 1,426 & \textbf{1,313} & \cellcolor{expTwoBetter}$-$7.9\% \\
$\mathcal{S}$ GPU (ms) &  & 36,800 & \textbf{6,369} & \cellcolor{expTwoBetter}$\times$5.8 & 10,281 & \textbf{5,391} & \cellcolor{expTwoBetter}$\times$1.9 & 90,082 & \textbf{9,704} & \cellcolor{expTwoBetter}$\times$9.3 \\
$\mathcal{R}$ TTFT (ms) &  & 59.9 & \textbf{6.6} & \cellcolor{expTwoBetter}$\times$9.0 & 48.5 & \textbf{12.2} & \cellcolor{expTwoBetter}$\times$4.0 & 52.4 & \textbf{6.0} & \cellcolor{expTwoBetter}$\times$8.7 \\
\addlinespace
\rowcolor{expTwoSection}
 & \textit{Medium} & \multicolumn{3}{c}{\textit{$\approx$6$\,\times$ compression ($P_{\mathrm{sum},5}$)}} & \multicolumn{3}{c}{\textit{$\approx$15$\,\times$ ($P_{\mathrm{sum},4}$)}} & \multicolumn{3}{c}{\textit{$\approx$6$\,\times$ ($P_{\mathrm{sum},4}$)}} \\
Score &  & 45.0 & \textbf{55.0} & \cellcolor{expTwoBetter}+10.0 & 40.2 & \textbf{51.7} & \cellcolor{expTwoBetter}+11.5 & 49.0 & \textbf{58.5} & \cellcolor{expTwoBetter}+9.5 \\
Tokens &  & 1,323 & \textbf{1,250} & \cellcolor{expTwoBetter}$-$5.5\% & 542 & \textbf{525} & \cellcolor{expTwoBetter}$-$3.1\% & 1,469 & \textbf{1,313} & \cellcolor{expTwoBetter}$-$10.6\% \\
$\mathcal{S}$ GPU (ms) &  & 40,319 & \textbf{7,095} & \cellcolor{expTwoBetter}$\times$5.7 & 12,653 & \textbf{5,640} & \cellcolor{expTwoBetter}$\times$2.2 & 93,712 & \textbf{9,704} & \cellcolor{expTwoBetter}$\times$9.7 \\
$\mathcal{R}$ TTFT (ms) &  & 66.4 & \textbf{6.7} & \cellcolor{expTwoBetter}$\times$9.9 & 58.5 & \textbf{12.3} & \cellcolor{expTwoBetter}$\times$4.8 & 54.1 & \textbf{6.0} & \cellcolor{expTwoBetter}$\times$9.0 \\
\bottomrule
\end{tabular}
\endgroup
\end{table*}

\begin{table*}[t]
\caption{
Experiment 2, MuSiQue. \ttot{}: sender summarizes to receiver. \swat{}+AM: the sender's cache rendered and compressed with Attention Matching. Budgets are determined by the sender agent in \ttot{} given one of five summary prompts and sorted from lowest to highest per pair. Change: score difference in points, relative token difference, and speedup.
}
\label{tab:exp2-appendix-musique}
\centering
\begingroup
\definecolor{expTwoSection}{HTML}{ECECEC}
\definecolor{expTwoBetter}{HTML}{D9EFD8}
\begin{tabular}{llrrrrrrrrr}
\toprule
 & & \multicolumn{3}{c}{Qwen3-4B $\rightarrow$ Llama3-8B} & \multicolumn{3}{c}{Llama3-8B $\rightarrow$ Qwen3-14B} & \multicolumn{3}{c}{Gemma3-12B $\rightarrow$ Qwen3-4B} \\
\cmidrule(lr){3-5}\cmidrule(lr){6-8}\cmidrule(lr){9-11}
Metric & Budget & \ttot{} & \swat{} & Change & \ttot{} & \swat{} & Change & \ttot{} & \swat{} & Change \\
\midrule
\rowcolor{expTwoSection}
 & \textit{Lowest} & \multicolumn{3}{c}{\textit{$\approx$14$\,\times$ compression ($P_{\mathrm{sum},1}$)}} & \multicolumn{3}{c}{\textit{$\approx$10$\,\times$ ($P_{\mathrm{sum},1}$)}} & \multicolumn{3}{c}{\textit{$\approx$15$\,\times$ ($P_{\mathrm{sum},1}$)}} \\
Score &  & 11.8 & \textbf{24.0} & \cellcolor{expTwoBetter}+12.2 & 11.8 & \textbf{32.6} & \cellcolor{expTwoBetter}+20.8 & 8.2 & \textbf{20.0} & \cellcolor{expTwoBetter}+11.7 \\
Tokens &  & 168 & 168 & $-$0.2\% & 254 & \textbf{248} & \cellcolor{expTwoBetter}$-$2.3\% & 171 & \textbf{166} & \cellcolor{expTwoBetter}$-$3.4\% \\
$\mathcal{S}$ GPU (ms) &  & 5,287 & \textbf{1,568} & \cellcolor{expTwoBetter}$\times$3.4 & 5,722 & \textbf{2,290} & \cellcolor{expTwoBetter}$\times$2.5 & 10,607 & \textbf{2,441} & \cellcolor{expTwoBetter}$\times$4.3 \\
$\mathcal{R}$ TTFT (ms) &  & 9.9 & \textbf{3.7} & \cellcolor{expTwoBetter}$\times$2.7 & 27.0 & \textbf{6.6} & \cellcolor{expTwoBetter}$\times$4.1 & 7.8 & \textbf{2.9} & \cellcolor{expTwoBetter}$\times$2.6 \\
\addlinespace
\rowcolor{expTwoSection}
 & \textit{Lower} & \multicolumn{3}{c}{\textit{$\approx$8$\,\times$ compression ($P_{\mathrm{sum},2}$)}} & \multicolumn{3}{c}{\textit{$\approx$8$\,\times$ ($P_{\mathrm{sum},2}$)}} & \multicolumn{3}{c}{\textit{$\approx$11$\,\times$ ($P_{\mathrm{sum},2}$)}} \\
Score &  & 13.0 & \textbf{31.5} & \cellcolor{expTwoBetter}+18.5 & 15.1 & \textbf{36.6} & \cellcolor{expTwoBetter}+21.5 & 8.4 & \textbf{21.0} & \cellcolor{expTwoBetter}+12.7 \\
Tokens &  & 305 & \textbf{294} & \cellcolor{expTwoBetter}$-$3.5\% & 346 & \textbf{310} & \cellcolor{expTwoBetter}$-$10.4\% & 230 & \textbf{226} & \cellcolor{expTwoBetter}$-$1.9\% \\
$\mathcal{S}$ GPU (ms) &  & 8,877 & \textbf{1,977} & \cellcolor{expTwoBetter}$\times$4.5 & 6,361 & \textbf{2,549} & \cellcolor{expTwoBetter}$\times$2.5 & 14,849 & \textbf{2,457} & \cellcolor{expTwoBetter}$\times$6.0 \\
$\mathcal{R}$ TTFT (ms) &  & 17.0 & \textbf{3.7} & \cellcolor{expTwoBetter}$\times$4.6 & 29.9 & \textbf{6.6} & \cellcolor{expTwoBetter}$\times$4.5 & 10.1 & \textbf{3.0} & \cellcolor{expTwoBetter}$\times$3.4 \\
\addlinespace
\rowcolor{expTwoSection}
 & \textit{Low} & \multicolumn{3}{c}{\textit{$\approx$7$\,\times$ compression ($P_{\mathrm{sum},3}$)}} & \multicolumn{3}{c}{\textit{$\approx$6$\,\times$ ($P_{\mathrm{sum},3}$)}} & \multicolumn{3}{c}{\textit{$\approx$6$\,\times$ ($P_{\mathrm{sum},3}$)}} \\
Score &  & 15.0 & \textbf{32.1} & \cellcolor{expTwoBetter}+17.1 & 16.9 & \textbf{39.8} & \cellcolor{expTwoBetter}+22.9 & 13.2 & \textbf{27.5} & \cellcolor{expTwoBetter}+14.3 \\
Tokens &  & 377 & \textbf{336} & \cellcolor{expTwoBetter}$-$10.7\% & 431 & \textbf{414} & \cellcolor{expTwoBetter}$-$3.9\% & 456 & \textbf{414} & \cellcolor{expTwoBetter}$-$9.2\% \\
$\mathcal{S}$ GPU (ms) &  & 9,464 & \textbf{2,070} & \cellcolor{expTwoBetter}$\times$4.6 & 9,174 & \textbf{3,059} & \cellcolor{expTwoBetter}$\times$3.0 & 29,586 & \textbf{3,199} & \cellcolor{expTwoBetter}$\times$9.2 \\
$\mathcal{R}$ TTFT (ms) &  & 19.0 & \textbf{3.8} & \cellcolor{expTwoBetter}$\times$5.0 & 41.1 & \textbf{6.8} & \cellcolor{expTwoBetter}$\times$6.1 & 17.2 & \textbf{3.2} & \cellcolor{expTwoBetter}$\times$5.5 \\
\addlinespace
\rowcolor{expTwoSection}
 & \textit{Medium-low} & \multicolumn{3}{c}{\textit{$\approx$3$\,\times$ compression ($P_{\mathrm{sum},5}$)}} & \multicolumn{3}{c}{\textit{$\approx$4$\,\times$ ($P_{\mathrm{sum},5}$)}} & \multicolumn{3}{c}{\textit{$\approx$3$\,\times$ ($P_{\mathrm{sum},4}$)}} \\
Score &  & 14.7 & \textbf{37.6} & \cellcolor{expTwoBetter}+22.9 & 17.6 & \textbf{45.7} & \cellcolor{expTwoBetter}+28.1 & 19.1 & \textbf{34.5} & \cellcolor{expTwoBetter}+15.4 \\
Tokens &  & 872 & \textbf{785} & \cellcolor{expTwoBetter}$-$10.0\% & 622 & \textbf{621} & \cellcolor{expTwoBetter}$-$0.3\% & 1,000 & \textbf{828} & \cellcolor{expTwoBetter}$-$17.2\% \\
$\mathcal{S}$ GPU (ms) &  & 29,256 & \textbf{4,519} & \cellcolor{expTwoBetter}$\times$6.5 & 13,224 & \textbf{4,078} & \cellcolor{expTwoBetter}$\times$3.2 & 68,038 & \textbf{5,578} & \cellcolor{expTwoBetter}$\times$12.2 \\
$\mathcal{R}$ TTFT (ms) &  & 46.3 & \textbf{4.0} & \cellcolor{expTwoBetter}$\times$11.6 & 54.3 & \textbf{6.9} & \cellcolor{expTwoBetter}$\times$7.8 & 38.5 & \textbf{3.4} & \cellcolor{expTwoBetter}$\times$11.3 \\
\addlinespace
\rowcolor{expTwoSection}
 & \textit{Medium} & \multicolumn{3}{c}{\textit{$\approx$3$\,\times$ compression ($P_{\mathrm{sum},4}$)}} & \multicolumn{3}{c}{\textit{$\approx$4$\,\times$ ($P_{\mathrm{sum},4}$)}} & \multicolumn{3}{c}{\textit{$\approx$3$\,\times$ ($P_{\mathrm{sum},5}$)}} \\
Score &  & 23.4 & \textbf{37.6} & \cellcolor{expTwoBetter}+14.2 & 20.7 & \textbf{45.7} & \cellcolor{expTwoBetter}+25.0 & 18.7 & \textbf{34.5} & \cellcolor{expTwoBetter}+15.8 \\
Tokens &  & 1,143 & \textbf{785} & \cellcolor{expTwoBetter}$-$31.4\% & 635 & \textbf{621} & \cellcolor{expTwoBetter}$-$2.2\% & 1,098 & \textbf{828} & \cellcolor{expTwoBetter}$-$24.7\% \\
$\mathcal{S}$ GPU (ms) &  & 38,766 & \textbf{4,519} & \cellcolor{expTwoBetter}$\times$8.6 & 14,258 & \textbf{4,078} & \cellcolor{expTwoBetter}$\times$3.5 & 73,704 & \textbf{5,578} & \cellcolor{expTwoBetter}$\times$13.2 \\
$\mathcal{R}$ TTFT (ms) &  & 57.9 & \textbf{4.0} & \cellcolor{expTwoBetter}$\times$14.5 & 61.2 & \textbf{6.9} & \cellcolor{expTwoBetter}$\times$8.8 & 41.2 & \textbf{3.4} & \cellcolor{expTwoBetter}$\times$12.1 \\
\bottomrule
\end{tabular}
\endgroup
\end{table*}

\begin{table*}[t]
\caption{
Experiment 2, NarrativeQA. \ttot{}: sender summarizes to receiver. \swat{}+AM: the sender's cache rendered and compressed with Attention Matching. Budgets are determined by the sender agent in \ttot{} given one of five summary prompts and sorted from lowest to highest per pair. Change: score difference in points, relative token difference, and speedup.
}
\label{tab:exp2-appendix-narrativeqa}
\centering
\begingroup
\definecolor{expTwoSection}{HTML}{ECECEC}
\definecolor{expTwoBetter}{HTML}{D9EFD8}
\begin{tabular}{llrrrrrrrrr}
\toprule
 & & \multicolumn{3}{c}{Qwen3-4B $\rightarrow$ Llama3-8B} & \multicolumn{3}{c}{Llama3-8B $\rightarrow$ Qwen3-14B} & \multicolumn{3}{c}{Gemma3-12B $\rightarrow$ Qwen3-4B} \\
\cmidrule(lr){3-5}\cmidrule(lr){6-8}\cmidrule(lr){9-11}
Metric & Budget & \ttot{} & \swat{} & Change & \ttot{} & \swat{} & Change & \ttot{} & \swat{} & Change \\
\midrule
\rowcolor{expTwoSection}
 & \textit{Lowest} & \multicolumn{3}{c}{\textit{$\approx$7$\,\times$ compression ($P_{\mathrm{sum},1}$)}} & \multicolumn{3}{c}{\textit{$\approx$7$\,\times$ ($P_{\mathrm{sum},1}$)}} & \multicolumn{3}{c}{\textit{$\approx$9$\,\times$ ($P_{\mathrm{sum},1}$)}} \\
Score &  & 29.2 & \textbf{65.6} & \cellcolor{expTwoBetter}+36.4 & 29.8 & \textbf{66.7} & \cellcolor{expTwoBetter}+36.9 & 19.1 & \textbf{53.1} & \cellcolor{expTwoBetter}+34.0 \\
Tokens &  & 108 & \textbf{103} & \cellcolor{expTwoBetter}$-$4.9\% & 115 & \textbf{103} & \cellcolor{expTwoBetter}$-$10.7\% & 84 & \textbf{80} & \cellcolor{expTwoBetter}$-$4.3\% \\
$\mathcal{S}$ GPU (ms) &  & 3,286 & \textbf{1,085} & \cellcolor{expTwoBetter}$\times$3.0 & 2,806 & \textbf{1,023} & \cellcolor{expTwoBetter}$\times$2.7 & 5,590 & \textbf{1,083} & \cellcolor{expTwoBetter}$\times$5.2 \\
$\mathcal{R}$ TTFT (ms) &  & 8.3 & \textbf{3.0} & \cellcolor{expTwoBetter}$\times$2.8 & 16.8 & \textbf{5.9} & \cellcolor{expTwoBetter}$\times$2.8 & 5.2 & \textbf{2.6} & \cellcolor{expTwoBetter}$\times$2.0 \\
\addlinespace
\rowcolor{expTwoSection}
 & \textit{Lower} & \multicolumn{3}{c}{\textit{$\approx$5$\,\times$ compression ($P_{\mathrm{sum},2}$)}} & \multicolumn{3}{c}{\textit{$\approx$5$\,\times$ ($P_{\mathrm{sum},2}$)}} & \multicolumn{3}{c}{\textit{$\approx$7$\,\times$ ($P_{\mathrm{sum},2}$)}} \\
Score &  & 35.3 & \textbf{66.9} & \cellcolor{expTwoBetter}+31.6 & 34.2 & \textbf{69.2} & \cellcolor{expTwoBetter}+35.0 & 22.9 & \textbf{56.6} & \cellcolor{expTwoBetter}+33.6 \\
Tokens &  & 155 & \textbf{144} & \cellcolor{expTwoBetter}$-$7.3\% & 151 & \textbf{144} & \cellcolor{expTwoBetter}$-$4.3\% & 116 & \textbf{103} & \cellcolor{expTwoBetter}$-$11.1\% \\
$\mathcal{S}$ GPU (ms) &  & 4,632 & \textbf{1,084} & \cellcolor{expTwoBetter}$\times$4.3 & 3,428 & \textbf{1,214} & \cellcolor{expTwoBetter}$\times$2.8 & 7,500 & \textbf{1,145} & \cellcolor{expTwoBetter}$\times$6.5 \\
$\mathcal{R}$ TTFT (ms) &  & 10.0 & \textbf{3.0} & \cellcolor{expTwoBetter}$\times$3.3 & 18.8 & \textbf{6.0} & \cellcolor{expTwoBetter}$\times$3.1 & 6.4 & \textbf{2.6} & \cellcolor{expTwoBetter}$\times$2.5 \\
\addlinespace
\rowcolor{expTwoSection}
 & \textit{Low} & \multicolumn{3}{c}{\textit{$\approx$4$\,\times$ compression ($P_{\mathrm{sum},3}$)}} & \multicolumn{3}{c}{\textit{$\approx$4$\,\times$ ($P_{\mathrm{sum},3}$)}} & \multicolumn{3}{c}{\textit{$\approx$4$\,\times$ ($P_{\mathrm{sum},3}$)}} \\
Score &  & 43.2 & \textbf{68.2} & \cellcolor{expTwoBetter}+25.0 & 38.2 & \textbf{70.0} & \cellcolor{expTwoBetter}+31.8 & 29.0 & \textbf{63.1} & \cellcolor{expTwoBetter}+34.1 \\
Tokens &  & 226 & \textbf{180} & \cellcolor{expTwoBetter}$-$20.3\% & 203 & \textbf{181} & \cellcolor{expTwoBetter}$-$11.0\% & 200 & \textbf{181} & \cellcolor{expTwoBetter}$-$9.9\% \\
$\mathcal{S}$ GPU (ms) &  & 7,466 & \textbf{1,324} & \cellcolor{expTwoBetter}$\times$5.6 & 5,085 & \textbf{1,379} & \cellcolor{expTwoBetter}$\times$3.7 & 13,666 & \textbf{1,457} & \cellcolor{expTwoBetter}$\times$9.4 \\
$\mathcal{R}$ TTFT (ms) &  & 14.6 & \textbf{3.0} & \cellcolor{expTwoBetter}$\times$4.8 & 25.0 & \textbf{6.0} & \cellcolor{expTwoBetter}$\times$4.2 & 9.5 & \textbf{2.6} & \cellcolor{expTwoBetter}$\times$3.6 \\
\addlinespace
\rowcolor{expTwoSection}
 & \textit{Medium-low} & \multicolumn{3}{c}{\textit{$\approx$2$\,\times$ compression ($P_{\mathrm{sum},4}$)}} & \multicolumn{3}{c}{\textit{$\approx$2$\,\times$ ($P_{\mathrm{sum},4}$)}} & \multicolumn{3}{c}{\textit{$\approx$2$\,\times$ ($P_{\mathrm{sum},4}$)}} \\
Score &  & 56.5 & \textbf{70.0} & \cellcolor{expTwoBetter}+13.5 & 49.5 & \textbf{71.6} & \cellcolor{expTwoBetter}+22.2 & 46.5 & \textbf{66.7} & \cellcolor{expTwoBetter}+20.2 \\
Tokens &  & 486 & \textbf{360} & \cellcolor{expTwoBetter}$-$26.0\% & 401 & \textbf{361} & \cellcolor{expTwoBetter}$-$9.9\% & 536 & \textbf{361} & \cellcolor{expTwoBetter}$-$32.6\% \\
$\mathcal{S}$ GPU (ms) &  & 15,394 & \textbf{2,722} & \cellcolor{expTwoBetter}$\times$5.7 & 10,329 & \textbf{2,671} & \cellcolor{expTwoBetter}$\times$3.9 & 35,625 & \textbf{2,554} & \cellcolor{expTwoBetter}$\times$13.9 \\
$\mathcal{R}$ TTFT (ms) &  & 26.7 & \textbf{3.1} & \cellcolor{expTwoBetter}$\times$8.6 & 44.2 & \textbf{6.2} & \cellcolor{expTwoBetter}$\times$7.1 & 21.0 & \textbf{2.8} & \cellcolor{expTwoBetter}$\times$7.6 \\
\addlinespace
\rowcolor{expTwoSection}
 & \textit{Medium} & \multicolumn{3}{c}{\textit{$\approx$2$\,\times$ compression ($P_{\mathrm{sum},5}$)}} & \multicolumn{3}{c}{\textit{$\approx$2$\,\times$ ($P_{\mathrm{sum},5}$)}} & \multicolumn{3}{c}{\textit{$\approx$2$\,\times$ ($P_{\mathrm{sum},5}$)}} \\
Score &  & 55.7 & \textbf{70.0} & \cellcolor{expTwoBetter}+14.4 & 47.2 & \textbf{71.6} & \cellcolor{expTwoBetter}+24.4 & 40.7 & \textbf{66.7} & \cellcolor{expTwoBetter}+26.0 \\
Tokens &  & 537 & \textbf{360} & \cellcolor{expTwoBetter}$-$33.0\% & 416 & \textbf{361} & \cellcolor{expTwoBetter}$-$13.3\% & 693 & \textbf{361} & \cellcolor{expTwoBetter}$-$47.9\% \\
$\mathcal{S}$ GPU (ms) &  & 16,040 & \textbf{2,722} & \cellcolor{expTwoBetter}$\times$5.9 & 10,028 & \textbf{2,671} & \cellcolor{expTwoBetter}$\times$3.8 & 46,570 & \textbf{2,554} & \cellcolor{expTwoBetter}$\times$18.2 \\
$\mathcal{R}$ TTFT (ms) &  & 27.5 & \textbf{3.1} & \cellcolor{expTwoBetter}$\times$8.9 & 43.3 & \textbf{6.2} & \cellcolor{expTwoBetter}$\times$7.0 & 26.9 & \textbf{2.8} & \cellcolor{expTwoBetter}$\times$9.8 \\
\bottomrule
\end{tabular}
\endgroup
\end{table*}

\begin{table*}[t]
\caption{
Experiment 2, QASPER. \ttot{}: sender summarizes to receiver. \swat{}+AM: the sender's cache rendered and compressed with Attention Matching. Budgets are determined by the sender agent in \ttot{} given one of five summary prompts and sorted from lowest to highest per pair. Change: score difference in points, relative token difference, and speedup.
}
\label{tab:exp2-appendix-qasper}
\centering
\begingroup
\definecolor{expTwoSection}{HTML}{ECECEC}
\definecolor{expTwoBetter}{HTML}{D9EFD8}
\begin{tabular}{llrrrrrrrrr}
\toprule
 & & \multicolumn{3}{c}{Qwen3-4B $\rightarrow$ Llama3-8B} & \multicolumn{3}{c}{Llama3-8B $\rightarrow$ Qwen3-14B} & \multicolumn{3}{c}{Gemma3-12B $\rightarrow$ Qwen3-4B} \\
\cmidrule(lr){3-5}\cmidrule(lr){6-8}\cmidrule(lr){9-11}
Metric & Budget & \ttot{} & \swat{} & Change & \ttot{} & \swat{} & Change & \ttot{} & \swat{} & Change \\
\midrule
\rowcolor{expTwoSection}
 & \textit{Lowest} & \multicolumn{3}{c}{\textit{$\approx$41$\,\times$ compression ($P_{\mathrm{sum},1}$)}} & \multicolumn{3}{c}{\textit{$\approx$27$\,\times$ ($P_{\mathrm{sum},1}$)}} & \multicolumn{3}{c}{\textit{$\approx$41$\,\times$ ($P_{\mathrm{sum},1}$)}} \\
Score &  & 21.7 & \textbf{34.3} & \cellcolor{expTwoBetter}+12.6 & 26.5 & \textbf{36.3} & \cellcolor{expTwoBetter}+9.8 & 23.0 & \textbf{31.9} & \cellcolor{expTwoBetter}+8.9 \\
Tokens &  & 106 & 106 & +0.2\% & 165 & \textbf{164} & \cellcolor{expTwoBetter}$-$0.5\% & 108 & 108 & $-$0.2\% \\
$\mathcal{S}$ GPU (ms) &  & 3,269 & \textbf{2,125} & \cellcolor{expTwoBetter}$\times$1.5 & 3,850 & \textbf{2,781} & \cellcolor{expTwoBetter}$\times$1.4 & 6,953 & \textbf{3,337} & \cellcolor{expTwoBetter}$\times$2.1 \\
$\mathcal{R}$ TTFT (ms) &  & 8.0 & \textbf{2.9} & \cellcolor{expTwoBetter}$\times$2.8 & 19.8 & \textbf{5.5} & \cellcolor{expTwoBetter}$\times$3.6 & 5.4 & \textbf{2.4} & \cellcolor{expTwoBetter}$\times$2.2 \\
\addlinespace
\rowcolor{expTwoSection}
 & \textit{Lower} & \multicolumn{3}{c}{\textit{$\approx$26$\,\times$ compression ($P_{\mathrm{sum},2}$)}} & \multicolumn{3}{c}{\textit{$\approx$20$\,\times$ ($P_{\mathrm{sum},2}$)}} & \multicolumn{3}{c}{\textit{$\approx$29$\,\times$ ($P_{\mathrm{sum},2}$)}} \\
Score &  & 26.5 & \textbf{37.8} & \cellcolor{expTwoBetter}+11.4 & 29.6 & \textbf{38.2} & \cellcolor{expTwoBetter}+8.6 & 25.8 & \textbf{34.7} & \cellcolor{expTwoBetter}+8.9 \\
Tokens &  & 167 & 167 & $-$0.1\% & 231 & \textbf{221} & \cellcolor{expTwoBetter}$-$4.4\% & \textbf{151} & 152 & +0.8\% \\
$\mathcal{S}$ GPU (ms) &  & 5,356 & \textbf{2,352} & \cellcolor{expTwoBetter}$\times$2.3 & 5,524 & \textbf{3,029} & \cellcolor{expTwoBetter}$\times$1.8 & 9,952 & \textbf{3,496} & \cellcolor{expTwoBetter}$\times$2.8 \\
$\mathcal{R}$ TTFT (ms) &  & 11.2 & \textbf{2.9} & \cellcolor{expTwoBetter}$\times$3.8 & 25.7 & \textbf{5.5} & \cellcolor{expTwoBetter}$\times$4.6 & 7.2 & \textbf{2.5} & \cellcolor{expTwoBetter}$\times$2.9 \\
\addlinespace
\rowcolor{expTwoSection}
 & \textit{Low} & \multicolumn{3}{c}{\textit{$\approx$16$\,\times$ compression ($P_{\mathrm{sum},3}$)}} & \multicolumn{3}{c}{\textit{$\approx$16$\,\times$ ($P_{\mathrm{sum},3}$)}} & \multicolumn{3}{c}{\textit{$\approx$15$\,\times$ ($P_{\mathrm{sum},3}$)}} \\
Score &  & 32.1 & \textbf{42.1} & \cellcolor{expTwoBetter}+10.0 & 30.5 & \textbf{39.0} & \cellcolor{expTwoBetter}+8.5 & 29.4 & \textbf{39.7} & \cellcolor{expTwoBetter}+10.4 \\
Tokens &  & 283 & \textbf{271} & \cellcolor{expTwoBetter}$-$4.0\% & 288 & \textbf{276} & \cellcolor{expTwoBetter}$-$4.0\% & 312 & \textbf{295} & \cellcolor{expTwoBetter}$-$5.5\% \\
$\mathcal{S}$ GPU (ms) &  & 8,225 & \textbf{2,674} & \cellcolor{expTwoBetter}$\times$3.1 & 7,229 & \textbf{3,277} & \cellcolor{expTwoBetter}$\times$2.2 & 19,936 & \textbf{3,995} & \cellcolor{expTwoBetter}$\times$5.0 \\
$\mathcal{R}$ TTFT (ms) &  & 15.4 & \textbf{3.0} & \cellcolor{expTwoBetter}$\times$5.1 & 32.2 & \textbf{5.6} & \cellcolor{expTwoBetter}$\times$5.8 & 12.3 & \textbf{2.6} & \cellcolor{expTwoBetter}$\times$4.7 \\
\addlinespace
\rowcolor{expTwoSection}
 & \textit{Medium-low} & \multicolumn{3}{c}{\textit{$\approx$7$\,\times$ compression ($P_{\mathrm{sum},5}$)}} & \multicolumn{3}{c}{\textit{$\approx$11$\,\times$ ($P_{\mathrm{sum},5}$)}} & \multicolumn{3}{c}{\textit{$\approx$6$\,\times$ ($P_{\mathrm{sum},4}$)}} \\
Score &  & 38.8 & \textbf{49.5} & \cellcolor{expTwoBetter}+10.7 & 32.2 & \textbf{41.3} & \cellcolor{expTwoBetter}+9.1 & 39.3 & \textbf{46.8} & \cellcolor{expTwoBetter}+7.5 \\
Tokens &  & 701 & \textbf{620} & \cellcolor{expTwoBetter}$-$11.5\% & 404 & \textbf{402} & \cellcolor{expTwoBetter}$-$0.6\% & 802 & \textbf{737} & \cellcolor{expTwoBetter}$-$8.1\% \\
$\mathcal{S}$ GPU (ms) &  & 19,936 & \textbf{3,770} & \cellcolor{expTwoBetter}$\times$5.3 & 8,799 & \textbf{3,685} & \cellcolor{expTwoBetter}$\times$2.4 & 52,999 & \textbf{5,723} & \cellcolor{expTwoBetter}$\times$9.3 \\
$\mathcal{R}$ TTFT (ms) &  & 33.2 & \textbf{3.3} & \cellcolor{expTwoBetter}$\times$10.1 & 37.6 & \textbf{5.7} & \cellcolor{expTwoBetter}$\times$6.5 & 29.0 & \textbf{3.0} & \cellcolor{expTwoBetter}$\times$9.6 \\
\addlinespace
\rowcolor{expTwoSection}
 & \textit{Medium} & \multicolumn{3}{c}{\textit{$\approx$6$\,\times$ compression ($P_{\mathrm{sum},4}$)}} & \multicolumn{3}{c}{\textit{$\approx$11$\,\times$ ($P_{\mathrm{sum},4}$)}} & \multicolumn{3}{c}{\textit{$\approx$5$\,\times$ ($P_{\mathrm{sum},5}$)}} \\
Score &  & 40.9 & \textbf{49.9} & \cellcolor{expTwoBetter}+9.1 & 33.9 & \textbf{41.3} & \cellcolor{expTwoBetter}+7.4 & 38.9 & \textbf{47.8} & \cellcolor{expTwoBetter}+8.9 \\
Tokens &  & \textbf{721} & 723 & +0.4\% & 414 & \textbf{402} & \cellcolor{expTwoBetter}$-$2.8\% & 945 & \textbf{884} & \cellcolor{expTwoBetter}$-$6.4\% \\
$\mathcal{S}$ GPU (ms) &  & 21,283 & \textbf{4,084} & \cellcolor{expTwoBetter}$\times$5.2 & 9,656 & \textbf{3,685} & \cellcolor{expTwoBetter}$\times$2.6 & 60,182 & \textbf{6,179} & \cellcolor{expTwoBetter}$\times$9.7 \\
$\mathcal{R}$ TTFT (ms) &  & 35.0 & \textbf{3.4} & \cellcolor{expTwoBetter}$\times$10.4 & 41.0 & \textbf{5.7} & \cellcolor{expTwoBetter}$\times$7.1 & 33.5 & \textbf{3.1} & \cellcolor{expTwoBetter}$\times$10.8 \\
\bottomrule
\end{tabular}
\endgroup
\end{table*}

\begin{table*}[t]
\caption{
Experiment 2, QuALITY. \ttot{}: sender summarizes to receiver. \swat{}+AM: the sender's cache rendered and compressed with Attention Matching. Budgets are determined by the sender agent in \ttot{} given one of five summary prompts and sorted from lowest to highest per pair. Change: score difference in points, relative token difference, and speedup.
}
\label{tab:exp2-appendix-quality}
\centering
\begingroup
\definecolor{expTwoSection}{HTML}{ECECEC}
\definecolor{expTwoBetter}{HTML}{D9EFD8}
\begin{tabular}{llrrrrrrrrr}
\toprule
 & & \multicolumn{3}{c}{Qwen3-4B $\rightarrow$ Llama3-8B} & \multicolumn{3}{c}{Llama3-8B $\rightarrow$ Qwen3-14B} & \multicolumn{3}{c}{Gemma3-12B $\rightarrow$ Qwen3-4B} \\
\cmidrule(lr){3-5}\cmidrule(lr){6-8}\cmidrule(lr){9-11}
Metric & Budget & \ttot{} & \swat{} & Change & \ttot{} & \swat{} & Change & \ttot{} & \swat{} & Change \\
\midrule
\rowcolor{expTwoSection}
 & \textit{Lowest} & \multicolumn{3}{c}{\textit{$\approx$42$\,\times$ compression ($P_{\mathrm{sum},1}$)}} & \multicolumn{3}{c}{\textit{$\approx$30$\,\times$ ($P_{\mathrm{sum},1}$)}} & \multicolumn{3}{c}{\textit{$\approx$52$\,\times$ ($P_{\mathrm{sum},1}$)}} \\
Score &  & 51.6 & \textbf{59.2} & \cellcolor{expTwoBetter}+7.6 & 57.8 & \textbf{71.8} & \cellcolor{expTwoBetter}+13.9 & 53.6 & \textbf{61.6} & \cellcolor{expTwoBetter}+8.0 \\
Tokens &  & 132 & \textbf{128} & \cellcolor{expTwoBetter}$-$2.5\% & 184 & \textbf{180} & \cellcolor{expTwoBetter}$-$2.0\% & 106 & \textbf{104} & \cellcolor{expTwoBetter}$-$2.3\% \\
$\mathcal{S}$ GPU (ms) &  & 3,721 & \textbf{1,476} & \cellcolor{expTwoBetter}$\times$2.5 & 4,174 & \textbf{2,085} & \cellcolor{expTwoBetter}$\times$2.0 & 7,229 & \textbf{2,208} & \cellcolor{expTwoBetter}$\times$3.3 \\
$\mathcal{R}$ TTFT (ms) &  & 11.8 & \textbf{6.0} & \cellcolor{expTwoBetter}$\times$2.0 & 27.3 & \textbf{11.6} & \cellcolor{expTwoBetter}$\times$2.3 & 8.2 & \textbf{4.9} & \cellcolor{expTwoBetter}$\times$1.7 \\
\addlinespace
\rowcolor{expTwoSection}
 & \textit{Lower} & \multicolumn{3}{c}{\textit{$\approx$27$\,\times$ compression ($P_{\mathrm{sum},2}$)}} & \multicolumn{3}{c}{\textit{$\approx$25$\,\times$ ($P_{\mathrm{sum},2}$)}} & \multicolumn{3}{c}{\textit{$\approx$40$\,\times$ ($P_{\mathrm{sum},2}$)}} \\
Score &  & 53.3 & \textbf{59.4} & \cellcolor{expTwoBetter}+6.1 & 57.5 & \textbf{73.1} & \cellcolor{expTwoBetter}+15.5 & 54.6 & \textbf{62.9} & \cellcolor{expTwoBetter}+8.3 \\
Tokens &  & 204 & \textbf{200} & \cellcolor{expTwoBetter}$-$2.4\% & 218 & \textbf{216} & \cellcolor{expTwoBetter}$-$0.8\% & 137 & \textbf{135} & \cellcolor{expTwoBetter}$-$1.3\% \\
$\mathcal{S}$ GPU (ms) &  & 5,085 & \textbf{1,613} & \cellcolor{expTwoBetter}$\times$3.2 & 4,903 & \textbf{2,035} & \cellcolor{expTwoBetter}$\times$2.4 & 8,950 & \textbf{2,281} & \cellcolor{expTwoBetter}$\times$3.9 \\
$\mathcal{R}$ TTFT (ms) &  & 13.7 & \textbf{6.1} & \cellcolor{expTwoBetter}$\times$2.3 & 29.9 & \textbf{11.7} & \cellcolor{expTwoBetter}$\times$2.6 & 9.1 & \textbf{4.9} & \cellcolor{expTwoBetter}$\times$1.9 \\
\addlinespace
\rowcolor{expTwoSection}
 & \textit{Low} & \multicolumn{3}{c}{\textit{$\approx$18$\,\times$ compression ($P_{\mathrm{sum},3}$)}} & \multicolumn{3}{c}{\textit{$\approx$22$\,\times$ ($P_{\mathrm{sum},3}$)}} & \multicolumn{3}{c}{\textit{$\approx$21$\,\times$ ($P_{\mathrm{sum},3}$)}} \\
Score &  & 53.6 & \textbf{61.2} & \cellcolor{expTwoBetter}+7.7 & 58.8 & \textbf{73.7} & \cellcolor{expTwoBetter}+14.9 & 57.6 & \textbf{64.6} & \cellcolor{expTwoBetter}+7.0 \\
Tokens &  & 303 & \textbf{299} & \cellcolor{expTwoBetter}$-$1.1\% & 254 & \textbf{246} & \cellcolor{expTwoBetter}$-$3.5\% & 263 & \textbf{257} & \cellcolor{expTwoBetter}$-$2.3\% \\
$\mathcal{S}$ GPU (ms) &  & 7,699 & \textbf{1,784} & \cellcolor{expTwoBetter}$\times$4.3 & 7,131 & \textbf{2,088} & \cellcolor{expTwoBetter}$\times$3.4 & 19,700 & \textbf{2,534} & \cellcolor{expTwoBetter}$\times$7.8 \\
$\mathcal{R}$ TTFT (ms) &  & 17.8 & \textbf{6.1} & \cellcolor{expTwoBetter}$\times$2.9 & 38.3 & \textbf{11.7} & \cellcolor{expTwoBetter}$\times$3.3 & 14.5 & \textbf{5.0} & \cellcolor{expTwoBetter}$\times$2.9 \\
\addlinespace
\rowcolor{expTwoSection}
 & \textit{Medium-low} & \multicolumn{3}{c}{\textit{$\approx$11$\,\times$ compression ($P_{\mathrm{sum},5}$)}} & \multicolumn{3}{c}{\textit{$\approx$17$\,\times$ ($P_{\mathrm{sum},5}$)}} & \multicolumn{3}{c}{\textit{$\approx$8$\,\times$ ($P_{\mathrm{sum},4}$)}} \\
Score &  & 55.6 & \textbf{62.7} & \cellcolor{expTwoBetter}+7.1 & 60.2 & \textbf{74.9} & \cellcolor{expTwoBetter}+14.7 & 63.5 & \textbf{67.8} & \cellcolor{expTwoBetter}+4.3 \\
Tokens &  & 528 & \textbf{490} & \cellcolor{expTwoBetter}$-$7.2\% & 336 & \textbf{318} & \cellcolor{expTwoBetter}$-$5.3\% & 769 & \textbf{675} & \cellcolor{expTwoBetter}$-$12.2\% \\
$\mathcal{S}$ GPU (ms) &  & 15,439 & \textbf{2,150} & \cellcolor{expTwoBetter}$\times$7.2 & 9,097 & \textbf{2,251} & \cellcolor{expTwoBetter}$\times$4.0 & 48,189 & \textbf{3,304} & \cellcolor{expTwoBetter}$\times$14.6 \\
$\mathcal{R}$ TTFT (ms) &  & 29.3 & \textbf{6.3} & \cellcolor{expTwoBetter}$\times$4.7 & 45.5 & \textbf{11.8} & \cellcolor{expTwoBetter}$\times$3.9 & 29.3 & \textbf{5.3} & \cellcolor{expTwoBetter}$\times$5.6 \\
\addlinespace
\rowcolor{expTwoSection}
 & \textit{Medium} & \multicolumn{3}{c}{\textit{$\approx$9$\,\times$ compression ($P_{\mathrm{sum},4}$)}} & \multicolumn{3}{c}{\textit{$\approx$16$\,\times$ ($P_{\mathrm{sum},4}$)}} & \multicolumn{3}{c}{\textit{$\approx$6$\,\times$ ($P_{\mathrm{sum},5}$)}} \\
Score &  & 55.6 & \textbf{63.1} & \cellcolor{expTwoBetter}+7.4 & 60.8 & \textbf{75.1} & \cellcolor{expTwoBetter}+14.3 & 61.8 & \textbf{68.2} & \cellcolor{expTwoBetter}+6.4 \\
Tokens &  & 658 & \textbf{599} & \cellcolor{expTwoBetter}$-$9.0\% & 353 & \textbf{338} & \cellcolor{expTwoBetter}$-$4.3\% & 996 & \textbf{900} & \cellcolor{expTwoBetter}$-$9.6\% \\
$\mathcal{S}$ GPU (ms) &  & 18,896 & \textbf{2,363} & \cellcolor{expTwoBetter}$\times$8.0 & 10,459 & \textbf{2,265} & \cellcolor{expTwoBetter}$\times$4.6 & 59,281 & \textbf{3,867} & \cellcolor{expTwoBetter}$\times$15.3 \\
$\mathcal{R}$ TTFT (ms) &  & 34.5 & \textbf{6.3} & \cellcolor{expTwoBetter}$\times$5.4 & 50.1 & \textbf{11.8} & \cellcolor{expTwoBetter}$\times$4.2 & 35.5 & \textbf{5.4} & \cellcolor{expTwoBetter}$\times$6.6 \\
\bottomrule
\end{tabular}
\endgroup
\end{table*}

\begin{table*}[t]
\caption{
Experiment 2, SQuAD. \ttot{}: sender summarizes to receiver. \swat{}+AM: the sender's cache rendered and compressed with Attention Matching. Budgets are determined by the sender agent in \ttot{} given one of five summary prompts and sorted from lowest to highest per pair. Change: score difference in points, relative token difference, and speedup.
}
\label{tab:exp2-appendix-squad}
\centering
\begingroup
\definecolor{expTwoSection}{HTML}{ECECEC}
\definecolor{expTwoBetter}{HTML}{D9EFD8}
\begin{tabular}{llrrrrrrrrr}
\toprule
 & & \multicolumn{3}{c}{Qwen3-4B $\rightarrow$ Llama3-8B} & \multicolumn{3}{c}{Llama3-8B $\rightarrow$ Qwen3-14B} & \multicolumn{3}{c}{Gemma3-12B $\rightarrow$ Qwen3-4B} \\
\cmidrule(lr){3-5}\cmidrule(lr){6-8}\cmidrule(lr){9-11}
Metric & Budget & \ttot{} & \swat{} & Change & \ttot{} & \swat{} & Change & \ttot{} & \swat{} & Change \\
\midrule
\rowcolor{expTwoSection}
 & \textit{Lowest} & \multicolumn{3}{c}{\textit{$\approx$3$\,\times$ compression ($P_{\mathrm{sum},1}$)}} & \multicolumn{3}{c}{\textit{$\approx$4$\,\times$ ($P_{\mathrm{sum},1}$)}} & \multicolumn{3}{c}{\textit{$\approx$4$\,\times$ ($P_{\mathrm{sum},1}$)}} \\
Score &  & 51.6 & \textbf{88.7} & \cellcolor{expTwoBetter}+37.1 & 48.4 & \textbf{87.5} & \cellcolor{expTwoBetter}+39.1 & 35.8 & \textbf{82.4} & \cellcolor{expTwoBetter}+46.6 \\
Tokens &  & 64 & \textbf{60} & \cellcolor{expTwoBetter}$-$5.5\% & 60 & \textbf{47} & \cellcolor{expTwoBetter}$-$22.9\% & 50 & \textbf{47} & \cellcolor{expTwoBetter}$-$7.2\% \\
$\mathcal{S}$ GPU (ms) &  & 1,852 & \textbf{1,068} & \cellcolor{expTwoBetter}$\times$1.7 & 1,272 & \textbf{935} & \cellcolor{expTwoBetter}$\times$1.4 & 2,923 & \textbf{1,220} & \cellcolor{expTwoBetter}$\times$2.4 \\
$\mathcal{R}$ TTFT (ms) &  & 6.2 & \textbf{3.1} & \cellcolor{expTwoBetter}$\times$2.0 & 11.5 & \textbf{5.9} & \cellcolor{expTwoBetter}$\times$1.9 & 4.0 & \textbf{2.5} & \cellcolor{expTwoBetter}$\times$1.6 \\
\addlinespace
\rowcolor{expTwoSection}
 & \textit{Lower} & \multicolumn{3}{c}{\textit{$\approx$3$\,\times$ compression ($P_{\mathrm{sum},2}$)}} & \multicolumn{3}{c}{\textit{$\approx$3$\,\times$ ($P_{\mathrm{sum},2}$)}} & \multicolumn{3}{c}{\textit{$\approx$3$\,\times$ ($P_{\mathrm{sum},2}$)}} \\
Score &  & 60.0 & \textbf{88.7} & \cellcolor{expTwoBetter}+28.7 & 57.2 & \textbf{89.4} & \cellcolor{expTwoBetter}+32.2 & 40.8 & \textbf{84.7} & \cellcolor{expTwoBetter}+43.9 \\
Tokens &  & 85 & \textbf{60} & \cellcolor{expTwoBetter}$-$29.2\% & 83 & \textbf{62} & \cellcolor{expTwoBetter}$-$25.2\% & 67 & \textbf{62} & \cellcolor{expTwoBetter}$-$7.5\% \\
$\mathcal{S}$ GPU (ms) &  & 2,438 & \textbf{1,068} & \cellcolor{expTwoBetter}$\times$2.3 & 1,756 & \textbf{970} & \cellcolor{expTwoBetter}$\times$1.8 & 3,956 & \textbf{1,103} & \cellcolor{expTwoBetter}$\times$3.6 \\
$\mathcal{R}$ TTFT (ms) &  & 7.2 & \textbf{3.1} & \cellcolor{expTwoBetter}$\times$2.3 & 13.6 & \textbf{5.9} & \cellcolor{expTwoBetter}$\times$2.3 & 4.7 & \textbf{2.6} & \cellcolor{expTwoBetter}$\times$1.8 \\
\addlinespace
\rowcolor{expTwoSection}
 & \textit{Low} & \multicolumn{3}{c}{\textit{$\approx$2$\,\times$ compression ($P_{\mathrm{sum},3}$)}} & \multicolumn{3}{c}{\textit{$\approx$2$\,\times$ ($P_{\mathrm{sum},3}$)}} & \multicolumn{3}{c}{\textit{$\approx$2$\,\times$ ($P_{\mathrm{sum},3}$)}} \\
Score &  & 67.5 & \textbf{88.8} & \cellcolor{expTwoBetter}+21.3 & 61.3 & \textbf{89.8} & \cellcolor{expTwoBetter}+28.5 & 47.0 & \textbf{87.0} & \cellcolor{expTwoBetter}+40.0 \\
Tokens &  & 104 & \textbf{90} & \cellcolor{expTwoBetter}$-$13.3\% & 105 & \textbf{93} & \cellcolor{expTwoBetter}$-$11.3\% & 98 & \textbf{93} & \cellcolor{expTwoBetter}$-$5.1\% \\
$\mathcal{S}$ GPU (ms) &  & 2,994 & \textbf{1,202} & \cellcolor{expTwoBetter}$\times$2.5 & 2,264 & \textbf{1,184} & \cellcolor{expTwoBetter}$\times$1.9 & 5,679 & \textbf{1,420} & \cellcolor{expTwoBetter}$\times$4.0 \\
$\mathcal{R}$ TTFT (ms) &  & 7.9 & \textbf{3.2} & \cellcolor{expTwoBetter}$\times$2.5 & 15.4 & \textbf{5.9} & \cellcolor{expTwoBetter}$\times$2.6 & 5.5 & \textbf{2.6} & \cellcolor{expTwoBetter}$\times$2.1 \\
\addlinespace
\rowcolor{expTwoSection}
 & \textit{Medium-low} & \multicolumn{3}{c}{\textit{$\approx$2$\,\times$ compression ($P_{\mathrm{sum},5}$)}} & \multicolumn{3}{c}{\textit{$\approx$2$\,\times$ ($P_{\mathrm{sum},4}$)}} & \multicolumn{3}{c}{\textit{$\approx$2$\,\times$ ($P_{\mathrm{sum},4}$)}} \\
Score &  & 74.3 & \textbf{88.8} & \cellcolor{expTwoBetter}+14.5 & 72.4 & \textbf{89.8} & \cellcolor{expTwoBetter}+17.4 & 59.4 & \textbf{87.0} & \cellcolor{expTwoBetter}+27.6 \\
Tokens &  & 169 & \textbf{90} & \cellcolor{expTwoBetter}$-$46.4\% & 171 & \textbf{93} & \cellcolor{expTwoBetter}$-$45.7\% & 178 & \textbf{93} & \cellcolor{expTwoBetter}$-$47.6\% \\
$\mathcal{S}$ GPU (ms) &  & 4,782 & \textbf{1,202} & \cellcolor{expTwoBetter}$\times$4.0 & 3,632 & \textbf{1,184} & \cellcolor{expTwoBetter}$\times$3.1 & 9,586 & \textbf{1,420} & \cellcolor{expTwoBetter}$\times$6.8 \\
$\mathcal{R}$ TTFT (ms) &  & 10.9 & \textbf{3.2} & \cellcolor{expTwoBetter}$\times$3.4 & 21.3 & \textbf{5.9} & \cellcolor{expTwoBetter}$\times$3.6 & 8.0 & \textbf{2.6} & \cellcolor{expTwoBetter}$\times$3.1 \\
\addlinespace
\rowcolor{expTwoSection}
 & \textit{Medium} & \multicolumn{3}{c}{\textit{$\approx$2$\,\times$ compression ($P_{\mathrm{sum},4}$)}} & \multicolumn{3}{c}{\textit{$\approx$2$\,\times$ ($P_{\mathrm{sum},5}$)}} & \multicolumn{3}{c}{\textit{$\approx$2$\,\times$ ($P_{\mathrm{sum},5}$)}} \\
Score &  & 75.7 & \textbf{88.8} & \cellcolor{expTwoBetter}+13.1 & 65.8 & \textbf{89.8} & \cellcolor{expTwoBetter}+24.0 & 38.6 & \textbf{87.0} & \cellcolor{expTwoBetter}+48.4 \\
Tokens &  & 172 & \textbf{90} & \cellcolor{expTwoBetter}$-$47.4\% & 177 & \textbf{93} & \cellcolor{expTwoBetter}$-$47.4\% & 179 & \textbf{93} & \cellcolor{expTwoBetter}$-$48.1\% \\
$\mathcal{S}$ GPU (ms) &  & 4,829 & \textbf{1,202} & \cellcolor{expTwoBetter}$\times$4.0 & 3,740 & \textbf{1,184} & \cellcolor{expTwoBetter}$\times$3.2 & 9,637 & \textbf{1,420} & \cellcolor{expTwoBetter}$\times$6.8 \\
$\mathcal{R}$ TTFT (ms) &  & 11.1 & \textbf{3.2} & \cellcolor{expTwoBetter}$\times$3.5 & 21.5 & \textbf{5.9} & \cellcolor{expTwoBetter}$\times$3.6 & 8.0 & \textbf{2.6} & \cellcolor{expTwoBetter}$\times$3.1 \\
\bottomrule
\end{tabular}
\endgroup
\end{table*}

\begin{table*}[t]
\caption{
Geometric mean performance of SwaT over three sender--receiver pairs, seven datasets and five budgets. T2T and SwaT columns are geometric means over all datasets and budgets; Change is their ratio. Score $\Delta$ is the arithmetic mean score difference in points.
}
\label{tab:exp2-geomean}
\centering
\begingroup
\definecolor{expTwoBetter}{HTML}{D9EFD8}
\begin{tabular}{lrrrrrrrrrrrr}
\toprule
 & \multicolumn{3}{c}{Qwen3-4B $\rightarrow$ Llama3-8B} & \multicolumn{3}{c}{Llama3-8B $\rightarrow$ Qwen3-14B} & \multicolumn{3}{c}{Gemma3-12B $\rightarrow$ Qwen3-4B} & \multicolumn{3}{c}{All pairs} \\
\cmidrule(lr){2-4}\cmidrule(lr){5-7}\cmidrule(lr){8-10}\cmidrule(lr){11-13}
Metric & T2T & SwaT & Change & T2T & SwaT & Change & T2T & SwaT & Change & T2T & SwaT & Change \\
\midrule
Score & 38.8 & \textbf{54.8} & \cellcolor{expTwoBetter}+41.2\% & 39.2 & \textbf{59.1} & \cellcolor{expTwoBetter}+50.6\% & 34.1 & \textbf{52.8} & \cellcolor{expTwoBetter}+54.9\% & 37.3 & \textbf{55.5} & \cellcolor{expTwoBetter}+48.8\% \\
Score $\Delta$ (pts) & 42.9 & \textbf{57.8} & \cellcolor{expTwoBetter}+14.9 & 42.9 & \textbf{61.9} & \cellcolor{expTwoBetter}+19.0 & 38.0 & \textbf{56.2} & \cellcolor{expTwoBetter}+18.2 & 41.3 & \textbf{58.6} & \cellcolor{expTwoBetter}+17.4 \\
Tokens & 293 & \textbf{254} & \cellcolor{expTwoBetter}$-$13.1\% & 269 & \textbf{243} & \cellcolor{expTwoBetter}$-$9.7\% & 309 & \textbf{269} & \cellcolor{expTwoBetter}$-$13.1\% & 290 & \textbf{255} & \cellcolor{expTwoBetter}$-$12.0\% \\
$\mathcal{S}$ GPU (ms) & 8,752 & \textbf{2,150} & \cellcolor{expTwoBetter}$\times$4.1 & 6,198 & \textbf{2,360} & \cellcolor{expTwoBetter}$\times$2.6 & 19,761 & \textbf{2,888} & \cellcolor{expTwoBetter}$\times$6.8 & 10,234 & \textbf{2,447} & \cellcolor{expTwoBetter}$\times$4.2 \\
$\mathcal{R}$ TTFT (ms) & 18.0 & \textbf{4.0} & \cellcolor{expTwoBetter}$\times$4.5 & 31.8 & \textbf{7.5} & \cellcolor{expTwoBetter}$\times$4.3 & 14.1 & \textbf{3.4} & \cellcolor{expTwoBetter}$\times$4.2 & 20.1 & \textbf{4.7} & \cellcolor{expTwoBetter}$\times$4.3 \\
\bottomrule
\end{tabular}
\endgroup
\end{table*}

\begin{figure*}[t]
  \centering
  \includegraphics[draft=false,width=\textwidth]{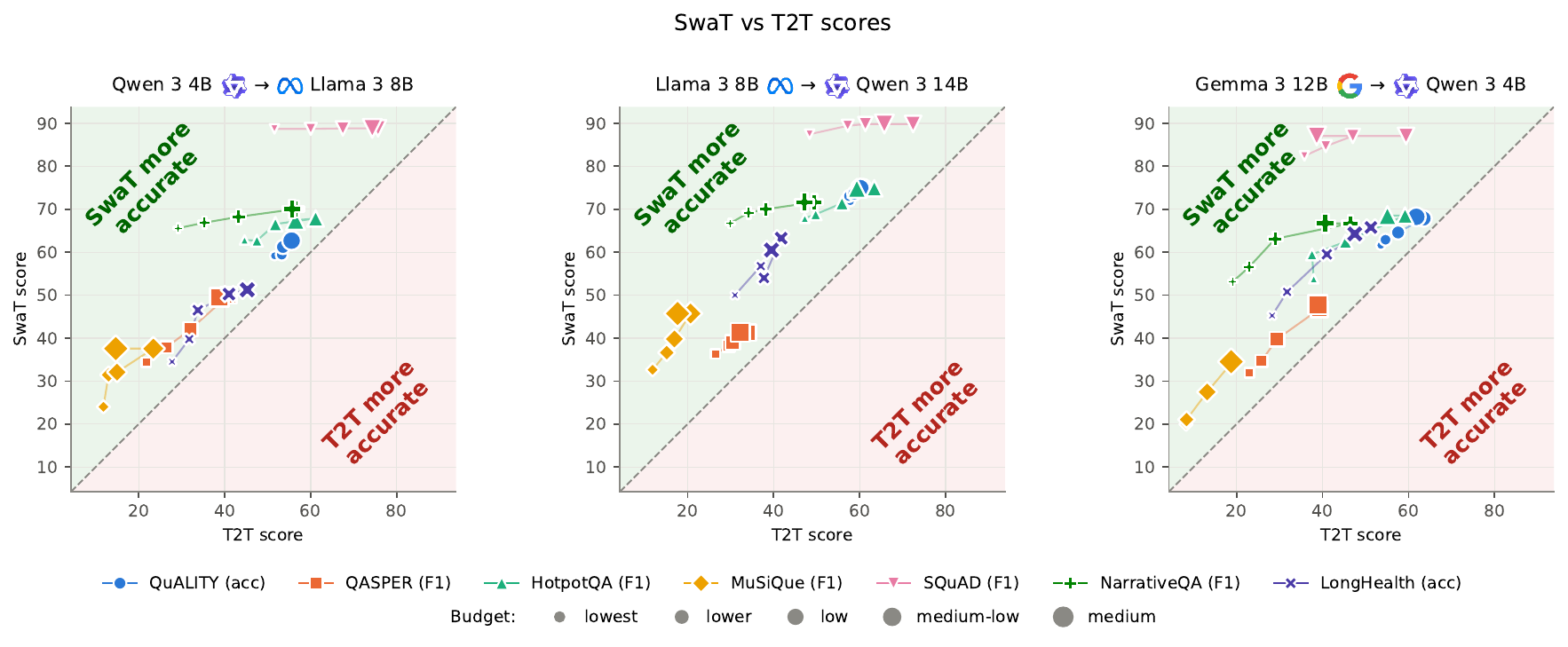}
  \par\medskip
  \includegraphics[draft=false,width=\textwidth]{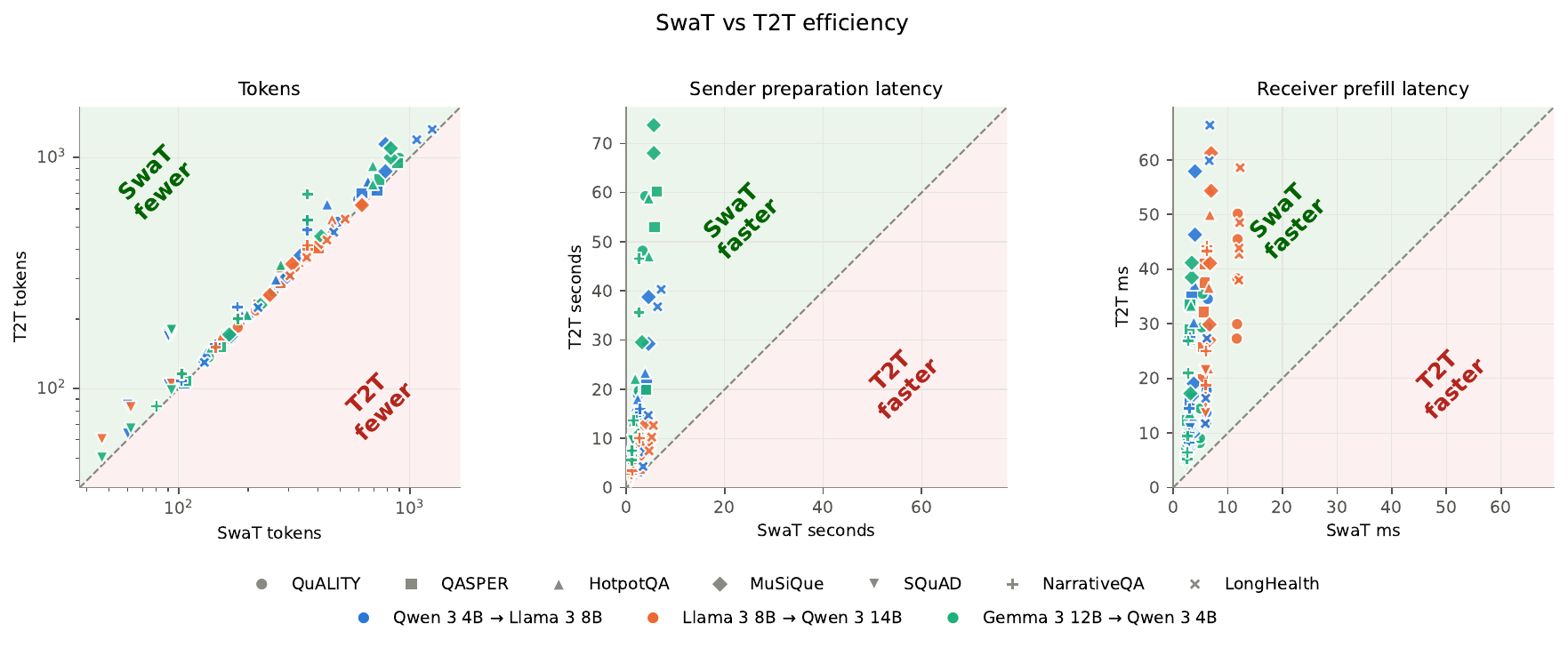}
  \caption{Scatter plot for all individual cells in \cref{tab:exp2-appendix-hotpotqa,tab:exp2-appendix-longhealth,tab:exp2-appendix-musique,tab:exp2-appendix-narrativeqa,tab:exp2-appendix-qasper,tab:exp2-appendix-quality,tab:exp2-appendix-squad}. We note that points in the tokens plot are close to the diagonal by construction as the two methods were picked to have close to matching budgets.}
  \Description{Description}
  \label{fig:exp2_6images}
\end{figure*}

% \clearpage

% \input{input/compute_flop}

% \input{input/compute_speed}

% \clearpage

% Arditi et al. 2024 ("Refusal in language models is mediated by a single direction") show a refusal direction written into the residual stream at the last prompt positions, and Qi et al. 2024 ("Safety alignment should be made more than just a few tokens deep") show the decision is made in the first few output tokens.
% \input{input/security}

% \clearpage

% \begin{figure*}[t]
%   \centering
%   \includegraphics[draft=false,width=\textwidth]{figures/refusal_direction.pdf}
%   \caption{caption for the two plots.}
%   \Description{Description}
%   \label{fig:refusal_causal}
% \end{figure*}

\clearpage

\section{Ablation training}

In \cref{tab:ablations} we provide ablations of our training procedure described in \cref{sec:method}.

\begin{table}[t]
\caption{
Ablations of our \swat{} architecture. Each method is compared against the average of two independently trained \swat{} models. Differences are in percentage points on the mean benchmark score (accuracy or F1). Metrics calculated over 13 benchmarks with the pair Qwen3-4B $\rightarrow$ Llama-3-8B.
}
\label{tab:ablations}
\centering
\begin{tabular}{p{0.55\columnwidth}r}
\toprule
Ablation & $\Delta$ \swat{} \\
\midrule
\swat{} & $0$ \\
\swat{} (w/ c2c \cite{kv_c2c} token alignment) & $-3.20$ \\
Headwise MLP & $-14.71$ \\
Phase 1 only & $-36.97$ \\
Phase 2 only & $-36.96$ \\
Dense loss only & $-55.11$ \\
\bottomrule
\end{tabular}
\end{table}

\clearpage

\section{Beyond the Receiver's Context Window}
\label{apdx-sec:exp-long-context}

Another zero-shot capability is the ability to provide the receiver with the ability to answer questions about a context beyond its context length. We are able to do this by taking a sender with support for a long context length (such as Qwen 3 which supports a 32k context length) and pairing it with a receiver that only supports an 8k context length. Then, we utilize the sender to generate the KV-cache of a long document, then we chunk the KV-cache into 8k chunks, transfer and compact each one using \swat{}+\am{}, then combine the compacted chunks into a single chunk that the receiver can use. We note that while we do chunk the KV-cache, this is only done after the sender has prefilled the entire document. Thus, each chunk contains cross-dependencies. This is unlike similar capabilities provided by \am{} which utilize the sender's context to compress each chunk which does not capture cross--chunk dependencies. We show the results for this on a version of HotpotQA where each context was inflated (using distractors from other samples) into 28k tokens thus far exceeding the receiver's 8k limit. We report the results in \cref{tab:beyond-context-hotpotqa}.

\begin{table}[thb]
\caption{
Interleaved HotpotQA (500 questions; EM/F1 in \%). Sequence lengths is number of text tokens or KV-cache tokens.
}
\label{tab:beyond-context-hotpotqa}
\centering
\begin{tabular}{lrrr}
\toprule
Method & \shortstack{Seq. \\ len} & EM & F1 \\
\midrule
\multicolumn{4}{l}{\textbf{Baselines}} \\
Qwen 3 4B & 28K & 20.4 & 36.75 \\
Llama 3 8B truncate & 8K & 5.4 & 13.32 \\
Llama 3 8B $\mathrm{AM}_{20}$ & 5.6K & 21.2 & 34.95 \\
\multicolumn{4}{l}{\textbf{\swat{}}} \\
Qwen $\rightarrow^{\mathrm{AM}_{20}}$ Llama & 5.6K & \textbf{24.8} & \textbf{41.87} \\
\bottomrule
\end{tabular}
\end{table}

\section{Additional Vision Transfer}
\label{apdx-sec:vision}

We add an additional ablation to the zero-shot vision transfer in \cref{tab:vision-subset-g2g} where we train a \swat{} to transfer from a Gemma~3~4B to itself; this measures the effect of the training data on the results. We do see that this ablation does improve the scores but not to the same degree as the Gemma to Granite pair.

Additionally, we show that zero-shot visual transfer is far from perfect. By testing two datasets in \cref{tab:vision-subset-detail} that require heavy visual processing, we show that the sender VLM model alone achieves a much higher accuracy than both \ttot{} and \swat{}. This shows that the zero-shot transfer is not transferring full vision capabilities especially on datasets that require such heavy processing.

\begin{table*}[t]
\caption{
Similar to \cref{tab:vision-subset}.
Test accuracy (\%) for multiple pairs including a control where sender and receiver are the same VLM (Gemma 3 4B $\rightarrow$ Gemma 3 4B pair).
}
\label{tab:vision-subset-g2g}
\centering
\begin{tabular}{lrrrrrr}
\toprule
Method & VQAv2 & TQA & OKVQA & GQA & VW & Avg. \\
\midrule
\multicolumn{7}{l}{\makebox[2ex]{\raisebox{-0.4ex}{\includegraphics[height=2ex,draft=false]{assets/logos/gemma.png}}} \textbf{Gemma 3 4B} $\rightarrow$ \makebox[2ex]{\raisebox{-0.4ex}{\includegraphics[height=2ex,draft=false]{assets/logos/gemma.png}}} \textbf{Gemma 3 4B}} \\
Sender & 45.3$^*$ & 53.4$^*$ & 22.0 & 31.4$^*$ & \textbf{19.0} & 34.2$\pm$.4$^*$ \\
\swat{} & \textbf{53.8} & \textbf{56.6} & \textbf{22.4} & \textbf{35.3} & \textbf{19.0} & \textbf{37.4$\pm$.4} \\
\multicolumn{7}{l}{\makebox[2ex]{\raisebox{-0.4ex}{\includegraphics[height=2ex,draft=false]{assets/logos/gemma.png}}} \textbf{Gemma 3 4B} $\rightarrow$ \makebox[2ex]{\raisebox{-0.4ex}{\includegraphics[height=2ex,draft=false]{assets/logos/granite.png}}} \textbf{Granite 3.3 8B}} \\
Sender & 45.3$^*$ & 53.4$^*$ & 22.0$^*$ & 31.4$^*$ & \textbf{19.0} & 34.2$\pm$.4$^*$ \\
\ttot{} & 33.8$^*$ & 32.0$^*$ & 15.7$^*$ & 16.3$^*$ & 7.7$^*$ & 21.1$\pm$.4$^*$ \\
\swat{} & \textbf{59.5} & \textbf{61.8} & \textbf{28.7} & \textbf{39.5} & 18.3 & \textbf{41.6$\pm$.4} \\
\multicolumn{7}{l}{\makebox[2ex]{\raisebox{-0.4ex}{\includegraphics[height=2ex,draft=false]{assets/logos/gemma.png}}} \textbf{Gemma 3 4B} $\rightarrow$ \makebox[2ex]{\raisebox{-0.4ex}{\includegraphics[height=2ex,draft=false]{assets/logos/llama.png}}} \textbf{Llama 3 8B}} \\
Sender & 45.3$^*$ & 53.4$^*$ & 22.0$^*$ & 31.4$^*$ & \textbf{19.0} & 34.2$\pm$.4$^*$ \\
\ttot{} & 16.4$^*$ & 12.3$^*$ & 18.3$^*$ & 10.9$^*$ & 8.4$^*$ & 13.3$\pm$.3$^*$ \\
\swat{} & \textbf{57.5} & \textbf{56.4} & \textbf{29.7} & \textbf{35.5} & 18.7 & \textbf{39.6$\pm$.4} \\
\multicolumn{7}{l}{\makebox[2ex]{\raisebox{-0.4ex}{\includegraphics[height=2ex,draft=false]{assets/logos/gemma.png}}} \textbf{Gemma 3 12B} $\rightarrow$ \makebox[2ex]{\raisebox{-0.4ex}{\includegraphics[height=2ex,draft=false]{assets/logos/qwen.png}}} \textbf{Qwen 3 4B}} \\
Sender & 44.5$^*$ & 22.4$^*$ & \textbf{24.4} & 28.3$^*$ & \textbf{20.0} & 27.9$\pm$.4$^*$ \\
\ttot{} & 47.0$^*$ & 37.6$^*$ & 19.1$^*$ & 29.7$^*$ & 12.2$^*$ & 29.1$\pm$.4$^*$ \\
\swat{} & \textbf{51.7} & \textbf{49.4} & 21.8 & \textbf{33.0} & 15.1 & \textbf{34.2$\pm$.4} \\
\bottomrule
\end{tabular}
\end{table*}

\begin{table*}[t]
\caption{
Similar to \cref{tab:vision-subset}.
Test accuracy (\%) for multiple pairs on
DocVQA and InfoVQA, which depend on dense text and fine detail in the image.
}
\label{tab:vision-subset-detail}
\centering
\begin{tabular}{lrrr}
\toprule
Method & DocVQA & InfoVQA & Avg. \\
\midrule
\multicolumn{4}{l}{\makebox[2ex]{\raisebox{-0.4ex}{\includegraphics[height=2ex,draft=false]{assets/logos/gemma.png}}} \textbf{Gemma 3 4B} $\rightarrow$ \makebox[2ex]{\raisebox{-0.4ex}{\includegraphics[height=2ex,draft=false]{assets/logos/gemma.png}}} \textbf{Gemma 3 4B}} \\
Sender & \textbf{73.3} & \textbf{41.6} & \textbf{57.4$\pm$.8} \\
\swat{} & 40.9 & 26.8 & 33.8$\pm$.8 \\
\multicolumn{4}{l}{\makebox[2ex]{\raisebox{-0.4ex}{\includegraphics[height=2ex,draft=false]{assets/logos/gemma.png}}} \textbf{Gemma 3 4B} $\rightarrow$ \makebox[2ex]{\raisebox{-0.4ex}{\includegraphics[height=2ex,draft=false]{assets/logos/granite.png}}} \textbf{Granite 3.3 8B}} \\
Sender & \textbf{73.3} & \textbf{41.6} & \textbf{57.4$\pm$.8} \\
\ttot{} & 39.3$^*$ & 25.5$^*$ & 32.4$\pm$.8$^*$ \\
\swat{} & 43.4 & 31.5 & 37.5$\pm$.8 \\
\multicolumn{4}{l}{\makebox[2ex]{\raisebox{-0.4ex}{\includegraphics[height=2ex,draft=false]{assets/logos/gemma.png}}} \textbf{Gemma 3 4B} $\rightarrow$ \makebox[2ex]{\raisebox{-0.4ex}{\includegraphics[height=2ex,draft=false]{assets/logos/llama.png}}} \textbf{Llama 3 8B}} \\
Sender & \textbf{73.3} & \textbf{41.6} & \textbf{57.4$\pm$.8} \\
\ttot{} & 38.4$^*$ & 26.6$^*$ & 32.5$\pm$.8$^*$ \\
\swat{} & 43.3 & 28.6 & 36.0$\pm$.8 \\
\multicolumn{4}{l}{\makebox[2ex]{\raisebox{-0.4ex}{\includegraphics[height=2ex,draft=false]{assets/logos/gemma.png}}} \textbf{Gemma 3 12B} $\rightarrow$ \makebox[2ex]{\raisebox{-0.4ex}{\includegraphics[height=2ex,draft=false]{assets/logos/qwen.png}}} \textbf{Qwen 3 4B}} \\
Sender & \textbf{78.1} & \textbf{46.3} & \textbf{62.2$\pm$.8} \\
\ttot{} & 48.1 & 29.3 & 38.7$\pm$.8 \\
\swat{} & 35.1 & 23.0 & 29.0$\pm$.7 \\
\bottomrule
\end{tabular}
\end{table*}

% \clearpage

% \input{figures/vllm}

%%%%%%%%%%%%%%%%%%%%%%%%%%%%%%%%%%%%%%%%%%%%%%%%%%%%%%%%%%%%%%%%%%%%%%%%

\end{document}